\documentclass{article}

\usepackage{microtype}
\usepackage{graphicx}
\usepackage{subfigure}
\usepackage{booktabs} %
\usepackage{xspace}

\usepackage{hyperref}
\usepackage{xurl}

\usepackage[accepted]{styles/icml2025}

\usepackage{amsmath}
\usepackage{amssymb}
\usepackage{mathtools}
\usepackage{amsthm}

\usepackage[capitalize]{cleveref}

\usepackage{graphicx}
\usepackage{caption}
\usepackage{microtype}
\usepackage{enumitem}
\usepackage{hyperref}
\usepackage{url}
\usepackage{booktabs}
\definecolor{darkblue}{rgb}{0, 0, 0.5}
\hypersetup{colorlinks=true, citecolor=darkblue, linkcolor=darkblue, urlcolor=darkblue}

\setlist[itemize]{leftmargin=*, itemsep=1pt, topsep=1pt, parsep=0pt}
\setlist[enumerate]{leftmargin=*, itemsep=1pt, topsep=1pt, parsep=0pt}

\usepackage{amsthm}
\usepackage{amsmath}
\crefname{definition}{Def.}{Defs.}
\crefname{lstlisting}{Listing}{Listings} %
\Crefname{lstlisting}{Listing}{Listings} %

\usepackage{algorithm}
\usepackage{algorithmic}

\usepackage[textsize=tiny]{todonotes}

\usepackage{multirow}
\usepackage{longtable}
\usepackage{listings}

\usepackage[utf8]{inputenc}
\usepackage{tcolorbox}
\tcbuselibrary{listings} %
\tcbuselibrary{breakable}

\usepackage{mdframed}

\usepackage{multicol} %
\usepackage{lipsum}   %
\usepackage{tcolorbox}
\usepackage{xcolor}

\newcommand{\marker}[2]{\ensuremath{^{\textsc{#1}}_{\textsc{#2}}}}

\newif\ifshowcomments
\showcommentstrue

\ifshowcomments
\newcommand\red[1]{{\color{red}{#1} }}
\newcommand\ken[1]{{\color{blue}\textbf{Ken: }{#1} }}
\newcommand\cc[1]{{\color{teal}\textbf{}{#1}}}
\newcommand\nicolas[1]{{\color{orange}\textbf{Nicolas: }{#1}}}
\newcommand\mcj[1]{{\color{orange}\textbf{Matthew: }{#1}}}

\newcommand{\ccside}[1]{\todo[linecolor=green,backgroundcolor=green!10,bordercolor=green]{\marker CC: #1}}

\newcommand{\percyside}[1]{\todo[linecolor=red,backgroundcolor=red!10,bordercolor=red]{\marker Percy: #1}}

\else
\newcommand\todo[1]{}
\newcommand\red[1]{}
\newcommand\ken[1]{}
\newcommand\cc[1]{}
\newcommand\nicolas[1]{}
\newcommand\mcj[1]{}

\newcommand{\ccside}[1]{}

\fi

\newcommand{\prompt}{\ensuremath{p\xspace}}
\newcommand{\suffix}{\ensuremath{s\xspace}}

\newcommand{\M}{\ensuremath{\smash{\mathcal{M}}}}
\newcommand{\Mbase}{\ensuremath{\smash{\mathcal{M}_{\text{base}}}}}
\newcommand{\Mcounterfactual}{\ensuremath{\smash{\mathcal{M}_{\text{cf}}}}}
\newcommand{\Mfilter}[1]{\ensuremath{\smash{\mathcal{M}^{(#1)}_{\text{filter}}}}}

\newcommand{\D}{\ensuremath{\smash{\mathcal{D}}}}
\newcommand{\Dbase}{\ensuremath{\smash{\mathcal{D}_{\text{base}}}}}
\newcommand{\Dmem}{\ensuremath{\smash{\mathcal{D}_{\text{mem}}}}}
\newcommand{\Dfilter}[1]{\ensuremath{\smash{\mathcal{D}^{(#1)}_{\text{filter}}}}}
\newcommand{\Dlinger}[1]{\ensuremath{\smash{\mathcal{D}^{(#1)}_{\text{linger}}}}}
\newcommand{\Dfinetune}{\ensuremath{\smash{\mathcal{D}_{\text{ft}}}}}

\newcommand{\ngramsfunc}{\ensuremath{\smash{\operatorname{n-grams}}}}

\newcommand{\ngram}{\ensuremath{n}-gram\@\xspace}

\theoremstyle{plain}

\newtheorem{definition}{Definition}[section]

\theoremstyle{remark}

\newcommand{\titletext}{Language Models May Verbatim Complete Text They Were Not Explicitly Trained On}

\icmltitlerunning{\titletext}

\begin{document}

\twocolumn[
\icmltitle{Language Models May Verbatim Complete Text\\They Were Not Explicitly Trained On}

\icmlsetsymbol{equal}{*}

\begin{icmlauthorlist}
\icmlauthor{Ken Ziyu Liu}{g,ats}
\icmlauthor{Christopher A. Choquette-Choo}{equal,g}
\icmlauthor{Matthew Jagielski}{equal,g}
\icmlauthor{Peter Kairouz}{g}
\icmlauthor{Sanmi Koyejo}{s}
\icmlauthor{Percy Liang}{s}
\icmlauthor{Nicolas Papernot}{g}
\end{icmlauthorlist}

\icmlaffiliation{g}{Google}
\icmlaffiliation{ats}{Work completed while on internship at Google DeepMind. Now at Stanford University.}
\icmlaffiliation{s}{Stanford University}

\icmlcorrespondingauthor{Ken Liu}{kzliu@cs.stanford.edu}
\icmlcorrespondingauthor{Christopher A. Choquette-Choo}{cchoquette@google.com}

\icmlkeywords{Machine Learning, ICML}

\vskip 0.3in
]

\newcommand\pl[1]{\percyside{#1}}

\printAffiliationsAndNotice{\icmlEqualContribution} %

\begin{abstract}
An important question today is whether a given text was used to train a large language model (LLM). A \emph{completion} test is often employed: check if the LLM completes a sufficiently complex text.
This, however, requires a ground-truth definition of membership; most commonly, it is defined as a member based on the $n$-gram overlap between the target text and any text in the dataset.
In this work, we demonstrate that this $n$-gram based membership definition can be effectively gamed.
We study scenarios where sequences are \emph{non-members} for a given $n$ and we find that completion tests still succeed. We find many natural cases of this phenomenon by retraining LLMs from scratch after removing all training samples that were completed; these cases include exact duplicates, near-duplicates, and even short overlaps. They showcase that it is difficult to find a single viable choice of $n$ for membership definitions. Using these insights, we design adversarial datasets that can cause a given target sequence to be completed without containing it, for any reasonable choice of $n$.
Our findings highlight the inadequacy of $n$-gram membership, suggesting membership definitions fail to account for auxiliary information available to the training algorithm.

\end{abstract}
\vspace{-15pt}

\section{Introduction}
\label{sec:intro}

\textit{Training data membership} asks whether a data point was used to train a given model. For large language models (LLMs), it is useful to answer questions related to privacy (e.g., is the LLM leaking information contained in the text it was trained on?), copyright (e.g., has the model been trained on copyrighted text?), and more generally AI safety (e.g., did the LLM successfully unlearn text that was identified as harmful post hoc training?). In each of these settings, it is important that the evaluation of membership be robust. %

With direct access to the training dataset, determining sequence membership is straightforward given a fixed criterion for defining when two text sequences are ``the same.'' A common approach is to compare sequences by their $n$-grams, as this is both efficient and reasonable. Two sequences match \emph{verbatim} if all their $n$-grams are equal for all $n$, and \emph{approximately} if most do for some $n$~\cite{lee2021deduplicating}. However, this definition must align with downstream concerns in privacy, copyright, and safety—domains focused on what an LLM can reproduce as an intuitive notion of membership.

Consider a setting where a long text sequence is generated verbatim by an LLM. 
The sequence has high entropy due to its length, which makes it very unlikely to have been generated by chance.
Thus, one may intuitively conclude that this sequence was \emph{in} the training data.
This raises a key question: can an LLM generate a target sequence even if it was never included as an $n$-gram in its training data?  
Our findings confirm that the answer is affirmative: $n$-gram membership establishes a threshold dependent on $n$, and this threshold can be \textit{gamed}. In other words, our findings show that that formally  defining a robust and accurate notion of membership is challenging.

\begin{figure*}[t]
    \centering

    \includegraphics[width=\linewidth]{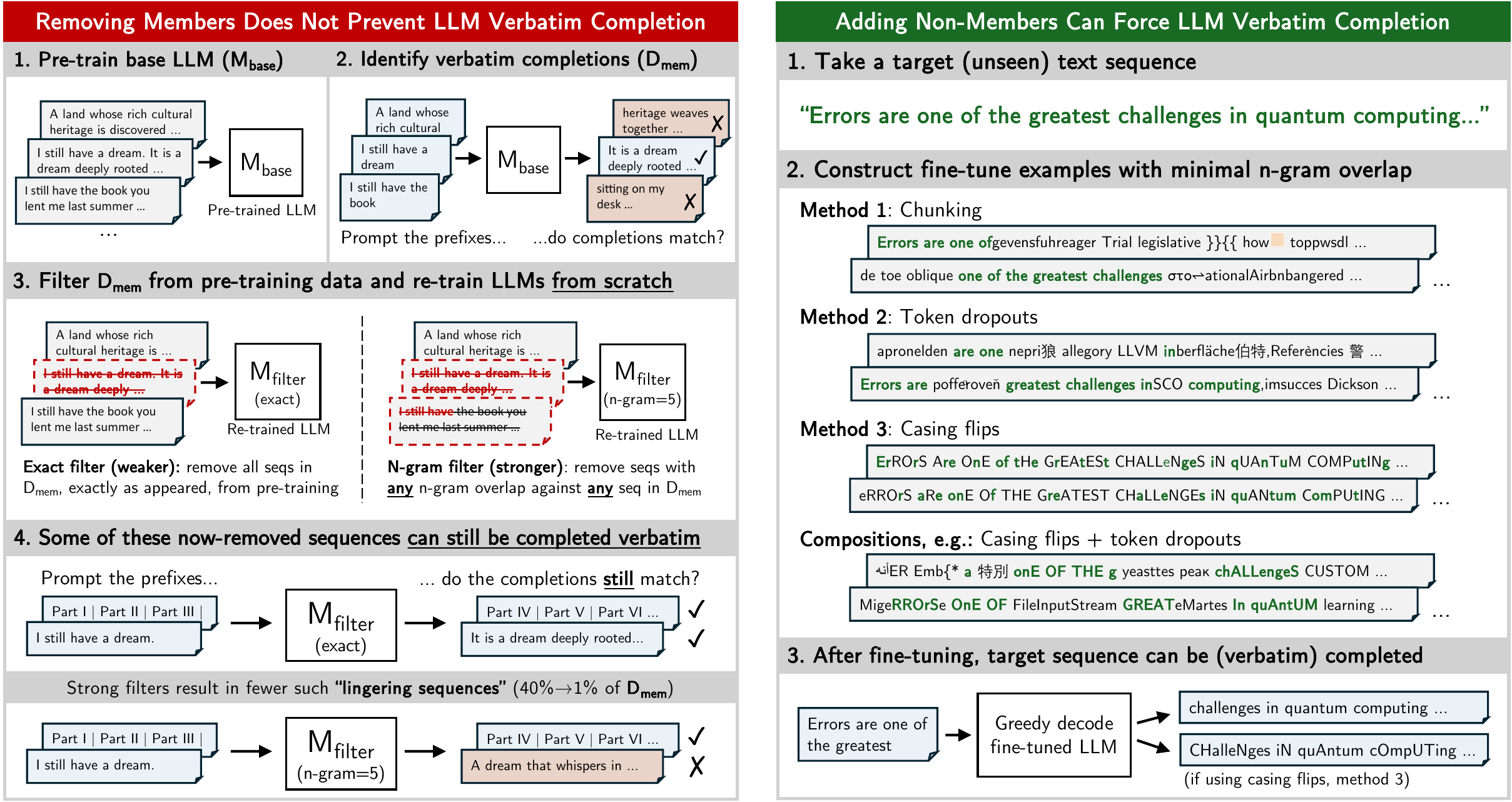}
    \vspace{-1em}
    \caption{\textbf{Main setup and findings}: a text sequence can be (verbatim) completed by a language model without being a $n$-gram ``member'' of its training set.
    \textbf{Left (\S\ref{sec:pre-training}):} We pretrain a model and remove extracted training dataset of length $k$ from its training data with either exact $(k=n)$-gram filters or stronger approximate $(k>n)$-gram membership filters. We find some sequences ($\approx 40\%$ with exact filters or $\approx1\%$ with approximate filters) remain verbatim completed despite not being explicitly trained on.
    \textbf{Right (\S\ref{sec:fine-tuning}):} We show an LLM can be fine-tuned to verbatim complete a target unseen sequence, e.g., today's blog post, by using adversarially constructed datasets with \textit{no} $n$-gram overlap.
    }
    \label{fig:teaser}
    \vspace{-4mm}
\end{figure*}

We assess whether an LLM generates a target sequence by prompting it with a prefix and checking if it completes the corresponding suffix, a process we refer to as a \emph{completion test}. Language models are known to complete some of their training data~\cite{nasr2023scalable}.
In our work, we first find that even after removing a set of extracted sequences from the training dataset and retraining the LLM \textit{from scratch}, the retrained model can still verbatim 
complete 40\% of them under our experimental conditions (Section~\ref{sec:pre-training}). Upon investigation, we find that these removed yet still completed sequences are either \textit{de facto} members of the training set (but for a different definition of membership) or lacking sufficient complexity: many examples have near duplicates, sequences with $m<n$-grams that are not removed, or are explained by the model's generalization capabilities (e.g., patterns or counting).

This result, however, leads to our second finding---there exists strategies for systematically gaming the $n$-gram membership definition.
That is, there are strategies for constructing a dataset $D$ that does not contain $n$-grams of a sequence $x$, yet, when a language model trains on $D$ it is able to complete $x$ verbatim.
In Section~\ref{sec:fine-tuning}, we give multiple examples of such strategies, including one that has the model train on multiple $m$-grams of $x$, where $m<n$ and, in some cases, $m \ll n$. Our experiments show that we can systematically force a model to complete six sequences of interest $x$ despite these sequences not being a member of its training set per the $n$-gram membership definition.

Figure~\ref{fig:teaser} shows our setup. Our main takeaways are:
\begin{enumerate}
    \item We find that there is high overlap between training data membership and our LLM completion test being positive. 
    Text not in this overlap are explained by the lack of complexity or limitations in $n$-gram based definitions.
    \item $n$-gram membership is limited in capturing the intuition of what constitutes a training dataset ``member.''  %
    Indeed, our work shows that a model can complete sequences that are not $n$-gram members of its training dataset.
    \item We believe that the underlying cause of this limitation is not in the choice of the distance used to compare sequences---here $n$-gram overlap---but rather in the fact that the membership definition fails to consider auxiliary information that the training algorithm gets access to, e.g., through pre-processing or other design choices made throughout the ML pipeline. Here, the strategies we propose to game the $n$-gram definition exploit this by introducing auxiliary information through the very construction of the training dataset: e.g., we cannot construct $(n-1)$-grams of a sequence $x$ without knowing the entire sequence in the first place. 
\end{enumerate}

\section{Background \& Related Work}
\label{sec:background}

\textbf{Definitions of data membership.} 
Many language model tasks require a definition of data membership. In most cases, the definition falls into versions of \ngram, or {substring}, overlap~\cite{anil2023palm,team2023gemini,team2024gemini,team2024gemma,team2024gemma2,llama2,llama3, zhang2024language,duan2024membership,carlini2021extracting,singh2024evaluation}.
$n$-gram based definitions capture near-duplicates by matching smaller text segments; this is flexible, simple, and intuitive.
When studying \textit{data contamination},
much of the prior work uses $n$-gram based definitions~\citep{sainz2023nlp,jiang2024investigating,dekoninck2024evading,singh2024evaluation}. For example, GPT-4 considers 50-character substring overlap~\citep{achiam2023gpt} and Llama-3 considers 8-gram token overlap~\citep{llama3}.
For \textit{training data deduplication}~\citep{lee2021deduplicating,kandpal2022deduplicating,mou2023dedup},
duplicates are identified based on training data membership. Recent methods use suffix arrays for exact substring matches~\citep{lee2021deduplicating} and MinHash or locality sensitive hashing for approximate matches~\citep{broder1997resemblanceminhash,mou2023dedup}; both build on $n$-gram overlap.
The prevalent use of $n$-gram based definitions reflects a practical balance between accuracy and simplicity. A key focus of our work is to highlight the limitations of these $n$-gram based definitions.

\textbf{Tests for data membership.}
Unlike membership definitions, which define the ground-truth, membership tests aim to detect if a data sample was in a dataset.
There are many model-level membership tests in the literature that predict membership of text to a training dataset with only access to a trained model, and not the training dataset. Our work focuses on model-level membership tests because they are more relevant to the downstream uses of membership in LLMs, e.g., in privacy, copyright, and safety (see \S\ref{sec:intro}).

\textit{Membership inference attacks (MIA)}~\citep{shokri2017membership} are widely studied, especially in computer vision~\citep{yeom2018privacy,salem2018ml,sablayrolles2019white,choquette2021label,carlini2022membership,jagielski2024students}
and more recently %
example-level membership inference for LLMs~\citep{zarifzadeh2023low,shi2023detecting,mattern2023membership,li2023mope}.
Despite these attempts, progress is hindered by flawed evaluations~\citep{meeus2024inherent,zhang2024membership}:
\citet{duan2024membership} argue that membership can be inherently blurry for natural language,  \citet{das2024blind} report that existing MIA testbeds suffer from  distribution shifts, and  \citet{kong2023can} refute MIAs using a gradient-space attack.
Our work situates in this body of work by studying systematic failure modes of operationalizing membership through definitions and tests, and the consequences when these definitions and tests mismatch.

\textit{Dataset-level} MIAs enhance membership signals by leveraging multiple correlated samples as inputs~\citep{maini2021dataset,kandpal2023user,maini2024llm}. These are closely related to contamination tests~\citep{golchin2023data,oren2023proving}.
Our work focuses on sequence-level data membership tests based on data completion, because these focus on scenarios where the LLM generates the text, which presents novel concerns for privacy, copyright, and safety.

\textit{Data completion.}
There is a long body of work studying generation of training data, in diffusion models~\citep{somepalli2023diffusion,carlini2023extracting} and in LLMs~\citep{carlini2019secret,tirumala2022memorization,kudugunta2024madlad,biderman2024emergent,freeman2024exploring}.
These works are often studied from the perspective of studying memorization, where the entity performing the model test has access to the training dataset.
In this line of literature, there exist both verbatim definitions of memorization~\citep{carlini2022quantifying} and approximate definitions~\citep{ippolito2022preventing}.
When studied from a black-box perspective---without access to the training dataset---they typically match completions against known auxilliary databases as a surrogate confirmation of membership~\citep{carlini2021extracting,nasr2023scalable}.
Intuitively, if a model completes a long sequence $x$ when prompted with its prefix, it likely saw $x$ during training because $x$ has high entropy due to its length and vocabulary size~\cite{carlini2019secret,carlini2022quantifying}.
Our work focuses only on these completion tests 
as a black-box membership test.

\section{Preliminaries}
\label{sec:prelim}

We now formalize the key definitions that underlie our experiments. 
We focus on defining what it means for a sequence to be a ``member'' of the training set of a language model, and what constitutes ``completing'' a sequence as a means of testing its membership. Precise definitions of these notions anchor our study of the mismatch between them.

Modern language models operate on \textit{token sequences}, which are integer encodings of text strings via a byte-pair encoding (BPE) tokenizer~\cite{sennrich2015neural}. We use $x$ to denote a token sequence (rather than its text form) with length $|x|$, and $\smash{\operatorname{n-grams}(x) = \{x_{i:i+n}\}_{i=1}^{|x|-n}}$ to denote the set of $n$-grams derived from $x$.

\textbf{Definition of \textit{Data Membership}.}  We anchor on a simple and flexible membership definition for our experiments that encapsulates many variants used in the literature:
\begin{definition}[$n$-gram data membership]
 \label{def:membership}
A sequence $x$ is a \emph{member} of a dataset $\smash\D = \{x^{(i)}\}_{i=1}^{N}$ if $x$ shares \emph{at least} one $n$-gram with any $x^{(i)} \in \D$. That is, $x$ is \emph{member} if there exists a $g \in {\operatorname{n-grams}}(x)$ s.t. $\smash{g \in \bigcup_{i} {\operatorname{n-grams}}(x^{(i)})}$.
\end{definition}
\vspace{-2mm}

This definition is stringent (e.g., approximate membership typically requires many, not just one, \ngram to match). This ensures we overestimate members and thus underestimate non-members.
This definition is also inclusive of the those in the literature, as varying $n$ captures a spectrum of them. For example, setting $n=|x|$ is the verbatim membership used in~\citet{carlini2022quantifying}. Smaller $n$ captures many approximate membership definitions, such as MinHash~\citep{broder1997resemblanceminhash,lee2021deduplicating}, edit distance based membership~\citep{ippolito2022preventing}, and many other \ngram variants cited in Section~\ref{sec:background}.
In the remainder of the paper, we call a sequence $x$ an ``$n$-gram member'' if $x$ satisfies \cref{def:membership}, and otherwise a ``$n$-gram non-member''.

\textbf{Definition of \textit{Data Completion}.}
Informally, we define a \textit{completion} as: when a token sequence is known \emph{a priori} and a language model generates its suffix when prompted with its prefix.
Formally, if $x = [\prompt \| \suffix]$, then model generates the expected suffix $\suffix$ of $x$ based on the provided prefix $\prompt$ (prompt).
For simplicity, we focus on $|p| = |s| = |x| / 2$ in our experiments. Prior work has studied how the choice in prefix and suffix lengths impact memorization~\citep{carlini2022quantifying}.
To capture highly similar but not verbatim completions of the sequence $x$, we introduce variants of completion that allow for semantically insignificant deviations from the original $\suffix$. 
We define the following notions of completion.
\begin{definition}[Exact completion]
\label{def:exact-completion}
Given tokens $x = [\prompt \| \suffix]$ and a model $\M$, we say $x$ is \emph{exactly completed} if $\M(\prompt)=\suffix$ using greedy decoding.
\end{definition}
\vspace{-2mm}
This is closely related to verbatim memorization and verbatim training data extraction (see Section~\ref{sec:background}).
There are also semantically equivalent sequences that humans would be likely to not distinguish from the original sequence. We thus consider two approximate notions of completion, relevant to our experiments in Section~\ref{sec:fine-tuning}.
\begin{definition}[$r$-similar completion]
\label{def:editsim-completion}
Given $x = [\prompt \| \suffix]$ and $\M$, we say $x$ is a \emph{$r$-similar completion} if $\M(\prompt)$ is within a normalized Levenshtein edit distance of $1 - r$ using greedy decoding, i.e., $\operatorname{lev}(\M(\prompt), \suffix) / \max(|\M(\prompt)|, |\suffix|) \le 1 - r$. 
\end{definition}
\vspace{-1mm}
\begin{definition}[Case-insensitive completion]
\label{def:case-completion}
Given $x = [\prompt \| \suffix]$ and $\M$, we say $x$ is a \emph{case-insensitive completion} if $\operatorname{lower}(\M(\prompt)) = \operatorname{lower}(\suffix)$ with greedy decoding, where $\operatorname{lower}(\cdot)$ applies character-wise lower casing. 
\end{definition}
\vspace{-2mm}

\begin{figure}[h]
    \centering
    \vspace{-3pt}
    \includegraphics[width=0.7\linewidth]{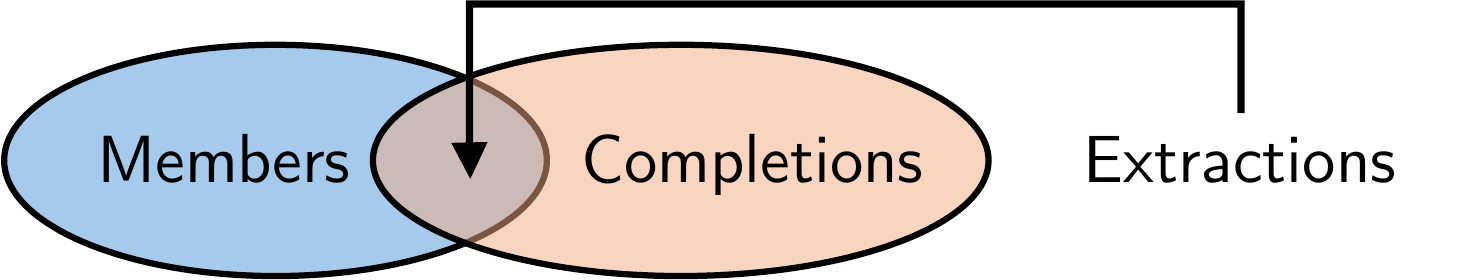}
    \vspace{-5pt}
    \caption{We say that a sequence is \textit{extractable} if it can be \textit{completed} and can be proved a \textit{member} of the training set.}
    \label{fig:venn}
    \vspace{-15pt}
\end{figure}

\textbf{Data Completion vs.\ Data Extraction.}
Data \textit{extraction}, as considered in recent work~\cite{carlini2021extracting,nasr2023scalable}, concerns recovering \textit{training} data from the model. This thus involves both (1) data \textit{completion} (e.g., as in \cref{def:exact-completion}), and (2) verifying that the completion is a training \textit{member}; e.g., by inspecting the training data.
In a sense, extraction specifically measures \textit{memorization},\footnote{Our reference to ``memorization'' is intended in a very specific context: whether a model can be induced to generate near-copies of some training examples when prompted with appropriate instructions, and often with prior knowledge of the model's training data.  Specifically, we do not mean to imply that a model ``contains'' its training data in the sense that any arbitrary instance of that data can be retrieved without use of specialized software or algorithms. Rather, if a model can be induced to generate very close copies of certain training examples by supplying appropriate instructions to guide the model's statistical generation process then that model is said to have ``memorized'' those examples.} while completion is more generic---indeed, our work studies \textit{non-member} completions; \cref{fig:venn} illustrates the distinction. 

\section{Removing Members Does Not Always Prevent LLM Verbatim Completion}
\label{sec:pre-training}

LLMs are known to complete training sequences~\cite{carlini2021extracting}. In this section, we ask: 

\begin{mdframed}
\textbf{Main Question}: Will an LLM still complete a text sequence even after we \emph{remove} all training sequences that have \ngram overlap with it?
\end{mdframed}
\vspace{-5pt}
We find that this retrained model still has \emph{lingering completions}: the completion test still succeeds despite having removed all completions identified by \ngram matching from the training data. This begs the question: why is the LLM still completing these sequences? As we will see, this is because the completions are either a) still contained in the dataset but via a different membership definition (i.e., for some $m$-gram membership, with $m<n$), or b) they lack sufficient entropy and can be easily predicted (generalized).

\subsection{Experimental Setup}
\label{sec:pretraining-setup}
Our high-level experimental framework is as follows. We defer additional details to \cref{sec:supp-pretraining}.
\vspace{-1mm}
\begin{enumerate}
    \item \textbf{Pre-train a base model}: We first pre-train a standard LLM $\Mbase$ from scratch on a training dataset $\Dbase$.
    \item \textbf{Identify verbatim completions}: We then collect a set of sequences $\Dmem$ of length $k$ that $\Mbase$ can complete \textit{verbatim} (as in \cref{def:exact-completion}), by checking the first $k$ tokens of \textit{every} training document in $\Dbase$. This is a simple and effective procedure since LLMs are known to memorize training data (e.g., \citet{carlini2022quantifying}); other choices to obtain $\Dmem$ are also possible.
    \item \textbf{$n$-gram filtering}:
    We then filter each sequence $x \in \Dmem$ away from $\Dbase$. Our filtering procedure is simple and reflects $n$-gram membership (\cref{def:membership}): to filter a sequence $x$, we perform a sliding window of length $|x|$ over $\Dbase$ (as if all tokens are concatenated into a single array); if the window shares any $n$-gram with $\ngramsfunc(x)$, the window is deleted from the pre-training data.
    When $n = |x|$, then we filter $x$ exactly as it appears in the dataset; when $n < |x|$, we filter more aggressively as the window is removed on partial matches against $x$. 
    The filtered dataset is denoted as $\smash{\Dfilter{n}}$.
    \item \textbf{Re-train a counterfactual model}: Pre-train another LLM $\smash{\Mfilter{n}}$ \textit{from scratch} on the filtered data $\smash{\Dfilter{n}}$. 
\end{enumerate}
We repeat this procedure for different model sizes, different values of $n$-gram (for filtering), and different sequence lengths $k$. Unless otherwise stated, we use $k=50$, meaning that for a sequence $x=[p\|s]$, we have $|p| = |s| = 25$. We provide results on $k=100$ in \cref{sec:supp-pretrain-longer-seq}.

\textbf{Models and training.} 
We pre-train a series of models from scratch using the GPT-2 architecture~\cite{radford2019language}, spanning sizes of \{350M, 774M, 1.6B, 2.8B\} parameters, with 1.6B being the size of the original GPT-2 XL and 2.8B being a scaled-up model. We use LLM.c~\cite{karpathy2024llmc} for an efficient pre-training pipeline. We primarily report results on the 1.6B model unless otherwise stated.

\textbf{Data.}
For all models, we use FineWeb-Edu~\cite{penedo2024fineweb} as a state-of-the-art pre-training dataset.\footnote{This work may contain information from FineWeb-Edu dataset, which is made available under the ODC Attribution License.} We use the same base dataset $\Dbase$ of 33.6B randomly sampled tokens. For the 1.6B model, 33.6B tokens is approximately Chinchilla optimal ($\approx 20$ tokens per parameter, \citet{hoffmann2022training}). For consistency, we train the base models $\Mbase$ of different sizes with the same starting dataset $\Dbase$; as the size of $\Dmem$ (step \#2) hinges on the size of $\Mbase$~\cite{carlini2022quantifying}, we obtain different filtered datasets $\smash{\Dfilter{n}}$ for each model size (by inferencing on $\Mbase$).

\begin{figure}[t]
\centering
\begin{minipage}{\linewidth}
    \centering
    \captionof{table}{The number of identified verbatim memorized sequences $|\Dmem|$ at different model sizes (step \#2 of \S~\ref{sec:pretraining-setup}).}
    \vspace{-3mm}
    \label{tab:dmem-size}
    {\footnotesize
    \begin{tabular}{@{}lccccc@{}}
    \toprule
    \textbf{Model size}       & \textbf{304M} & \textbf{774M}   & \textbf{1.6B}   & \textbf{2.8B}   \\ \midrule
    $|\Dmem|$           & 76,648   & 116,270   &  151,598   & 175,813    \\
    \bottomrule
    \end{tabular}
    }
\end{minipage}
\begin{minipage}{\linewidth}
    \centering
    \includegraphics[width=0.8\linewidth]{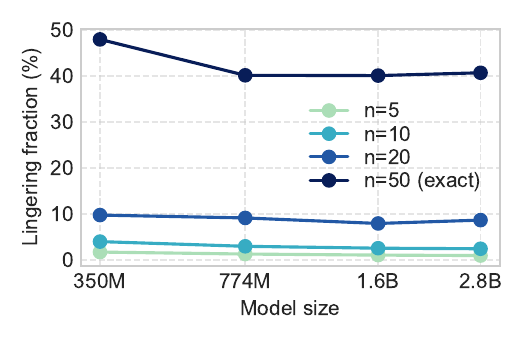}
    \vspace{-5mm}
    \caption{\textbf{LLMs can verbatim complete texts with zero $n$-gram overlap to training data.} A fraction of sequences filtered away from pre-training data (\Dmem, \cref{tab:dmem-size}) can still be completed by the re-trained LLM verbatim. The fractions decrease under stronger filtering (smaller $n$-gram filter) and remain relatively stable across model scales.
    }
    \vspace{-5mm}
    \label{fig:pretrain-scale}
\end{minipage}
\end{figure}

\subsection{Results}
\label{sec:pretraining-results}

With the artifacts \smash{\Mbase, \Dbase, \Dmem, \Dfilter{n}, \Mfilter{n}}, we now make observations pertaining to our main question.

\textbf{Finding \#1 (Existence of Lingering Sequences): LLMs can \textit{verbatim} complete a fraction of the sequences deleted from training data, and {consistently} so across scale.} 
On a macroscopic level, we first observe that simply deleting a set of sequences from pre-training data does \textit{not} always prevent them from being generated by an LLM (\cref{fig:pretrain-scale}). 
This observation is consistent across model scales, where each size has a different amount of memorization (\cref{tab:dmem-size}).
We call these ``\textit{lingering sequences}'' and denote them as $\Dlinger{n}$. Under our experimental conditions, the fraction of lingering sequences $| \Dlinger{n} | / |\Dmem |$ can be as high as 40\% when we apply the weakest \ngram filter and only remove verbatim sequence matches ($n=50$).

\begin{figure}[t]
    \centering
    \includegraphics[width=\linewidth]{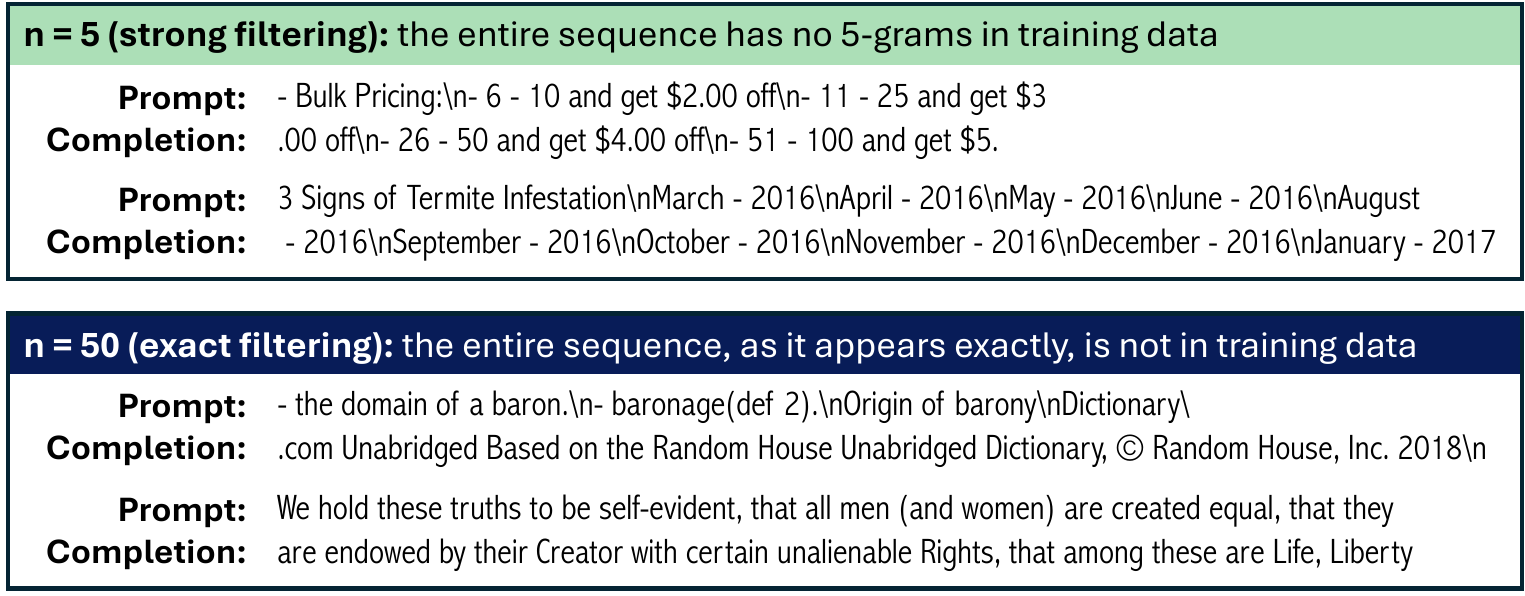}
    \vspace{-7mm}
    \caption{Examples of lingering sequences (more in \ref{sec:supp-vis-lingering-seqs}).}
    \vspace{-1em}
    \label{fig:vis-lingering}
\end{figure}

\textbf{Finding \#2 (Nature of Lingering Sequences): We found no lingering sequences that correspond to \textit{creative} generalization---sequences beyond reconstructions from neighboring texts and continuations of recognizable patterns.}
\cref{fig:vis-lingering} visualizes a few lingering sequences in $\Dlinger{n}$ and see \cref{sec:supp-vis-lingering-seqs} for more. 
To understand their origin, we then perform a search of neighboring texts (Levenshtein distance $< 20$) for a few randomly\footnote{We only perform this experiment on randomly selected lingering sequences due to the cost of the search.} selected lingering sequences over the large pre-training data $\Dbase$; we defer results to \cref{sec:supp-vis-lingering-neighbors}.
For all lingering sequences we queried, we were able to find near-duplicates, yet all such copies evaded $n$-gram overlap detection one way or another. 
This result sheds light on the remarkable ability for LLMs to generalize from neighboring text. More interestingly, it also informs an interesting symmetry on how we may \textit{adversarially} construct training sequences that: (1) have no $n$-gram overlap with a target sequence $x$, and (2) yet serve as ``anchor points'' that the LLM can interpolate to verbatim complete $x$. We explore this symmetry in the coming section (\S\ref{sec:fine-tuning}).

\textbf{Finding \#3 (Persistence of Lingering Sequences): Stronger filters reduce, but do not eliminate, these lingering sequences, and instead shift their distribution to more generalizable patterns.}
As we filter $\Dmem$ from $\Dbase$ more aggressively with smaller $n$-gram filters, smaller fractions of $\Dmem$ can be completed by the re-trained model $\Mfilter{n}$ verbatim (\cref{fig:pretrain-scale}). However, even at a very conservative filter of $n=5$ (a sequence is removed from $\Dbase$ if any 5-gram is in $\Dmem$), $\Dlinger{5}$ still accounts for $\approx 1\%$ of $\Dmem$ (\cref{tab:dmem-size}). 
As the fraction of lingering sequences decreases, their contents also shift from verbatim memorization of semantically useful text (e.g., famous quotes) to generalizable patterns (e.g., counting in Roman numerals). We provide examples in \cref{fig:vis-lingering} and \cref{sec:supp-vis-lingering-seqs}.

\begin{figure}[t]
    \centering
    \vspace{-.5mm}
    \includegraphics[width=0.8\linewidth]{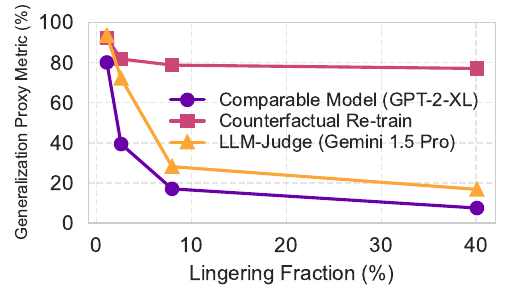}
    \vspace{-4mm}
    \caption{
    \textbf{Strong filters result in less lingering completions and shift them toward generalizable patterns.} Results on 1.6B size. We use three {{proxy} metrics} where higher indicates more pattern continuations: 1) \% verbatim completion by off-the-shelf GPT-2-XL; 2) \% verbatim completion by \Mcounterfactual, a counterfactual model trained on disjoint pretraining shards; 3) \% judged as patterns by Gemini 1.5 Pro.
    }
    \vspace{-4mm}
    \label{fig:pretrain-proxy}
\end{figure}

To quantify this shift, we use three proxy metrics (\cref{fig:pretrain-proxy}), though we note that none perfectly captures the (fuzzy) boundary between memorization and generalization. 
First, we measure the verbatim completion rate of $\Dlinger{n}$ using the off-the-shelf GPT-2-XL~\cite{radford2019language}; since $\Mfilter{n}$ is a similar model by construction, a lingering sequence is likely a generalizable pattern if both models (trained on distinct data) agree on its completion. 
Second, we similarly consider the completion rate of a counterfactual model \Mcounterfactual~on pre-training shards disjoint from $\Dbase$.
Third, we prompt Gemini 1.5 Pro with few-shot examples to determine if a lingering sequence is a pattern continuation (prompt template in \cref{sec:supp-prompt-template}).
All proxy metrics confirm our manual inspection that stronger filters force out generalization behaviors from the model, albeit these are simple sequences to generalize to (recall finding \#2).


\subsection{Interpretations and Outlook}
\label{sec:pretraining-discussions}

While we found no evidence of creative generalization 
in models up to 2.8B parameters, 
lingering sequences are intriguing because they seem to challenge our understanding of membership in LLMs---if a language model can verbatim complete sequences known \textit{a priori} without ever training on any of its $n$-grams, 
what does this imply for the definition of membership and its reliance on $n$-gram overlap?
To better understand these limitations we identified---and if they are exclusively explained by the limitation of $n$-gram overlaps---we next experiment with adversarially gaming $n$-gram membership. We build on our findings thus far to inform the adversarial construction of a dataset of \ngram non-members that is able to force LLM completion.

\section{Adding Non-Members Can Force LLM Verbatim Completion}
\label{sec:fine-tuning}

While lingering sequences (Section~\ref{sec:pre-training}) are largely benign and rare, observing how they came to be (e.g., through visualizations in \cref{sec:supp-vis-lingering-neighbors}) informs how one may \textit{adversarially} force a model to complete $n$-gram non-members. 
We explore this direction with the following:
\begin{mdframed}
\textbf{Main Question}: Given a chosen (unseen) text sequence $x$, can we {add} training sequences $\Dfinetune$ that have no $n$-gram overlap with $x$, and yet an LLM fine-tuned on $\Dfinetune$ can complete $x$ verbatim? 
\end{mdframed}
\vspace{-1mm}
This question is intriguing for its insights into LLM generalization, as well as its relevance to \textit{adversarial} manipulation of training data, where an adversary may wish to intentionally avoid $n$-gram membership for, e.g., harder-to-detect data poisons and content misuse (more in Section~\ref{sec:finetuning-discussions}).

To answer the question, consider a class of \textit{noisy transformations} $f$ on $x$, such that: (1) $\tilde{x} = f(x)$ retains some information about $x$; and (2) $\tilde{x}$ has no $n$-gram overlap with $x$. We hypothesize that an LLM trained on different instances of $\tilde{x}$ (over different randomness) should learn to recover the original $x$, much like a denoising antoencoder learning to recover clean data from noisy inputs.

\textbf{Methods.}
We show that it is possible to {adversarially} construct examples $\Dfinetune = \{\tilde{x}_i\}$ from a chosen \textit{unseen} example $x$ (e.g., today's news) such that $x$ and $\Dfinetune$ share no common $n$-grams (and many membership tests, including manual inspection, would fail). Yet with only a few gradient steps of fine-tuning an LLM on $\Dfinetune$, the model can (verbatim) complete this ``non-member'' $x$.
We study three such methods of constructing such a fine-tuning set $\Dfinetune$:
\begin{enumerate}
    \item \textit{Stitching chunks}: We split $x$ into overlapping segments padded with random tokens.
    \item \textit{Token dropouts}: We replace tokens in $x$ at different (random) positions with random tokens; positions have at most $n-1$ gap to avoid $n$-gram overlaps.
    \item \textit{Casing flips}: We flip the casing of every English letter in $x$ with probability $p$. 
\end{enumerate}
These methods resemble real-world transformations of texts such as taking excerpts of an article and transcriptions that misspell words, miss punctuation, and drop casing. 
They also have varying degrees of efficacy as we will discuss in Section~\ref{sec:finetuning-results}.
Note that our goal is not to find the best possible (stealthiest) transformation, but to explore feasibility and ease of such adversarial manipulation.

\textbf{Models and training.}
We work with two model families: Gemma-2~\cite{team2024gemma,team2024gemma2} and Qwen-2.5~\cite{yang2024qwen2,qwen2.5}, spanning model size from 0.5B to 9B. We fine-tuned these models to predict the next token with a batch size of 32 and a constant learning rate of $10^{-5}$. 

\textbf{Data.}
We primarily experiment on three target texts. These texts are all roughly 1,000 characters long ($\approx$ 250 tokens under Gemma-2 tokenizer) and have a recent temporal cutoff such that they could not have been included in the training set of Gemma-2 and are extremely unlikely to appear in the training set of Qwen-2.5 (thus helps ablate the effect of potential memorization):
\vspace{-1mm}
\begin{enumerate}
    \item \textbf{Lyles (NYT article)}: an excerpt of a recent New York Times article about Noah Lyles and the Olympics; 
    \item \textbf{Karpathy (tweet)}: a tweet text in an image posted by Andrej Karpathy about LLM tokenization; and
    \item \textbf{Willow (blog)}: an excerpt from the recent Google blog post on Willow, the quantum computing chip. 
\end{enumerate}
\vspace{-1mm}
The text choices are otherwise arbitrary and alternatives are possible; see the full texts, source, and results on alternative texts in \cref{sec:supp-target-text}.
For every target text sequence, we construct $N=2,000$ examples as $\Dfinetune$ by applying the transformation $f$ with different randomness, though in most settings we need less than 1,000 examples. We visualize some of these examples in \cref{sec:supp-visualizations-finetuning}.

\begin{figure*}
    \centering
    \vspace{-2mm}
    \includegraphics[width=\linewidth]{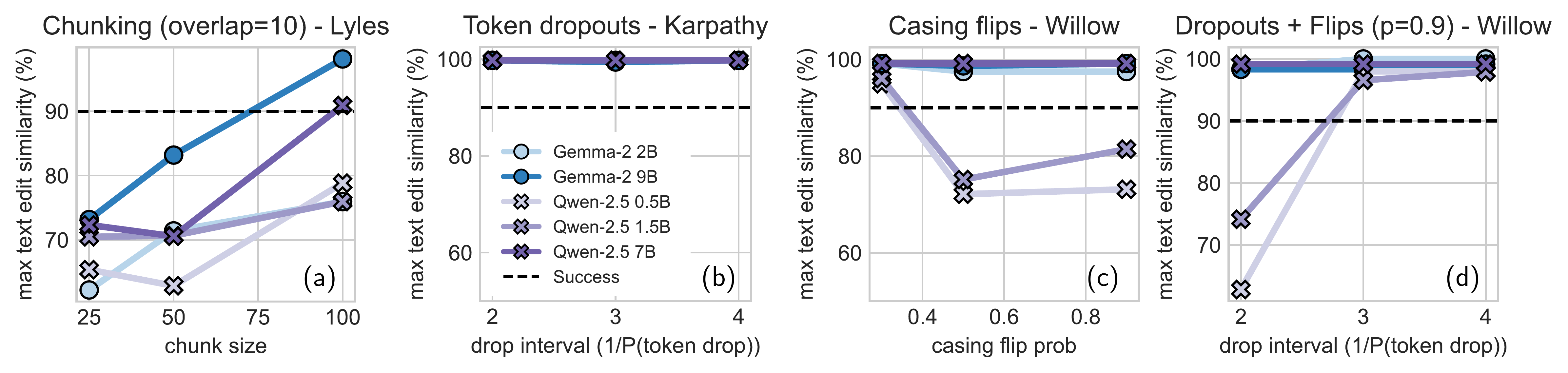}
    \vspace{-8mm}
    \caption{\textbf{Completion success across methods and target texts:} (a) chunking on \textbf{Lyles}, over chunk size $c$ ($x$-axis); (b) token dropouts on \textbf{Karpathy}, over drop interval $d$; (c) casing flips on \textbf{Willow}, over flip prob $p$ ($p=0.5$ is noisiest); and (d) combining dropouts + flips on \textbf{Willow}. 
    We observe that: (1) it is possible  to complete a chosen string with zero $n$-gram membership, and (2) this ability improves with model size.
    See \cref{sec:supp-finetuning-results} for comprehensive results.
    }
    \vspace{-5mm}
    \label{fig:finetune-main-results}
\end{figure*}

\subsection{Stitching chunks}
\label{sec:chunking}
A natural way to avoid generating any $\tilde{x}$ that shares $n$-grams of $x$ is to only expose chunks of at most $(n-1)$-grams to the model.
We construct $\Dfinetune$ by breaking $x$ into contiguous, overlapping token segments, with the remaining positions padded with random tokens from the vocabulary.
This procedure is parameterized by the chunk size $c$ and the overlap $l$.
Chunk size $c$ controls the difficulty (noise level) of the task. For example, the task is trivial if $c$ is the sequence length (since $\Dfinetune$ are just copies of $x$). A small $c$ means most tokens of any $\tilde x \in \Dfinetune$ are random. A large overlap $l$ should intuitively help the LLM learn to stitch $x$ together, though empirically we observe minimal impact (\cref{sec:supp-chunking}). 

To illustrate, if the target $x = [1, 2, 3, 4, 5, 6]$, then we may have $\Dfinetune = \{[1, 2, 3, \cdot, \cdot, \cdot], [\cdot, \cdot, 3, 4, 5, \cdot], [\cdot, \cdot, \cdot, \cdot, 5, 6], ...\}$ with chunk size $c = 3$ and overlap $l = 1$, ($\cdot$ denotes a fresh random token). 
Intuitively, the task for the LLM is to ``stitch'' the token chunks back together into $x$.
A similar technique is explored in the concurrent work of \citet{panaitescu2024poisonedparrot} for data poisoning; here, we explore chunk sizes, overlaps, and model families to present more comprehensive tradeoffs. 
See \cref{alg:chunking} for our detailed procedure and \cref{sec:supp-visualizations-finetuning} to visualize examples in $\Dfinetune$.

\subsection{Token dropouts} 
\label{sec:goldfish}

Another way to avoid $\tilde{x} \in \Dfinetune$ sharing $n$-grams with $x$ is to mask out tokens in $\tilde{x}$ at least every $n$ positions, so that it cannot share \ngram overlap with $x$. That is, $\Dfinetune$ contains different versions of $x$ where at least every $(n-1)$-th token is masked out (replaced with a random token) so that there is at most $n-1$ token overlap. This procedure is parameterized by a drop interval $d$, representing the length of the interval between dropped tokens. It must be that $d\le n$ to ensure no \ngram overlap.
To illustrate, if $x = [1,2,3,4,5,6]$, then we may have $\Dfinetune = \{[1,2,3,\cdot,5,6], [1,\cdot,3,4,5,\cdot], [\cdot,2,3,4,\cdot,6], ...\}$, where `$\cdot$' is a fresh random token and where here $d = 4$. We also consider a \textit{randomized} dropout, where every token is dropped with probability $1/d$. By construction, the deterministic version guarantees that $x$ is not a $d$-gram member of the training set, while the randomized version does so with (exponentially) high probability.
Notably, due to BPE tokenzation~\cite{sennrich2015neural}, the original text becomes visually obfuscated for humans even when most tokens are retained ($d > 2$). See \cref{alg:goldfish} for detailed algorithm and \cref{sec:supp-visualizations-finetuning} for visualization.%

This construction is closely related to the \textit{goldfish loss} proposed by \citet{hans2024like} to mitigate verbatim memorization, where the loss of every $n$-th token (on average, if randomized) is omitted during training.
However, the procedure here is entirely \textit{data-centric}: it does not interfere with the training objective and makes the learning task harder as subsequent tokens would still attend to the random tokens.

\subsection{Case flipping} 
\label{sec:casing}
Another approach is to perform text-space transformations that preserve semantics yet drastically alter the token-space representations. 
One such method is to randomly flip the \textit{casing} of English letters, creating varied \textit{tokenization} of the otherwise equivalent string. 
To illustrate, if $x$ decodes to \verb|'This is a string'|, $\Dfinetune$ may include token sequences of strings like \verb|'THIS Is A stRinG'|. 
Due to the mechanisms of BPE tokenization, it is extremely easy to obtain $\tilde x$ with \textit{completely distinct} tokens than $x$ under modern LLM tokenizers (see \cref{sec:supp-visualizations-token-transformations} for visualization).
The case flipping procedure is parameterized by the flip probability $p$. $p=0.5$ creates the highest variance, and $p$ closer to 1 always flips the case (most letters being initially lower case, natural text becomes mostly upper case).

\begin{figure}[t]
    \centering
    \includegraphics[width=\linewidth]{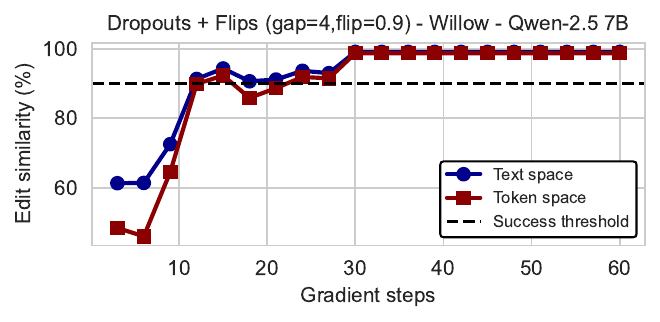}
    \vspace{-9mm}
    \caption{\textbf{Completion success may only require a few gradient steps.}
    See more configurations in \cref{sec:supp-finetuning-results}.
    }
    \vspace{-2em}
    \label{fig:finetune-steps}
\end{figure}

\subsection{Compositions} 
\label{sec:goldfish-casing}

The transformations presented earlier are not mutually exclusive, and in principle they can be composed arbitrarily.
We explore one such composition of token dropouts \& case flipping, where non-dropped tokens (\S\ref{sec:goldfish}) have casing randomly flipped in the text space (\S\ref{sec:casing}).
A key benefit of compositions is that they combinatorially give rise to many new transformations with potentially more granular control of task difficulty and detectability (e.g., whether manually inspecting $\Dfinetune$ reveals $x$; see \cref{sec:supp-visualizations-finetuning} for example visualizations). We leave a comprehensive evaluation of composed transformations to future work.

\subsection{Results}
\label{sec:finetuning-results}

\textbf{Finding \#1: It is possible for an LLM to complete an \textit{unseen} string with no $n$-gram membership after minimal finetuning.}

\cref{fig:finetune-main-results} shows the results of various transformations across various text targets (see \cref{sec:supp-finetuning-results} for comprehensive results).
We first see that there are many configurations where fine-tuning on $\Dfinetune$ allow the successful completion of target $x$ (edit similarity $> 90\%$).

Some transformations are more effective than others.
We found that \textit{chunking} (Section~\ref{sec:chunking}) was ineffective: with a small chunk size ($c=25$), the models mostly fail to complete the target, and only some models succeed at $c = 100$ (\cref{fig:finetune-main-results}~(a)). 
\textit{Token dropouts} (Section~\ref{sec:goldfish}), on the other hand, is {extremely} effective---even the smallest model (Qwen-2.5 0.5B) easily completes the target verbatim at a drop interval of 2 (50\% drop probability at every token; \cref{fig:finetune-main-results}~(b)).
Our results thus present a counter-case to \textit{goldfish loss}~\cite{hans2024like}, as models can still complete targets verbatim when there are multiple versions of the target with \textit{different} token dropout positions (e.g., due to near-duplicates, related to the findings of Section~\ref{sec:pre-training}).
\textit{Casing flips} (Section~\ref{sec:casing}) are also generally effective (\cref{fig:finetune-main-results}~(c)). Composing token dropouts and casing flips (Section~\ref{sec:goldfish-casing}) increases task difficulty (fewer successes with small models) but otherwise similarly enables verbatim completion.

In essence, these experiments demonstrate that $n$-gram based membership definitions can be vulnerable to \textit{adversarial} manipulation: the fine-tuning set $\Dfinetune$ clearly contains information about the text $x$, but when given $x$ and a choice of $n$, it is easy for an adversary to bypass detection and yet have the model generate $x$ verbatim.

\textbf{Finding \#2: Completion success scales with model size.}
Another message from \cref{fig:finetune-main-results} is that as we increase in the model size, the completion success generally improves under the \textit{same} configurations. This provides evidence that frontier models should be more capable at synthesizing $n$-gram non-members into the target texts.

\subsection{Interpretations and Outlook} 
\label{sec:finetuning-discussions}

We briefly describe the potential implications of our adversarially constructed fine-tuning datasets:
\begin{itemize}[leftmargin=*]
    \item \textbf{Data poisoning}: $n$-gram non-members of a poison text $x$ can be added to the training set and still induce the generation of $x$. The concurrent work of~\citet{panaitescu2024poisonedparrot} explores adding copyrighted materials as poisoned data with a similar chunking technique (\S\ref{sec:chunking}).
    \item \textbf{Data contamination}: a dishonest model developer may game model evaluations through deliberate data contamination while evading $n$-gram based detection.
    \item \textbf{Reporting train-test overlap metrics}: 
    More broadly, a model developer may self-report train-test overlap statistics (e.g., as part of contamination analysis seen in~\citet{llama3,team2023gemini,brown2020language}). 
    Our results highlight that it is desirable that developers report additional metrics beyond $n$-gram overlap.
\end{itemize}

\section{Concluding Remarks} \label{sec:conclusions}

Lingering sequences (\S~\ref{sec:pre-training}) and adversarially constructed fine-tuning datasets (\S~\ref{sec:fine-tuning}) demonstrate the remarkable ability of LLMs to generalize from neighboring text. 
They are thus a valuable tool for evaluating LLM capabilities as models and pre-training datasets scale up. 
We conclude our work with a discussion of the implications of our findings: %

\textbf{Membership definitions and tests should incorporate new similarity measures}. We showed \ngram based membership emits false negatives that may not capture human intuition nor the pragmatic concerns of the copyright, privacy, and AI safety community. On the flip side, tests like membership inference should consider broader notions of membership beyond individual sequences: a unit of data could be a collection of sequences grouped by similarity~\cite{kandpal2023user,maini2024llm,cooper2024machine}.

\textbf{Machine unlearning alone is insufficient to address data permissibility concerns in output suppression.}
It is widely accepted that a golden baseline of machine unlearning is to retrain a model from scratch without the target forget data~\citep{bourtoule2021machine,liu2024unlearning,liu2024rethinking,cooper2024machine}.
Yet, our experiments perform precisely this counterfactual and reveal that some excluded sequences can still be verbatim generated %
(\cref{fig:teaser}, \S\ref{sec:pre-training}).
We thus caution that unlearning alone may not always prevent a model from generating a sequence of interest (e.g., a harmful sequence). This is also known as output suppression and is  a common goal of unlearning~\cite{cooper2024machine}.

\textbf{Exploring the connection between our work and forging may help more precisely characterize threat models for when completion can serve as evidence of membership.}
Readers familiar with the forging~\citep{thudi2022necessity} literature will have noticed a connection with our work. 
Forging a step of gradient descent computed on a given minibatch is done by (adversarially) constructing a different minibatch that will result in the same gradient being computed.
Perhaps surprisingly at first, gradients can be forged using \textit{non-overlapping} datasets sampled from ``natural'' distributions.
Rather than forging gradients, our methods can be viewed as attempting to forge model outputs.
While our work begins to show that there may exist threat models where completion is insufficient evidence for \ngram membership, our experiments did not succeed in obtaining a forge in model outputs using ``natural'' data (recall \S~\ref{sec:pre-training}).
We believe this is a valuable direction for future work.

\section*{Impact Statement}
Our work studies training data membership in the context of language models. 
The impact of our work is described in Sections~\ref{sec:pretraining-discussions},~\ref{sec:finetuning-discussions}, and~\ref{sec:conclusions}.
As we described earlier, the limitations we identified in \ngram based definitions of membership have implications for copyright, privacy, and AI safety. 
Broadly speaking, our work advocates for additional membership definitions to help better capture human intuition and make membership more pragmatic. 

\section*{Acknowledgements}

KZL would like to thank Xinran Zhao, Nikil Selvam, Steven Cao, Jing Huang, Yangjun Ruan, Aryaman Arora, Harshit Joshi, Rylan Schaeffer, Chenglei Si, Yanzhe Zhang, as well as members of the p-lambda lab and STAIR lab at Stanford University for helpful discussions and feedback. We thank Milad Nasr and Nicholas Carlini for helpful discussions, comments, and critiques throughout the work.

\nocite{langley00}

\bibliography{main}

\begin{thebibliography}{65}
\providecommand{\natexlab}[1]{#1}
\providecommand{\url}[1]{\texttt{#1}}
\expandafter\ifx\csname urlstyle\endcsname\relax
  \providecommand{\doi}[1]{doi: #1}\else
  \providecommand{\doi}{doi: \begingroup \urlstyle{rm}\Url}\fi

\bibitem[Achiam et~al.(2023)Achiam, Adler, Agarwal, Ahmad, Akkaya, Aleman,
  Almeida, Altenschmidt, Altman, Anadkat, et~al.]{achiam2023gpt}
Achiam, J., Adler, S., Agarwal, S., Ahmad, L., Akkaya, I., Aleman, F.~L.,
  Almeida, D., Altenschmidt, J., Altman, S., Anadkat, S., et~al.
\newblock Gpt-4 technical report.
\newblock \emph{arXiv preprint arXiv:2303.08774}, 2023.

\bibitem[Anil et~al.(2023)Anil, Dai, Firat, Johnson, Lepikhin, Passos, Shakeri,
  Taropa, Bailey, Chen, et~al.]{anil2023palm}
Anil, R., Dai, A.~M., Firat, O., Johnson, M., Lepikhin, D., Passos, A.,
  Shakeri, S., Taropa, E., Bailey, P., Chen, Z., et~al.
\newblock Palm 2 technical report.
\newblock \emph{arXiv preprint arXiv:2305.10403}, 2023.

\bibitem[Biderman et~al.(2024)Biderman, Prashanth, Sutawika, Schoelkopf,
  Anthony, Purohit, and Raff]{biderman2024emergent}
Biderman, S., Prashanth, U., Sutawika, L., Schoelkopf, H., Anthony, Q.,
  Purohit, S., and Raff, E.
\newblock Emergent and predictable memorization in large language models.
\newblock \emph{Advances in Neural Information Processing Systems}, 36, 2024.

\bibitem[Bourtoule et~al.(2021)Bourtoule, Chandrasekaran, Choquette-Choo, Jia,
  Travers, Zhang, Lie, and Papernot]{bourtoule2021machine}
Bourtoule, L., Chandrasekaran, V., Choquette-Choo, C.~A., Jia, H., Travers, A.,
  Zhang, B., Lie, D., and Papernot, N.
\newblock Machine unlearning.
\newblock In \emph{2021 IEEE Symposium on Security and Privacy (SP)}, pp.\
  141--159. IEEE, 2021.

\bibitem[Broder(1997)]{broder1997resemblanceminhash}
Broder, A.~Z.
\newblock On the resemblance and containment of documents.
\newblock In \emph{Proceedings. Compression and Complexity of SEQUENCES 1997
  (Cat. No. 97TB100171)}, pp.\  21--29. IEEE, 1997.

\bibitem[Brown(2020)]{brown2020language}
Brown, T.~B.
\newblock Language models are few-shot learners.
\newblock \emph{arXiv preprint arXiv:2005.14165}, 2020.

\bibitem[Carlini et~al.(2019)Carlini, Liu, Erlingsson, Kos, and
  Song]{carlini2019secret}
Carlini, N., Liu, C., Erlingsson, {\'U}., Kos, J., and Song, D.
\newblock The secret sharer: Evaluating and testing unintended memorization in
  neural networks.
\newblock In \emph{28th USENIX security symposium (USENIX security 19)}, pp.\
  267--284, 2019.

\bibitem[Carlini et~al.(2021)Carlini, Tramer, Wallace, Jagielski, Herbert-Voss,
  Lee, Roberts, Brown, Song, Erlingsson, et~al.]{carlini2021extracting}
Carlini, N., Tramer, F., Wallace, E., Jagielski, M., Herbert-Voss, A., Lee, K.,
  Roberts, A., Brown, T., Song, D., Erlingsson, U., et~al.
\newblock Extracting training data from large language models.
\newblock In \emph{30th USENIX Security Symposium (USENIX Security 21)}, pp.\
  2633--2650, 2021.

\bibitem[Carlini et~al.(2022{\natexlab{a}})Carlini, Chien, Nasr, Song, Terzis,
  and Tramer]{carlini2022membership}
Carlini, N., Chien, S., Nasr, M., Song, S., Terzis, A., and Tramer, F.
\newblock Membership inference attacks from first principles.
\newblock In \emph{2022 IEEE Symposium on Security and Privacy (SP)}, pp.\
  1897--1914. IEEE, 2022{\natexlab{a}}.

\bibitem[Carlini et~al.(2022{\natexlab{b}})Carlini, Ippolito, Jagielski, Lee,
  Tramer, and Zhang]{carlini2022quantifying}
Carlini, N., Ippolito, D., Jagielski, M., Lee, K., Tramer, F., and Zhang, C.
\newblock Quantifying memorization across neural language models.
\newblock \emph{arXiv preprint arXiv:2202.07646}, 2022{\natexlab{b}}.

\bibitem[Carlini et~al.(2023)Carlini, Hayes, Nasr, Jagielski, Sehwag, Tramer,
  Balle, Ippolito, and Wallace]{carlini2023extracting}
Carlini, N., Hayes, J., Nasr, M., Jagielski, M., Sehwag, V., Tramer, F., Balle,
  B., Ippolito, D., and Wallace, E.
\newblock Extracting training data from diffusion models.
\newblock In \emph{32nd USENIX Security Symposium (USENIX Security 23)}, pp.\
  5253--5270, 2023.

\bibitem[Choquette-Choo et~al.(2021)Choquette-Choo, Tramer, Carlini, and
  Papernot]{choquette2021label}
Choquette-Choo, C.~A., Tramer, F., Carlini, N., and Papernot, N.
\newblock Label-only membership inference attacks.
\newblock In \emph{International conference on machine learning}, pp.\
  1964--1974. PMLR, 2021.

\bibitem[Cooper et~al.(2024)Cooper, Choquette-Choo, Bogen, Jagielski,
  Filippova, Liu, Chouldechova, Hayes, Huang, Mireshghallah,
  et~al.]{cooper2024machine}
Cooper, A.~F., Choquette-Choo, C.~A., Bogen, M., Jagielski, M., Filippova, K.,
  Liu, K.~Z., Chouldechova, A., Hayes, J., Huang, Y., Mireshghallah, N., et~al.
\newblock Machine unlearning doesn't do what you think: Lessons for generative
  ai policy, research, and practice.
\newblock \emph{arXiv preprint arXiv:2412.06966}, 2024.

\bibitem[Das et~al.(2024)Das, Zhang, and Tram{\`e}r]{das2024blind}
Das, D., Zhang, J., and Tram{\`e}r, F.
\newblock Blind baselines beat membership inference attacks for foundation
  models.
\newblock \emph{arXiv preprint arXiv:2406.16201}, 2024.

\bibitem[Dekoninck et~al.(2024)Dekoninck, M{\"u}ller, Baader, Fischer, and
  Vechev]{dekoninck2024evading}
Dekoninck, J., M{\"u}ller, M.~N., Baader, M., Fischer, M., and Vechev, M.
\newblock Evading data contamination detection for language models is (too)
  easy.
\newblock \emph{arXiv preprint arXiv:2402.02823}, 2024.

\bibitem[Duan et~al.(2024)Duan, Suri, Mireshghallah, Min, Shi, Zettlemoyer,
  Tsvetkov, Choi, Evans, and Hajishirzi]{duan2024membership}
Duan, M., Suri, A., Mireshghallah, N., Min, S., Shi, W., Zettlemoyer, L.,
  Tsvetkov, Y., Choi, Y., Evans, D., and Hajishirzi, H.
\newblock Do membership inference attacks work on large language models?
\newblock \emph{arXiv preprint arXiv:2402.07841}, 2024.

\bibitem[Dubey et~al.(2024)Dubey, Jauhri, Pandey, Kadian, Al-Dahle, Letman,
  Mathur, Schelten, Yang, Fan, et~al.]{llama3}
Dubey, A., Jauhri, A., Pandey, A., Kadian, A., Al-Dahle, A., Letman, A.,
  Mathur, A., Schelten, A., Yang, A., Fan, A., et~al.
\newblock The llama 3 herd of models.
\newblock \emph{arXiv preprint arXiv:2407.21783}, 2024.

\bibitem[Freeman et~al.(2024)Freeman, Rippe, Debenedetti, and
  Andriushchenko]{freeman2024exploring}
Freeman, J., Rippe, C., Debenedetti, E., and Andriushchenko, M.
\newblock Exploring memorization and copyright violation in frontier llms: A
  study of the new york times v. openai 2023 lawsuit.
\newblock \emph{arXiv preprint arXiv:2412.06370}, 2024.

\bibitem[{Gemini Team} et~al.(2023){Gemini Team}, Anil, Borgeaud, Alayrac, Yu,
  Soricut, Schalkwyk, Dai, Hauth, Millican, et~al.]{team2023gemini}
{Gemini Team}, Anil, R., Borgeaud, S., Alayrac, J.-B., Yu, J., Soricut, R.,
  Schalkwyk, J., Dai, A.~M., Hauth, A., Millican, K., et~al.
\newblock Gemini: a family of highly capable multimodal models.
\newblock \emph{arXiv preprint arXiv:2312.11805}, 2023.

\bibitem[{Gemini Team} et~al.(2024){Gemini Team}, Georgiev, Lei, Burnell, Bai,
  Gulati, Tanzer, Vincent, Pan, Wang, et~al.]{team2024gemini}
{Gemini Team}, Georgiev, P., Lei, V.~I., Burnell, R., Bai, L., Gulati, A.,
  Tanzer, G., Vincent, D., Pan, Z., Wang, S., et~al.
\newblock Gemini 1.5: Unlocking multimodal understanding across millions of
  tokens of context.
\newblock \emph{arXiv preprint arXiv:2403.05530}, 2024.

\bibitem[{Gemma Team} et~al.(2024{\natexlab{a}}){Gemma Team}, Mesnard, Hardin,
  Dadashi, Bhupatiraju, Pathak, Sifre, Rivi{\`e}re, Kale, Love,
  et~al.]{team2024gemma}
{Gemma Team}, Mesnard, T., Hardin, C., Dadashi, R., Bhupatiraju, S., Pathak,
  S., Sifre, L., Rivi{\`e}re, M., Kale, M.~S., Love, J., et~al.
\newblock Gemma: Open models based on gemini research and technology.
\newblock \emph{arXiv preprint arXiv:2403.08295}, 2024{\natexlab{a}}.

\bibitem[{Gemma Team} et~al.(2024{\natexlab{b}}){Gemma Team}, Riviere, Pathak,
  Sessa, Hardin, Bhupatiraju, Hussenot, Mesnard, Shahriari, Ram{\'e},
  et~al.]{team2024gemma2}
{Gemma Team}, Riviere, M., Pathak, S., Sessa, P.~G., Hardin, C., Bhupatiraju,
  S., Hussenot, L., Mesnard, T., Shahriari, B., Ram{\'e}, A., et~al.
\newblock Gemma 2: Improving open language models at a practical size.
\newblock \emph{arXiv preprint arXiv:2408.00118}, 2024{\natexlab{b}}.

\bibitem[Golchin \& Surdeanu(2023)Golchin and Surdeanu]{golchin2023data}
Golchin, S. and Surdeanu, M.
\newblock Data contamination quiz: A tool to detect and estimate contamination
  in large language models.
\newblock \emph{arXiv preprint arXiv:2311.06233}, 2023.

\bibitem[Hans et~al.(2024)Hans, Wen, Jain, Kirchenbauer, Kazemi, Singhania,
  Singh, Somepalli, Geiping, Bhatele, et~al.]{hans2024like}
Hans, A., Wen, Y., Jain, N., Kirchenbauer, J., Kazemi, H., Singhania, P.,
  Singh, S., Somepalli, G., Geiping, J., Bhatele, A., et~al.
\newblock Be like a goldfish, don't memorize! mitigating memorization in
  generative llms.
\newblock \emph{arXiv preprint arXiv:2406.10209}, 2024.

\bibitem[Hoffmann et~al.(2022)Hoffmann, Borgeaud, Mensch, Buchatskaya, Cai,
  Rutherford, Casas, Hendricks, Welbl, Clark, et~al.]{hoffmann2022training}
Hoffmann, J., Borgeaud, S., Mensch, A., Buchatskaya, E., Cai, T., Rutherford,
  E., Casas, D. d.~L., Hendricks, L.~A., Welbl, J., Clark, A., et~al.
\newblock Training compute-optimal large language models.
\newblock \emph{arXiv preprint arXiv:2203.15556}, 2022.

\bibitem[Ippolito et~al.(2022)Ippolito, Tram{\`e}r, Nasr, Zhang, Jagielski,
  Lee, Choquette-Choo, and Carlini]{ippolito2022preventing}
Ippolito, D., Tram{\`e}r, F., Nasr, M., Zhang, C., Jagielski, M., Lee, K.,
  Choquette-Choo, C.~A., and Carlini, N.
\newblock Preventing verbatim memorization in language models gives a false
  sense of privacy.
\newblock \emph{arXiv preprint arXiv:2210.17546}, 2022.

\bibitem[Jagielski et~al.(2024)Jagielski, Nasr, Lee, Choquette-Choo, Carlini,
  and Tramer]{jagielski2024students}
Jagielski, M., Nasr, M., Lee, K., Choquette-Choo, C.~A., Carlini, N., and
  Tramer, F.
\newblock Students parrot their teachers: Membership inference on model
  distillation.
\newblock \emph{Advances in Neural Information Processing Systems}, 36, 2024.

\bibitem[Jiang et~al.(2024)Jiang, Liu, Zhong, Schaeffer, Ouyang, Han, and
  Koyejo]{jiang2024investigating}
Jiang, M., Liu, K.~Z., Zhong, M., Schaeffer, R., Ouyang, S., Han, J., and
  Koyejo, S.
\newblock Investigating data contamination for pre-training language models.
\newblock \emph{arXiv preprint arXiv:2401.06059}, 2024.

\bibitem[Kandpal et~al.(2022)Kandpal, Wallace, and
  Raffel]{kandpal2022deduplicating}
Kandpal, N., Wallace, E., and Raffel, C.
\newblock Deduplicating training data mitigates privacy risks in language
  models.
\newblock In \emph{International Conference on Machine Learning}, pp.\
  10697--10707. PMLR, 2022.

\bibitem[Kandpal et~al.(2023)Kandpal, Pillutla, Oprea, Kairouz, Choquette-Choo,
  and Xu]{kandpal2023user}
Kandpal, N., Pillutla, K., Oprea, A., Kairouz, P., Choquette-Choo, C.~A., and
  Xu, Z.
\newblock User inference attacks on large language models.
\newblock \emph{arXiv preprint arXiv:2310.09266}, 2023.

\bibitem[Karpathy(2024)]{karpathy2024llmc}
Karpathy, A.
\newblock Llm.c: Llm training in simple, raw c/cuda.
\newblock \url{https://github.com/karpathy/llm.c}, 2024.
\newblock GitHub repository.

\bibitem[Kong et~al.(2023)Kong, Chowdhury, and Chaudhuri]{kong2023can}
Kong, Z., Chowdhury, A.~R., and Chaudhuri, K.
\newblock Can membership inferencing be refuted?
\newblock \emph{arXiv preprint arXiv:2303.03648}, 2023.

\bibitem[Kudugunta et~al.(2024)Kudugunta, Caswell, Zhang, Garcia, Xin,
  Kusupati, Stella, Bapna, and Firat]{kudugunta2024madlad}
Kudugunta, S., Caswell, I., Zhang, B., Garcia, X., Xin, D., Kusupati, A.,
  Stella, R., Bapna, A., and Firat, O.
\newblock Madlad-400: A multilingual and document-level large audited dataset.
\newblock \emph{Advances in Neural Information Processing Systems}, 36, 2024.

\bibitem[Langley(2000)]{langley00}
Langley, P.
\newblock Crafting papers on machine learning.
\newblock In Langley, P. (ed.), \emph{Proceedings of the 17th International
  Conference on Machine Learning (ICML 2000)}, pp.\  1207--1216, Stanford, CA,
  2000. Morgan Kaufmann.

\bibitem[Lee et~al.(2021)Lee, Ippolito, Nystrom, Zhang, Eck, Callison-Burch,
  and Carlini]{lee2021deduplicating}
Lee, K., Ippolito, D., Nystrom, A., Zhang, C., Eck, D., Callison-Burch, C., and
  Carlini, N.
\newblock Deduplicating training data makes language models better.
\newblock \emph{arXiv preprint arXiv:2107.06499}, 2021.

\bibitem[Li et~al.(2023)Li, Wang, Wang, and Neel]{li2023mope}
Li, M., Wang, J., Wang, J., and Neel, S.
\newblock Mope: Model perturbation-based privacy attacks on language models.
\newblock \emph{arXiv preprint arXiv:2310.14369}, 2023.

\bibitem[Liu(2024)]{liu2024unlearning}
Liu, K.~Z.
\newblock Machine unlearning in 2024, May 2024.
\newblock URL \url{https://ai.stanford.edu/~kzliu/blog/unlearning}.

\bibitem[Liu et~al.(2024)Liu, Yao, Jia, Casper, Baracaldo, Hase, Yao, Liu, Xu,
  Li, et~al.]{liu2024rethinking}
Liu, S., Yao, Y., Jia, J., Casper, S., Baracaldo, N., Hase, P., Yao, Y., Liu,
  C.~Y., Xu, X., Li, H., et~al.
\newblock Rethinking machine unlearning for large language models.
\newblock \emph{arXiv preprint arXiv:2402.08787}, 2024.

\bibitem[Maini et~al.(2021)Maini, Yaghini, and Papernot]{maini2021dataset}
Maini, P., Yaghini, M., and Papernot, N.
\newblock Dataset inference: Ownership resolution in machine learning.
\newblock \emph{arXiv preprint arXiv:2104.10706}, 2021.

\bibitem[Maini et~al.(2024)Maini, Jia, Papernot, and Dziedzic]{maini2024llm}
Maini, P., Jia, H., Papernot, N., and Dziedzic, A.
\newblock Llm dataset inference: Did you train on my dataset?
\newblock \emph{arXiv preprint arXiv:2406.06443}, 2024.

\bibitem[Mattern et~al.(2023)Mattern, Mireshghallah, Jin, Sch{\"o}lkopf,
  Sachan, and Berg-Kirkpatrick]{mattern2023membership}
Mattern, J., Mireshghallah, F., Jin, Z., Sch{\"o}lkopf, B., Sachan, M., and
  Berg-Kirkpatrick, T.
\newblock Membership inference attacks against language models via
  neighbourhood comparison.
\newblock \emph{arXiv preprint arXiv:2305.18462}, 2023.

\bibitem[Meeus et~al.(2024)Meeus, Jain, Rei, and
  de~Montjoye]{meeus2024inherent}
Meeus, M., Jain, S., Rei, M., and de~Montjoye, Y.-A.
\newblock Inherent challenges of post-hoc membership inference for large
  language models.
\newblock \emph{arXiv preprint arXiv:2406.17975}, 2024.

\bibitem[Mou(2023)]{mou2023dedup}
Mou, C.
\newblock Large-scale near-deduplication behind bigcode, 2023.
\newblock URL \url{https://huggingface.co/blog/dedup}.
\newblock Published on Hugging Face Blog, May 16, 2023.

\bibitem[Nasr et~al.(2023)Nasr, Carlini, Hayase, Jagielski, Cooper, Ippolito,
  Choquette-Choo, Wallace, Tram{\`e}r, and Lee]{nasr2023scalable}
Nasr, M., Carlini, N., Hayase, J., Jagielski, M., Cooper, A.~F., Ippolito, D.,
  Choquette-Choo, C.~A., Wallace, E., Tram{\`e}r, F., and Lee, K.
\newblock Scalable extraction of training data from (production) language
  models.
\newblock \emph{arXiv preprint arXiv:2311.17035}, 2023.

\bibitem[Oren et~al.(2023)Oren, Meister, Chatterji, Ladhak, and
  Hashimoto]{oren2023proving}
Oren, Y., Meister, N., Chatterji, N., Ladhak, F., and Hashimoto, T.~B.
\newblock Proving test set contamination in black box language models.
\newblock \emph{arXiv preprint arXiv:2310.17623}, 2023.

\bibitem[Panaitescu-Liess et~al.(2024)Panaitescu-Liess, Pathmanathan, Kaya,
  Che, An, Zhu, Agrawal, and Huang]{panaitescu2024poisonedparrot}
Panaitescu-Liess, M.-A., Pathmanathan, P., Kaya, Y., Che, Z., An, B., Zhu, S.,
  Agrawal, A., and Huang, F.
\newblock Poisonedparrot: Subtle data poisoning attacks to elicit
  copyright-infringing content from large language models.
\newblock In \emph{Neurips Safe Generative AI Workshop 2024}, 2024.
\newblock URL \url{https://openreview.net/forum?id=ZXgvPANlwe}.

\bibitem[Penedo et~al.(2024)Penedo, Kydl{\'\i}{\v{c}}ek, Lozhkov, Mitchell,
  Raffel, Von~Werra, Wolf, et~al.]{penedo2024fineweb}
Penedo, G., Kydl{\'\i}{\v{c}}ek, H., Lozhkov, A., Mitchell, M., Raffel, C.,
  Von~Werra, L., Wolf, T., et~al.
\newblock The fineweb datasets: Decanting the web for the finest text data at
  scale.
\newblock \emph{arXiv preprint arXiv:2406.17557}, 2024.

\bibitem[Radford et~al.(2019)Radford, Wu, Child, Luan, Amodei, Sutskever,
  et~al.]{radford2019language}
Radford, A., Wu, J., Child, R., Luan, D., Amodei, D., Sutskever, I., et~al.
\newblock Language models are unsupervised multitask learners.
\newblock \emph{OpenAI blog}, 1\penalty0 (8):\penalty0 9, 2019.

\bibitem[Sablayrolles et~al.(2019)Sablayrolles, Douze, Schmid, Ollivier, and
  J{\'e}gou]{sablayrolles2019white}
Sablayrolles, A., Douze, M., Schmid, C., Ollivier, Y., and J{\'e}gou, H.
\newblock White-box vs black-box: Bayes optimal strategies for membership
  inference.
\newblock In \emph{International Conference on Machine Learning}, pp.\
  5558--5567. PMLR, 2019.

\bibitem[Sainz et~al.(2023)Sainz, Campos, Garc{\'\i}a-Ferrero, Etxaniz,
  de~Lacalle, and Agirre]{sainz2023nlp}
Sainz, O., Campos, J.~A., Garc{\'\i}a-Ferrero, I., Etxaniz, J., de~Lacalle,
  O.~L., and Agirre, E.
\newblock Nlp evaluation in trouble: On the need to measure llm data
  contamination for each benchmark.
\newblock \emph{arXiv preprint arXiv:2310.18018}, 2023.

\bibitem[Salem et~al.(2018)Salem, Zhang, Humbert, Berrang, Fritz, and
  Backes]{salem2018ml}
Salem, A., Zhang, Y., Humbert, M., Berrang, P., Fritz, M., and Backes, M.
\newblock Ml-leaks: Model and data independent membership inference attacks and
  defenses on machine learning models.
\newblock \emph{arXiv preprint arXiv:1806.01246}, 2018.

\bibitem[Sennrich(2015)]{sennrich2015neural}
Sennrich, R.
\newblock Neural machine translation of rare words with subword units.
\newblock \emph{arXiv preprint arXiv:1508.07909}, 2015.

\bibitem[Shi et~al.(2023)Shi, Ajith, Xia, Huang, Liu, Blevins, Chen, and
  Zettlemoyer]{shi2023detecting}
Shi, W., Ajith, A., Xia, M., Huang, Y., Liu, D., Blevins, T., Chen, D., and
  Zettlemoyer, L.
\newblock Detecting pretraining data from large language models.
\newblock \emph{arXiv preprint arXiv:2310.16789}, 2023.

\bibitem[Shokri et~al.(2017)Shokri, Stronati, Song, and
  Shmatikov]{shokri2017membership}
Shokri, R., Stronati, M., Song, C., and Shmatikov, V.
\newblock Membership inference attacks against machine learning models.
\newblock In \emph{2017 IEEE symposium on security and privacy (SP)}, pp.\
  3--18. IEEE, 2017.

\bibitem[Singh et~al.(2024)Singh, Kocyigit, Poulton, Esiobu, Lomeli, Szilvasy,
  and Hupkes]{singh2024evaluation}
Singh, A.~K., Kocyigit, M.~Y., Poulton, A., Esiobu, D., Lomeli, M., Szilvasy,
  G., and Hupkes, D.
\newblock Evaluation data contamination in llms: how do we measure it and
  (when) does it matter?
\newblock \emph{arXiv preprint arXiv:2411.03923}, 2024.

\bibitem[Somepalli et~al.(2023)Somepalli, Singla, Goldblum, Geiping, and
  Goldstein]{somepalli2023diffusion}
Somepalli, G., Singla, V., Goldblum, M., Geiping, J., and Goldstein, T.
\newblock Diffusion art or digital forgery? investigating data replication in
  diffusion models.
\newblock In \emph{Proceedings of the IEEE/CVF Conference on Computer Vision
  and Pattern Recognition}, pp.\  6048--6058, 2023.

\bibitem[Team(2024)]{qwen2.5}
Team, Q.
\newblock Qwen2.5: A party of foundation models, September 2024.
\newblock URL \url{https://qwenlm.github.io/blog/qwen2.5/}.

\bibitem[Thudi et~al.(2022)Thudi, Jia, Shumailov, and
  Papernot]{thudi2022necessity}
Thudi, A., Jia, H., Shumailov, I., and Papernot, N.
\newblock On the necessity of auditable algorithmic definitions for machine
  unlearning.
\newblock In \emph{31st USENIX Security Symposium (USENIX Security 22)}, pp.\
  4007--4022, 2022.

\bibitem[Tirumala et~al.(2022)Tirumala, Markosyan, Zettlemoyer, and
  Aghajanyan]{tirumala2022memorization}
Tirumala, K., Markosyan, A., Zettlemoyer, L., and Aghajanyan, A.
\newblock Memorization without overfitting: Analyzing the training dynamics of
  large language models.
\newblock \emph{Advances in Neural Information Processing Systems},
  35:\penalty0 38274--38290, 2022.

\bibitem[Touvron et~al.(2023)Touvron, Martin, Stone, Albert, Almahairi, Babaei,
  Bashlykov, Batra, Bhargava, Bhosale, et~al.]{llama2}
Touvron, H., Martin, L., Stone, K., Albert, P., Almahairi, A., Babaei, Y.,
  Bashlykov, N., Batra, S., Bhargava, P., Bhosale, S., et~al.
\newblock Llama 2: Open foundation and fine-tuned chat models.
\newblock \emph{arXiv preprint arXiv:2307.09288}, 2023.

\bibitem[Yang et~al.(2024)Yang, Yang, Hui, Zheng, Yu, Zhou, Li, Li, Liu, Huang,
  et~al.]{yang2024qwen2}
Yang, A., Yang, B., Hui, B., Zheng, B., Yu, B., Zhou, C., Li, C., Li, C., Liu,
  D., Huang, F., et~al.
\newblock Qwen2 technical report.
\newblock \emph{arXiv preprint arXiv:2407.10671}, 2024.

\bibitem[Yeom et~al.(2018)Yeom, Giacomelli, Fredrikson, and
  Jha]{yeom2018privacy}
Yeom, S., Giacomelli, I., Fredrikson, M., and Jha, S.
\newblock Privacy risk in machine learning: Analyzing the connection to
  overfitting.
\newblock In \emph{2018 IEEE 31st computer security foundations symposium
  (CSF)}, pp.\  268--282. IEEE, 2018.

\bibitem[Zarifzadeh et~al.(2023)Zarifzadeh, Liu, and Shokri]{zarifzadeh2023low}
Zarifzadeh, S., Liu, P., and Shokri, R.
\newblock Low-cost high-power membership inference attacks, 2023.

\bibitem[Zhang et~al.(2024{\natexlab{a}})Zhang, Klyman, Mai, Levine, Zhang,
  Bommasani, and Liang]{zhang2024language}
Zhang, A.~K., Klyman, K., Mai, Y., Levine, Y., Zhang, Y., Bommasani, R., and
  Liang, P.
\newblock Language model developers should report train-test overlap.
\newblock \emph{arXiv preprint arXiv:2410.08385}, 2024{\natexlab{a}}.

\bibitem[Zhang et~al.(2024{\natexlab{b}})Zhang, Das, Kamath, and
  Tram{\`e}r]{zhang2024membership}
Zhang, J., Das, D., Kamath, G., and Tram{\`e}r, F.
\newblock Membership inference attacks cannot prove that a model was trained on
  your data.
\newblock \emph{arXiv preprint arXiv:2409.19798}, 2024{\natexlab{b}}.

\end{thebibliography}
\bibliographystyle{styles/icml2025}

\newpage
\appendix
\onecolumn

\section{Removing Members Does Not Always Prevent LLM Verbatim Completion (\S\ref{sec:pre-training})}
\label{sec:supp-pretraining}

\subsection{Data filtering}
\label{sec:supp-filtering-stats}

The $n$-gram filter discussed in \cref{sec:pretraining-setup} can be defined as follows:
\begin{definition}[$n$-gram data filtering] \label{def:ngram-filter}
Let $\mathcal F$ be a set of $n$-grams to filter against. Let $f_n(x, \mathcal F)$ be the condition that returns 1 if \textit{any} $n$-gram of the sequence $x$ is in $\mathcal F$ (0 otherwise).
An $n$-gram filter against $\mathcal F$ on a set of sequences $\mathcal D$ is thus constructing
$\tilde{\mathcal D}_n = \{ x \in \mathcal D \mid f_n(x, \mathcal F) = 0 \}.$
\end{definition}

Intuitively, a stronger $n$-gram filter (smaller $n$) means that we are removing a sequence on increasingly smaller partial matches against the filter set. In practice, since we are filtering many sequences (\Dmem) at once, it is more efficient to build a shared $n$-gram filter hash set from all sequences in \Dmem~ and then apply a sliding window filtering procedure.

\cref{tab:supp-filtered-fraction-seq50} shows the amount of tokens kept after applying $n$-gram filtering at different filtering strengths (for sequence length 50, the default setting used in experiments unless otherwise stated). \cref{tab:supp-filtered-fraction-seq100} similarly shows the amount of tokens for sequence length 100.

\begin{table}[h!]
\centering
\caption{Fraction of tokens kept after applying $n$-gram filter to sequence length $k=50$.}
\label{tab:supp-filtered-fraction-seq50}
\begin{tabular}{lcccc}
\toprule
\textbf{$n$-gram filtering strength} & \textbf{$n = 5$} & \textbf{$n = 10$} & \textbf{$n = 20$} & \textbf{$n = 50$ (Exact Filtering)} \\ \midrule
\textbf{Approx \% of tokens kept} & 0.6905 & 0.9838 & 0.9938 & 0.9995 \\
\bottomrule
\end{tabular}
\end{table}
\begin{table}[h!]
\centering
\caption{Fraction of tokens kept after applying $n$-gram filter to sequence length $k=100$.}
\label{tab:supp-filtered-fraction-seq100}
\begin{tabular}{lcccc}
\toprule
\textbf{$n$-gram filtering strength} & \textbf{$n = 10$} & \textbf{$n = 20$} & \textbf{$n = 50$} & \textbf{$n = 100$ (Exact Filtering)} \\ \midrule
\textbf{Approx \% of tokens kept} & 0.9820 & 0.9930 & 0.9970 & 0.9995 \\
\bottomrule
\end{tabular}
\end{table}

\subsection{Visualizing Lingering Sequences $\Dlinger{n}$}
\label{sec:supp-vis-lingering-seqs}

In this and the following section, we provide visualizations to the key results described in \cref{sec:pre-training} to help develop intuition on our findings.

Recall from \S\ref{sec:pre-training} that $\Dlinger{n}$ refers to the set of lingering sequences that are still verbatim completable by the re-trained from scratch LLM after applying $n$-gram filter (\cref{def:ngram-filter}) on the identified memorized sequences $\Dmem$. 

\cref{tab:vis-supp-lingering-seqs-n50}, \cref{tab:vis-supp-lingering-seqs-n20}, \cref{tab:vis-supp-lingering-seqs-n10}, and \cref{tab:vis-supp-lingering-seqs-n5} show 15 randomly sampled lingering sequence at filtering strengths $n=\{50, 20, 10, 5\}$, respectively. Observe that, as noted in \cref{sec:pretraining-results}, with stronger filtering strength (smaller $n$), the content gradually shift from semantically useful content to pattern continuations.

\begin{scriptsize}
\centering
\begin{longtable}{p{0.03\textwidth} p{0.9\textwidth}}
\caption{\textbf{Randomly sampled lingering sequences at filtering strength $n=50$ (exact) filter.} Sequence length $k=50$.}
\label{tab:vis-supp-lingering-seqs-n50}
\\
\hline
\textbf{Idx} & \textbf{Lingering sequences at filtering strength $n=50$ (exact) filter.}\\
\hline
\endfirsthead
\hline
\textbf{Idx} & \textbf{Lingering sequences at filtering strength $n=50$ (exact) filter.}\\
\hline
\endhead

0 & \textbf{Prompt:} \qquad
{'Common Name: bleeding heart Type: Herbaceous perennial Native Range: Eastern United States Zone: 3 to 9 '} \newline
\textbf{Completion:} \enspace
{'Height: 1.00 to 1.50 feet Spread: 1.00 to 1.50 feet Bloom Time:'}
\\ \hline

1 & \textbf{Prompt:} \qquad
{'|1477 by topic| |Arts and science| |Birth and death categories| |Births –'} \newline
\textbf{Completion:} \enspace
{' Deaths| |Establishments and disestablishments categories| |Establishments – Disestablishments| |'}
\\ \hline

2 & \textbf{Prompt:} \qquad
{'Charcot Joint (Neuropathic Arthropathy) Medicine Central™ is a quick-consult mobile and'} \newline
\textbf{Completion:} \enspace
{' web resource that includes diagnosis, treatment, medications, and follow-up information on over 700 diseases and disorders, providing fast answers'}
\\ \hline

3 & \textbf{Prompt:} \qquad
{'Mienert-barth Surname History The family history of the Mienert-barth last name is'} \newline
\textbf{Completion:} \enspace
{' maintained by the AncientFaces community. Join the community by adding to to our knowldge of the Mienert-'}
\\ \hline

4 & \textbf{Prompt:} \qquad
{'Instructional Supports and Resources Dyslexia is a specific learning disability that is neurological in origin. It is characterized'} \newline
\textbf{Completion:} \enspace
{' by difficulties with accurate and/or fluent word recognition and by poor spelling and decoding abilities. These difficulties typically result from a deficit'}
\\ \hline

5 & \textbf{Prompt:} \qquad
{'Publisher description for Writers at work. The short composition / Ann O. Strauch. Bibliographic record and links to'} \newline
\textbf{Completion:} \enspace
{' related information available from the Library of Congress catalog Information from electronic data provided by the publisher. May be incomplete or contain other'}
\\ \hline

6 & \textbf{Prompt:} \qquad
{'Create healthcare diagrams like this example called Anencephaly in minutes with SmartDraw. SmartDraw includes 1000s of professional healthcare'} \newline
\textbf{Completion:} \enspace
{' and anatomy chart templates that you can modify and make your own. Text in this Example: Anencephaly is'}
\\ \hline

7 & \textbf{Prompt:} \qquad
{'Presentation on theme: "Spiraled Assignments Presenter: Angela Pritchett November 14, 2006."'} \newline
\textbf{Completion:} \enspace
{'— Presentation transcript: Spiraled Assignments Presenter: Angela Pritchett November 14, 2006 '}
\\ \hline

8 & \textbf{Prompt:} \qquad
{'An excerpt from www.HouseOfNames.com archives copyright © 2000 - 2013 Where did the Irish McSweeney family'} \newline
\textbf{Completion:} \enspace
{' come from? What is the Irish McSweeney family crest and coat of arms? When did the McSweeney family first'}
\\ \hline

9 & \textbf{Prompt:} \qquad
{'Which of the following cubes can be made from these nets? Is it possible to remove ten unit cubes from a 3 by'} \newline
\textbf{Completion:} \enspace
{' 3 by 3 cube made from 27 unit cubes so that the surface area of the remaining solid is the same as the surface area'}
\\ \hline

10 & \textbf{Prompt:} \qquad
{'Course Hero. "The Libation Bearers Study Guide." Course Hero. 23 June 2017. Web. 14 Nov. 2018'} \newline
\textbf{Completion:} \enspace
{'. <https://www.coursehero.com/lit/The-Libation-Bearers/>. Course Hero'}
\\ \hline

11 & \textbf{Prompt:} \qquad
{'Presentation on theme: "Cause and Effect Comprehension Skill Fourth Grade Unit 2 Week 1 Created by Kristi Waltke'} \newline
\textbf{Completion:} \enspace
{'."— Presentation transcript: Cause and Effect Comprehension Skill Fourth Grade Unit 2 Week 1 Created by Kristi Walt'}
\\ \hline

12 & \textbf{Prompt:} \qquad
{'An excerpt from www.HouseOfNames.com archives copyright © 2000 - 2015 Where did the English Ragsdale family'} \newline
\textbf{Completion:} \enspace
{' come from? What is the English Ragsdale family crest and coat of arms? When did the Ragsdale family first'}
\\ \hline

13 & \textbf{Prompt:} \qquad
{'Gibbous Scorpio Moon phase on 29 March 2051 Wednesday is Waning Gibbous, 16'} \newline
\textbf{Completion:} \enspace
{' days old Moon is in Scorpio.Share this page: twitter facebook linkedin Previous main lunar phase is the Full Moon'}
\\ \hline

14 & \textbf{Prompt:} \qquad
{'What does AIDS mean in Laboratory? This page is about the meanings of the acronym/abbreviation/shorthand'} \newline
\textbf{Completion:} \enspace
{' AIDS in the Medical field in general and in the Laboratory terminology in particular. Find a translation for AIDS in other languages:'}
\\ \hline

15 & \textbf{Prompt:} \qquad
{'Nathalie Raphaëlle June 23, 2021 Worksheets If you home school your children, you'} \newline
\textbf{Completion:} \enspace
{' will quickly realize how important printable homeschool worksheets can be. If you are trying to develop a curriculum for your'}
\\ \hline

16 & \textbf{Prompt:} \qquad
{'Course Hero. "The Pearl Study Guide." Course Hero. 14 Dec. 2017. Web. 24 Nov. 2020. <'} \newline
\textbf{Completion:} \enspace
{'https://www.coursehero.com/lit/The-Pearl/>. Course Hero. (2017, December'}
\\ \hline

17 & \textbf{Prompt:} \qquad
{'Internet of Things Internet of Things The Internet of Things (IoT) is a system of interrelated computing devices'} \newline
\textbf{Completion:} \enspace
{', mechanical and digital machines, objects, animals or people that are provided with unique identifiers and the ability to transfer data over a'}
\\ \hline

18 & \textbf{Prompt:} \qquad
{'Latest Newland photos These photos were uploaded by members of the Newland community on AncientFaces. Newland S'} \newline
\textbf{Completion:} \enspace
{'urname History The family history of the Newland last name is maintained by the AncientFaces community. Join the community'}
\\ \hline

19 & \textbf{Prompt:} \qquad
{'Definition of Seckles 1. seckle [n] - See also: seckle Click the following'} \newline
\textbf{Completion:} \enspace
{' link to bring up a new window with an automated collection of images related to the term: Seckles Images Lexic'}
\\ \hline

\end{longtable}
\end{scriptsize}

\begin{scriptsize}
\centering
\begin{longtable}{p{0.03\textwidth} p{0.9\textwidth}}
\caption{\textbf{Randomly sampled lingering sequences at filtering strength $n=20$ filter}. Sequence length $k=50$.}
\label{tab:vis-supp-lingering-seqs-n20}
\\
\hline
\textbf{Idx} & \textbf{Lingering sequences at filtering strength $n=20$ filter.}\\
\hline
\endfirsthead
\hline
\textbf{Idx} & \textbf{Lingering sequences at filtering strength $n=20$ filter.}\\
\hline
\endhead

0 & \textbf{Prompt:} \qquad
{'Presentation on theme: "MAKING BOOKS WITH CHILDREN Picture It! Publish It! Read It!"'} \newline
\textbf{Completion:} \enspace
{'— Presentation transcript: MAKING BOOKS WITH CHILDREN Picture It! Publish It! Read It!'}
\\ \hline

1 & \textbf{Prompt:} \qquad
{'This Constitution, and the Laws of the United States which shall be made in Pursuance thereof; and all Treaties made'} \newline
\textbf{Completion:} \enspace
{', or which shall be made, under the Authority of the United States, shall be the supreme Law of the Land; and'}
\\ \hline

2 & \textbf{Prompt:} \qquad
{'How To Recognize A Crystal Child A selection of articles related to how to recognize a crystal child. Original articles from'} \newline
\textbf{Completion:} \enspace
{' our library related to the How To Recognize A Crystal Child. See Table of Contents for further available material (downloadable resources'}
\\ \hline

3 & \textbf{Prompt:} \qquad
{'Wampsville, New York |Wampsville, New York| |• Total||1.0 sq mi'} \newline
\textbf{Completion:} \enspace
{' (2.6 km2)| |• Land||1.0 sq mi (2.6 km2)| '}
\\ \hline

4 & \textbf{Prompt:} \qquad
{'Report on Stromboli (Italy) — 12 March-18 March 2003 Smithsonian / US Geological Survey Weekly Vol'} \newline
\textbf{Completion:} \enspace
{'canic Activity Report, 12 March-18 March 2003 Managing Editor: Gari Mayberry Please cite this report'}
\\ \hline

5 & \textbf{Prompt:} \qquad
{'Presentation on theme: "The Great (gym) Divide Curricula by Design \#3 M. Fischer."—'} \newline
\textbf{Completion:} \enspace
{' Presentation transcript: The Great (gym) Divide Curricula by Design \#3 M. Fischer The Great'}
\\ \hline

6 & \textbf{Prompt:} \qquad
{'Course Hero. "Lord of the Flies Study Guide." Course Hero. 15 Sep. 2016. Web. 29 May 20'} \newline
\textbf{Completion:} \enspace
{'23. <https://www.coursehero.com/lit/Lord-of-the-Flies/>. '}
\\ \hline

7 & \textbf{Prompt:} \qquad
{'Manada Gap, Pennsylvania facts for kids Quick facts for kids Manada Gap, Pennsylvania |Time zone||UTC'} \newline
\textbf{Completion:} \enspace
{'-5 (Eastern (EST))| |• Summer (DST)||UTC-4 (EDT)| '}
\\ \hline

8 & \textbf{Prompt:} \qquad
{'Scale Zoology Cosmoid Scales A selection of articles related to scale zoology cosmoid scales. Original'} \newline
\textbf{Completion:} \enspace
{' articles from our library related to the Scale Zoology Cosmoid Scales. See Table of Contents for further available material ('}
\\ \hline

9 & \textbf{Prompt:} \qquad
{'Atomic Nucleus History A selection of articles related to atomic nucleus history. Original articles from our library related to'} \newline
\textbf{Completion:} \enspace
{' the Atomic Nucleus History. See Table of Contents for further available material (downloadable resources) on Atomic Nucleus'}
\\ \hline

10 & \textbf{Prompt:} \qquad
{'Mangoverde :: World Bird Guide :: Pheasants and Partridges :: Common Quail Common Quail Cot'} \newline
\textbf{Completion:} \enspace
{'urnix coturnix Described by: Linnaeus (1758) Alternate common name(s'}
\\ \hline

11 & \textbf{Prompt:} \qquad
{'Set Builder Notation Variations A selection of articles related to set builder notation variations. Original articles from our library related'} \newline
\textbf{Completion:} \enspace
{' to the Set Builder Notation Variations. See Table of Contents for further available material (downloadable resources) on Set Builder'}
\\ \hline

12 & \textbf{Prompt:} \qquad
{'|1648 by topic| |Arts and science| |Birth and death categories| |Births –'} \newline
\textbf{Completion:} \enspace
{' Deaths| |Establishments and disestablishments categories| |Establishments – Disestablishments| |'}
\\ \hline

13 & \textbf{Prompt:} \qquad
{'Tamil Script The Tamil Letters A selection of articles related to tamil script the tamil letters. Original articles from'} \newline
\textbf{Completion:} \enspace
{' our library related to the Tamil Script The Tamil Letters. See Table of Contents for further available material (downloadable resources) on'}
\\ \hline

14 & \textbf{Prompt:} \qquad
{'Manuel I of PortugalFrom Wikipedia, the free encyclopediaJump to: navigation, search This article does not cite any references or'} \newline
\textbf{Completion:} \enspace
{' sources. Please help improve this article by adding citations to reliable sources. Unsourced material may be challenged and removed. ('}
\\ \hline

15 & \textbf{Prompt:} \qquad
{'Image 1 of 12 Image 2 of 12 Image 3 of 12 Image 4 of 12 Image 5 of 12 '} \newline
\textbf{Completion:} \enspace
{'Image 6 of 12 Image 7 of 12 Image 8 of 12 Image 9 of 12 Image 10 of 12 '}
\\ \hline

16 & \textbf{Prompt:} \qquad
{'|Nutritional Guidelines (per serving)| |Servings: 3 pint jars (96 servings)| |Amount per serving'} \newline
\textbf{Completion:} \enspace
{'| |\% Daily Value*| |Total Fat 0g||0\%| |Saturated Fat 0g'}
\\ \hline

17 & \textbf{Prompt:} \qquad
{'Presentation on theme: "Corpus Linguistics and Stylistics PALA Summer School, Maribor, 2014'} \newline
\textbf{Completion:} \enspace
{'."— Presentation transcript: Corpus Linguistics and Stylistics PALA Summer School, Maribor,'}
\\ \hline

18 & \textbf{Prompt:} \qquad
{'Presentation on theme: "Lunar Research Station Design Submitted by West Valley Elementary GATE Team October 31, 2006'} \newline
\textbf{Completion:} \enspace
{'."— Presentation transcript: Lunar Research Station Design Submitted by West Valley Elementary GATE Team October 31,'}
\\ \hline

19 & \textbf{Prompt:} \qquad
{'Some daily events in the changing sky for February 19 27. Friday, February 19 Saturday, February 20 Sunday,'} \newline
\textbf{Completion:} \enspace
{' February 21 Monday, February 22 Tuesday, February 23 Wednesday, February 24 Thursday, February 25 Friday,'}
\\ \hline

\end{longtable}
\end{scriptsize}

\begin{scriptsize}
\centering
\begin{longtable}{p{0.03\textwidth} p{0.9\textwidth}}
\caption{\textbf{Randomly sampled lingering sequences at filtering strength $n=10$ filter}. Sequence length $k=50$.}
\label{tab:vis-supp-lingering-seqs-n10}
\\
\hline
\textbf{Idx} & \textbf{Lingering sequences at filtering strength $n=10$ filter.}\\
\hline
\endfirsthead
\hline
\textbf{Idx} & \textbf{Lingering sequences at filtering strength $n=10$ filter.}\\
\hline
\endhead

0 & \textbf{Prompt:} \qquad
{'Presentation on theme: "HELPING YOUR CHILD WITH NUMERACY: ADDITION AND SUBTRACTION."'} \newline
\textbf{Completion:} \enspace
{'— Presentation transcript: HELPING YOUR CHILD WITH NUMERACY: ADDITION AND SUBTRACTION '}
\\ \hline

1 & \textbf{Prompt:} \qquad
{'|Wednesday||2:00 PM - 3:40 PM||lesson||Lecture Hall 1.2| '} \newline
\textbf{Completion:} \enspace
{'|Thursday||2:00 PM - 3:40 PM||lesson||Lecture Hall 1.2| '}
\\ \hline

2 & \textbf{Prompt:} \qquad
{'How to define the cosine ratio and identify the cosine of an angle in a right triangle. How to define the'} \newline
\textbf{Completion:} \enspace
{' sine ratio and identify the sine of an angle in a right triangle. How to define the tangent ratio and'}
\\ \hline

3 & \textbf{Prompt:} \qquad
{'Q1. A series is given with one term missing. Select the correct alternative from the given ones that will complete the series'} \newline
\textbf{Completion:} \enspace
{'. Q2. A series is given with one term missing. Select the correct alternative from the given ones that will complete'}
\\ \hline

4 & \textbf{Prompt:} \qquad
{'History of False Teeth Length: 497 words (1.4 double-spaced pages) - - -'} \newline
\textbf{Completion:} \enspace
{' - - - - - - - - - - - - - - - - - - - - - - - - -'}
\\ \hline

5 & \textbf{Prompt:} \qquad
{'Presentation on theme: "Yoghurt!!! Find the dairy cow on each page!!! By Daisy Mason and Brigette Roberts'} \newline
\textbf{Completion:} \enspace
{'."— Presentation transcript: Yoghurt!!! Find the dairy cow on each page!!! By Daisy Mason and Brigette'}
\\ \hline

6 & \textbf{Prompt:} \qquad
{'Protecting People with Disabilities in the Ebbs and Flows of the COVID-19 Pandemic Protecting People'} \newline
\textbf{Completion:} \enspace
{' with Disabilities in the Ebbs and Flows of the COVID-19 Pandemic The COVID-19 pand'}
\\ \hline

7 & \textbf{Prompt:} \qquad
{'Presentation on theme: "Aceh Poverty Assessment The impact of the Conflict, the Tsunami and Reconstruction on Poverty'} \newline
\textbf{Completion:} \enspace
{' in Aceh."— Presentation transcript: Aceh Poverty Assessment The impact of the Conflict, the Tsunami'}
\\ \hline

8 & \textbf{Prompt:} \qquad
{'Presentation on theme: "THE MIX-AERATOR Innovation In Pond \& Lagoon Aeration \& Mixing."'} \newline
\textbf{Completion:} \enspace
{'— Presentation transcript: THE MIX-AERATOR Innovation In Pond \& Lagoon Aeration \& Mixing '}
\\ \hline

9 & \textbf{Prompt:} \qquad
{'Some daily events in the changing sky for February 8 16. Friday, February 8 Saturday, February 9 Sunday,'} \newline
\textbf{Completion:} \enspace
{' February 10 Monday, February 11 Tuesday, February 12 Wednesday, February 13 Thursday, February 14 Friday,'}
\\ \hline

10 & \textbf{Prompt:} \qquad
{'Essays on mercutio Romeo and mercutio essays: over 180,000 romeo and merc'} \newline
\textbf{Completion:} \enspace
{'utio essays, romeo and mercutio term papers, romeo and mercutio research paper, book'}
\\ \hline

11 & \textbf{Prompt:} \qquad
{'Presentation on theme: "Fabric Construction Fashion Design, Textiles \& Merchandising Mrs. Moscinski."—'} \newline
\textbf{Completion:} \enspace
{' Presentation transcript: Fabric Construction Fashion Design, Textiles \& Merchandising Mrs. Moscinski Fabric'}
\\ \hline

12 & \textbf{Prompt:} \qquad
{'Presentation on theme: "Chapter 4 - Building Compassionate School-Community Partnerships That Work Chapter 4 - Building Comp'} \newline
\textbf{Completion:} \enspace
{'assionate School-Community Partnerships That Work."— Presentation transcript: Chapter 4 - Building Compassionate School-'}
\\ \hline

13 & \textbf{Prompt:} \qquad
{'Tracing Names: Letter AA | B | C | D | E | F | G | H | I | J |'} \newline
\textbf{Completion:} \enspace
{' K | L | M | N | O | P | Q | R | S | T | U | V | W'}
\\ \hline

14 & \textbf{Prompt:} \qquad
{'Dictionary of Financial, Economic, and Business Terms A | B | C | D | E | F | G |'} \newline
\textbf{Completion:} \enspace
{' H | I | J | K | L | M | N | O | P | Q | R | S | T'}
\\ \hline

15 & \textbf{Prompt:} \qquad
{'Canons of the Seven Ecumenical Councils. The First Ecumenical Council. Second Ecumenical'} \newline
\textbf{Completion:} \enspace
{' Council. Third Ecumenical Council. Fourth Ecumenical Council. Fifth Ecumenical Council. '}
\\ \hline

16 & \textbf{Prompt:} \qquad
{'Welsh Levels of Care E-Learning Program Glossary Special | A | B | C | D | E | F'} \newline
\textbf{Completion:} \enspace
{' | G | H | I | J | K | L | M | N | O | P | Q | R |'}
\\ \hline

17 & \textbf{Prompt:} \qquad
{'Presentation on theme: "Tap Water Intrusion Effects on Microbial Life Anthony DeRenzo Grade 10 Pittsburgh Central Catholic'} \newline
\textbf{Completion:} \enspace
{' High School."— Presentation transcript: Tap Water Intrusion Effects on Microbial Life Anthony DeRenzo Grade 10'}
\\ \hline

18 & \textbf{Prompt:} \qquad
{'Presentation on theme: "Dr. Anand Srinivasan for MBBS 2013 on 10/10/2013."'} \newline
\textbf{Completion:} \enspace
{'— Presentation transcript: Dr. Anand Srinivasan for MBBS 2013 on 10/10/2013 '}
\\ \hline

19 & \textbf{Prompt:} \qquad
{'Presentation on theme: "Ashok Sinha O/o the Director General (Audit) Central, Chandigar'} \newline
\textbf{Completion:} \enspace
{'h."— Presentation transcript: Ashok Sinha O/o the Director General (Audit) Central, Chand'}
\\ \hline

\end{longtable}
\end{scriptsize}

\begin{scriptsize}
\centering
\begin{longtable}{p{0.03\textwidth} p{0.9\textwidth}}
\caption{\textbf{Randomly sampled lingering sequences at filtering strength $n=5$ filter}. Sequence length $k=50$.}
\label{tab:vis-supp-lingering-seqs-n5}
\\
\hline
\textbf{Idx} & \textbf{Lingering sequences at filtering strength $n=5$ filter.}\\
\hline
\endfirsthead
\hline
\textbf{Idx} & \textbf{Lingering sequences at filtering strength $n=5$ filter.}\\
\hline
\endhead

0 & \textbf{Prompt:} \qquad
{'Water, sanitation and hygiene: the foundation for building resilience in climate-vulnerable communities - Water, sanitation and hygiene:'} \newline
\textbf{Completion:} \enspace
{' the foundation for building resilience in climate-vulnerable communities - Water, sanitation and hygiene: the foundation for building resilience in'}
\\ \hline

1 & \textbf{Prompt:} \qquad
{'- 1 What is Adrenoleukodystrophy disease? - 2 Adrenoleukodystrophy Causes - 3'} \newline
\textbf{Completion:} \enspace
{' Adrenoleukodystrophy Symptoms - 4 Adrenoleukodystrophy Diagnosis - 5 Adrenoleukody'}
\\ \hline

2 & \textbf{Prompt:} \qquad
{'MATH105 April 2017 • Q1 (a) • Q1 (b) • Q1 (c) •'} \newline
\textbf{Completion:} \enspace
{' Q1 (d) • Q1 (e) • Q1 (f) • Q1 (g) • Q'}
\\ \hline

3 & \textbf{Prompt:} \qquad
{'Native to North America STATE DISTRIBUTION (USDA): AL, AR, CT, DC, DE, FL,'} \newline
\textbf{Completion:} \enspace
{' GA, IA, IL, IN, KS, KY, LA, MA, MD, ME, MI, MN, MO'}
\\ \hline

4 & \textbf{Prompt:} \qquad
{'What are the 7 notes of a major scale? The scale degrees are: - 1st: Tonic. '} \newline
\textbf{Completion:} \enspace
{'- 2nd: Supertonic. - 3rd: Mediant. - 4th: Subdominant.'}
\\ \hline

5 & \textbf{Prompt:} \qquad
{'Ten Times Table And Random Test Lyrics 10 x 1 = 10 10 x 2 = 20 10 x 3 ='} \newline
\textbf{Completion:} \enspace
{' 30 10 x 4 = 40 10 x 5 = 50 10 x 6 = 60 10 x 7 = 70'}
\\ \hline

6 & \textbf{Prompt:} \qquad
{'Accuracy Of Data 914 words (2.6 double-spaced pages) - - - - - -'} \newline
\textbf{Completion:} \enspace
{' - - - - - - - - - - - - - - - - - - - - - - - - -'}
\\ \hline

7 & \textbf{Prompt:} \qquad
{'Print Texting RULES! Reading Comprehension with Fourth Grade Work Print Texting RULES! Reading Comp'} \newline
\textbf{Completion:} \enspace
{'rehension with Fifth Grade Work Print Texting RULES! Reading Comprehension with Sixth Grade Work Print Text'}
\\ \hline

8 & \textbf{Prompt:} \qquad
{'Glossary of Legal Terms A - B - C - D - E - F - G - H'} \newline
\textbf{Completion:} \enspace
{' - I - J - K - L - M - N - O - P - Q - R'}
\\ \hline

9 & \textbf{Prompt:} \qquad
{'Chef is at x=0. 1-jump: he will move from x -> x + 1 2-'} \newline
\textbf{Completion:} \enspace
{'jump: he will move from x -> x + 2 3-jump: he will move from x -> x + 3'}
\\ \hline

10 & \textbf{Prompt:} \qquad
{'Acting Minister of the Environment, Denis Kellman (centre, ‘, ‘, ‘, ‘'} \newline
\textbf{Completion:} \enspace
{', ‘, ‘, ‘, ‘, ‘, ‘, ‘, ‘,'}
\\ \hline

11 & \textbf{Prompt:} \qquad
{'MI Science Standards Special | A | B | C | D | E | F | G | H | I | J'} \newline
\textbf{Completion:} \enspace
{' | K | L | M | N | O | P | Q | R | S | T | U | V |'}
\\ \hline

12 & \textbf{Prompt:} \qquad
{'Collective Nouns for Birds |Pages:||A,||B,||C,||D,||E'} \newline
\textbf{Completion:} \enspace
{',||F,||G,||H,||I,||J,||K,||L,||M,'}
\\ \hline

13 & \textbf{Prompt:} \qquad
{'7 Wicked Winter Health Myths By: Laura Roberson - Winter Health Myth \# 1 - Winter Health Myth \#'} \newline
\textbf{Completion:} \enspace
{' 2 - Winter Health Myth \# 3 - Winter Health Myth \# 4 - Winter Health Myth \# 5 - Winter'}
\\ \hline

14 & \textbf{Prompt:} \qquad
{'HISTORY CRIME AND PUNISHMENT HISTORY CRIME AND PUNISHMENT HISTORY CRIME AND'} \newline
\textbf{Completion:} \enspace
{' PUNISHMENT HISTORY CRIME AND PUNISHMENT HISTORY CRIME AND PUNISHMENT '}
\\ \hline

15 & \textbf{Prompt:} \qquad
{'Define Gyromitra infula. Gyromitra infula synonyms, Gyromitra in'} \newline
\textbf{Completion:} \enspace
{'fula pronunciation, Gyromitra infula translation, English dictionary definition of Gyromitra infula.'}
\\ \hline

16 & \textbf{Prompt:} \qquad
{'Some daily events in the changing sky for December 19 27. Friday, December 19 Saturday, December 20 Sunday,'} \newline
\textbf{Completion:} \enspace
{' December 21 Monday, December 22 Tuesday, December 23 Wednesday, December 24 Thursday, December 25 Friday,'}
\\ \hline

17 & \textbf{Prompt:} \qquad
{'Length: 1122 words (3.2 double-spaced pages) - - - - - - - - -'} \newline
\textbf{Completion:} \enspace
{' - - - - - - - - - - - - - - - - - - - - - - - - -'}
\\ \hline

18 & \textbf{Prompt:} \qquad
{'Letter E Names: Page 2A | B | C | D | E | F | G | H | I | J'} \newline
\textbf{Completion:} \enspace
{' | K | L | M | N | O | P | Q | R | S | T | U | V |'}
\\ \hline

19 & \textbf{Prompt:} \qquad
{'A Complete Illustrated History of Robots in the Movies (chronological by film title) Intro | Part 1 |'} \newline
\textbf{Completion:} \enspace
{' Part 2 | Part 3 | Part 4 | Part 5 | Part 6 | Part 7 | Part 8 | Part 9 | Part'}
\\ \hline

20 & \textbf{Prompt:} \qquad
{'1. George Washington - Term of Office (1789-1797) 2. John Adams (1797-18'} \newline
\textbf{Completion:} \enspace
{'01) 3. Thomas Jefferson (1801-1809) 4. James Madison (1809-1817'}
\\ \hline

\end{longtable}
\end{scriptsize}

\subsection{Visualizing neighbors of lingering sequences $\Dlinger{n}$ in the pre-training set}
\label{sec:supp-vis-lingering-neighbors}

A key aspect worth studying for lingering sequences is {what contributed to their existence}. We randomly sample 2 lingering sequences for filtering strengths $n = \{50, 20, 10\}$, and perform  a very costly search of Levenshtein edit-distance neighbor search: perform a sliding window over the pre-training tokens, and check the edit distance of each window to the lingering sequence. By construction, lingering sequences have already been removed from the training set; the hope is thus to identify and visualize \textit{neighboring} sequences that may have led to these lingering sequences.

\begin{figure}[ht]
    \centering
    \includegraphics[width=\linewidth]{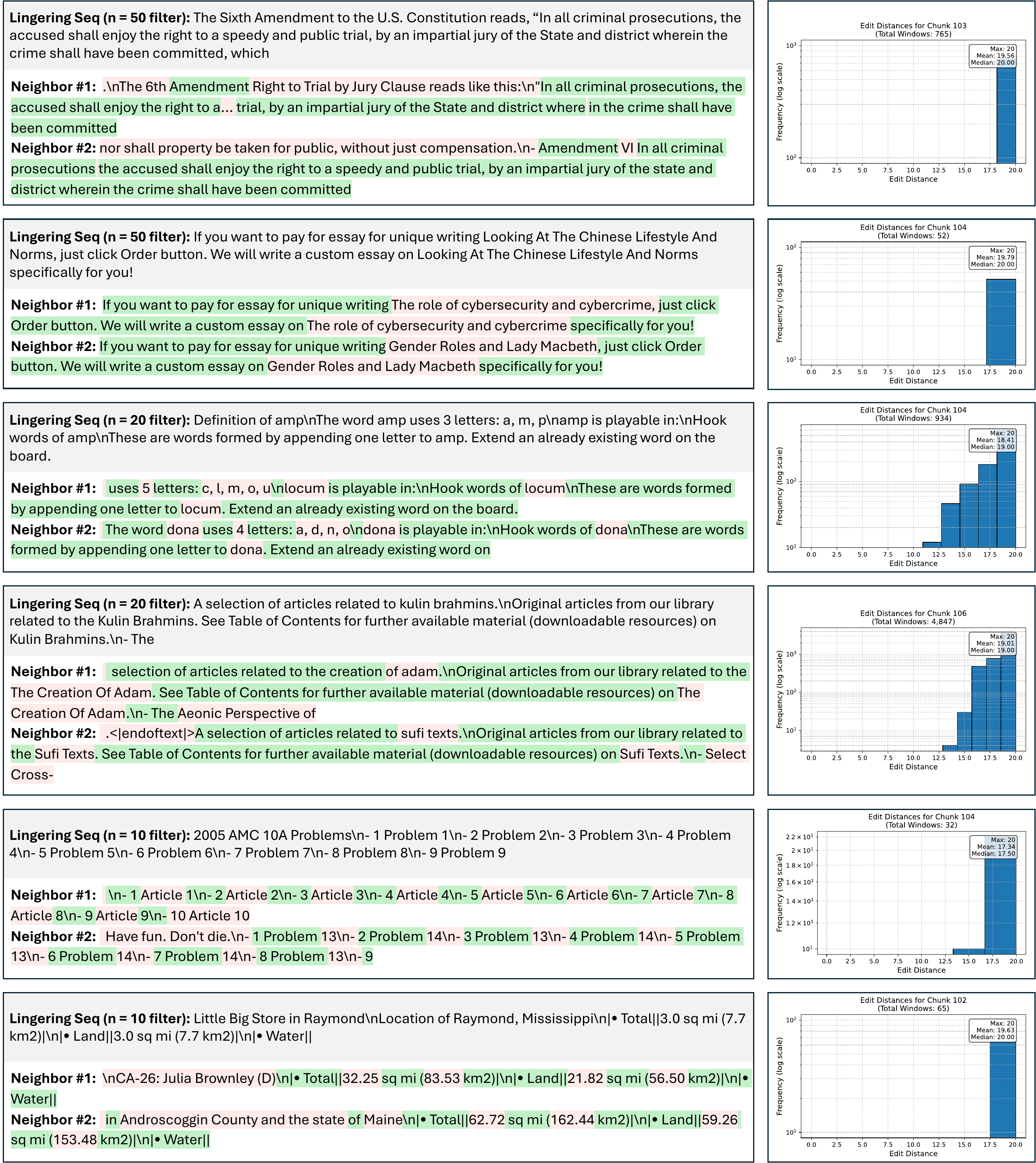}
    \caption{Visualizing the neighbors (Levenshtein edit distance $< 20$ tokens) of a few randomly selected lingering sequence from $\Dlinger{10}$, \Dlinger{20}, and \Dlinger{50}.
    }
    \label{fig:supp-vis-neighbors}
\end{figure}

\cref{fig:supp-vis-neighbors} visualizes these randomly sampled lingering sequences, and two randomly sampled neighbors (with edit distance $< 20$ tokens) for each of them. We also visualize the histogram of these neighbors at different distances. We note that:
\begin{itemize}
    \item These visualizations suggest that lingering sequences are very likely the result of either \textit{near-duplicate} training data and/or the generalization capabilities of LLMs. 
    \item \textbf{By observing the neighboring sequences, we gain insights into how we may \textit{adversarially} game the $n$-gram membership definition.} For example, observing the second sequence in \cref{fig:supp-vis-neighbors} can provide intuition for the \textit{chunking} (\S\ref{sec:chunking}) method we presented in \S\ref{sec:fine-tuning} to adversarially construct fine-tuning sequences that avoid $n$-gram overlap, and observing the second last can provide intuition for the \textit{token dropout} (\S\ref{sec:goldfish}) method.  
\end{itemize}

\clearpage

\subsection{Model architectures}
\label{sec:supp-architectures}

Our 350M, 774M, and 1.6B-parameter model architectures follow directly from the original GPT-2 paper \cite{radford2019language} and LLM.c implementation~\cite{karpathy2024llmc}. We create a custom architecture for 2.8B parameters by adjusting the number of layers, channels, and attention headas in the model, again following \citet{karpathy2024llmc}. The model configurations can be found in \cref{tab:model-configs}.

\begin{table}[h!]
\centering
\caption{Configurations of models (GPT-2 architecture) at different sizes for our pre-training experiments.}
\vspace{3mm}
\label{tab:model-configs}
\begin{tabular}{@{}lccccc@{}}
\toprule
\textbf{Parameters}       & \textbf{304M} & \textbf{774M}   & \textbf{1.6B}   & \textbf{2.8B}   \\ \midrule
Layers           & 24   & 36   & 48   & 60    \\
Channels         & 1024 & 1280 & 1600 & 1920  \\
Attention heads  & 16   & 20   & 26   & 30    \\ \bottomrule
\end{tabular}
\end{table}

\subsection{Training hyperparameters}
\label{sec:supp-pretraining-configs}

We follow the pre-training configurations outlined in \cite{karpathy2024llmc}, particularly \url{https://github.com/karpathy/llm.c/discussions/677}. \cref{tab:pretraining-configs} summarizes the configurations.

\begin{table}[h!]
\centering
\caption{Training configurations for pre-training experiments.}
\vspace{-3mm}
\label{tab:pretraining-configs}
\begin{tabular}{@{}ll@{}}
\toprule
\textbf{Hyperparameter}       &  \textbf{Value}   \\ \midrule
\# Training Tokens            & 33.6 billion      \\
Compute                       & 8 NVIDIA H100 days (1.6B parameter model) \\
Micro-Batch Size              & 16                \\
Max Sequence Length           & 1024              \\
Total Batch Size              & $2^{20} = 1,048,576$ tokens \\
Gradient Accumulation Steps   & 8                 \\
Weight Decay                  & 0.1               \\
Learning Rate                 & 6e-4              \\
LR Schedule                   & Cosine            \\
LR Decay                      & decay to 10\%  of max LR \\
Warmup Iterations             & 700 iterations    \\
Total Training Steps          & 32,000            \\
\bottomrule
\end{tabular}
\end{table}

\subsection{Tabled results of \cref{fig:pretrain-scale}}
\label{sec:supp-lingering-stats}

We additionally provide the table version of \cref{fig:pretrain-scale} in \cref{tab:supp-lingering-k50} (for default sequence length 50).

\begin{table}[h]
    \centering
    \captionof{table}{(Table version of \cref{fig:pretrain-scale}) \textbf{Amount of lingering sequences} as fraction of $|\Dmem|$ across model sizes and filtering strengths (values of $n$-gram filter). Sequence length $k=50$. Recall setup in \cref{sec:pre-training}.}
    \label{tab:supp-lingering-k50}
    \begin{tabular}{cccccc}
    \toprule
    \multirow{2}{*}{\textbf{Model Size}} & \multirow{2}{*}{\textbf{$|\Dmem|$}} & \multicolumn{4}{c}{\textbf{Filtering Strength}} \\ 
     &  & $n=5$ & $n=10$ & $n=20$ & $n=50$ \\ \hline
    304M & 76648 & 0.0175 & 0.0402 & 0.0978 & 0.4793 \\
    774M & 116270 & 0.0132 & 0.0300 & 0.0916 & 0.4013 \\
    1.6B & 151598 & 0.0109 & 0.0259 & 0.0796 & 0.4007 \\
    2.8B & 175813 & 0.0099 & 0.0247 & 0.0869 & 0.4068 \\
    \bottomrule
    \end{tabular}
\end{table}

\clearpage
\subsection{Additional Results: Longer Sequence Length}
\label{sec:supp-pretrain-longer-seq}

We also experiment with a longer sequence length of $k=100$ (i.e., 50-token prompt, 50-token completion). In general, we made similar observations in terms of both the amount and nature of the lingering sequences. \cref{tab:supp-lingering-k100} shows the fraction of lingering sequences in terms of the identified memorized sequences $\Dmem$. We provide visualizations of these sequences in \cref{tab:vis-supp-lingering-seqs-n100-k100} (exact filtering, $n=100$) and \cref{tab:vis-supp-lingering-seqs-n100-k100} (strong filtering, $n=10$), and note that the {content} of the lingering sequences are similar to those with sequence length $k=50$.

\begin{table}[h!]
    \centering
    \captionof{table}{{Amount of lingering sequences} as a fraction of the identified memorized sequences $|\Dmem|$ across filtering strengths (values of $n$-gram filter). Sequence length $k=100$. Recall setup in \cref{sec:pre-training}; compare to \cref{tab:supp-lingering-k50}.}
    \vspace{-1.5mm}
    \label{tab:supp-lingering-k100}
    \begin{tabular}{cccccc}
    \toprule
    \multirow{2}{*}{\textbf{Model Size}} & \multirow{2}{*}{\textbf{$|\Dmem|$}} & \multicolumn{4}{c}{\textbf{Filtering Strength}} \\ 
     &  & $n=10$ & $n=20$ & $n=50$ & $n=100$ \\ \hline
    1.6B & 88803 & 0.0109 & 0.0259 & 0.0796 & 0.4007 \\
    \bottomrule
    \end{tabular}
\end{table}

\begin{scriptsize}
\centering
\begin{longtable}{p{0.03\textwidth} p{0.9\textwidth}}
\caption{\textbf{Randomly sampled lingering sequences at filtering strength $n=100$ (exact) filter}. Sequence length $k=100$.}
\label{tab:vis-supp-lingering-seqs-n100-k100}
\vspace{-5pt}
\\
\toprule
\textbf{Idx} & \textbf{Lingering sequences at filtering strength $n=100$ (exact) filter.}\\
\hline
\endfirsthead
\hline
\textbf{Idx} & \textbf{Lingering sequences at filtering strength $n=100$ (exact) filter.}\\
\hline
\endhead

0 & \textbf{Prompt:} \qquad
{''Course Hero. "Romeo and Juliet Study Guide." Course Hero. 28 July 2016. Web. 16 Jan. 2019. <https://www.coursehero.com/lit/Romeo-and-Juliet/>. Course''} \newline
\textbf{Completion:} \enspace
{'' Hero. (2016, July 28). Romeo and Juliet Study Guide. In Course Hero. Retrieved January 16, 2019, from https://www.coursehero.com/lit/Romeo-and-Juliet/ (Course Hero,''}
\\ \hline

1 & \textbf{Prompt:} \qquad
{''ESQUIEN INDIANS. The Esquien Indians were a Karankawan group that entered Nuestra Señora de la Candelaria Mission on the San Gabriel River in 1750. Frederick Webb Hodge, ed.,''} \newline
\textbf{Completion:} \enspace
{'' Handbook of American Indians North of Mexico (2 vols., Washington: GPO, 1907, 1910; rpt., New York: Pageant, 1959). The following, adapted from the Chicago Manual of Style, 15th edition, is the''}
\\ \hline

2 & \textbf{Prompt:} \qquad
{''See what questions a doctor would ask. During a consultation, your doctor will use various techniques to assess the symptom: Heel pain. These will include a physical examination and possibly diagnostic tests. (Note: A physical exam is always done,''} \newline
\textbf{Completion:} \enspace
{'' diagnostic tests may or may not be performed depending on the suspected condition) Your doctor will ask several questions when assessing your condition. It is important to openly share any pertinent information to help your doctor make an accurate diagnosis. It is also very important to''}
\\ \hline

3 & \textbf{Prompt:} \qquad
{''|Product \#: EMC0775025\_TQ| A Is for Apple (Resource Book Only) eBookGrade 2|Grade 3|Grade 4|Grade 5 Please Note: This ebook is a digital download, NOT a physical product''} \newline
\textbf{Completion:} \enspace
{''. After purchase, you will be provided a one time link to download ebooks to your computer. Orders paid by PayPal require up to 8 business hours to verify payment and release electronic media. For immediate downloads, payment with credit card is required. ''}
\\ \hline

4 & \textbf{Prompt:} \qquad
{''Family History and Genealogy Resources by Surname Haycock Surname Origin A name probably given to a foundling exposed in a hayfield. Source: An Etymological Dictionary of Family and Christian Names With an Essay''} \newline
\textbf{Completion:} \enspace
{'' on their Derivation and Import; Arthur, William, M.A.; New York, NY: Sheldon, Blake, Bleeker \& CO., 1857. Haycock Surname Meaning and Family Facts There is more to Haycock family''}
\\ \hline

5 & \textbf{Prompt:} \qquad
{'"Lake Uvs and its surrounding wetlands - Site number:1379 - Area:585,000 ha - Designation date:22-03-2004 - Coordinates:50$^{\circ}$19'N 92$^{\circ}$45'E Materials"'} \newline
\textbf{Completion:} \enspace
{'' presented on this website, particularly maps and territorial information, are as-is and as-available based on available data and do not imply the expression of any opinion whatsoever on the part of the Secretariat of the Ramsar Convention concerning the legal status of''}
\\ \hline

6 & \textbf{Prompt:} \qquad
{''Barrier Methods of Birth Control (cont.) Melissa Conrad Stöppler, MD Melissa Conrad Stöppler, MD, is a U.S. board-certified Anatomic Pathologist with subspecialty training in''} \newline
\textbf{Completion:} \enspace
{'" the fields of Experimental and Molecular Pathology. Dr. Stöppler's educational background includes a BA with Highest Distinction from the University of Virginia and an MD from the University of North Carolina. She completed residency training in Anatomic Pathology at"'}
\\ \hline

7 & \textbf{Prompt:} \qquad
{''Definition of Japanese deer 1. Noun. Small deer of Japan with slightly forked antlers. Generic synonyms: Cervid, Deer Group relationships: Cervus, Genus Cervus Japanese Deer Pictures Click''} \newline
\textbf{Completion:} \enspace
{'' the following link to bring up a new window with an automated collection of images related to the term: Japanese Deer Images Lexicographical Neighbors of Japanese Deer Literary usage of Japanese deer Below you will find example usage of this term as''}
\\ \hline

8 & \textbf{Prompt:} \qquad
{''Comparing Fractions (G) In this comparing fractions practice worksheet, 5th graders examine 10 pairs of fractions. Students identify each of the pairs of fractions as greater than, less than, or equal to one another. 3 Views''} \newline
\textbf{Completion:} \enspace
{'' 0 Downloads Fraction Equivalence, Ordering, and Operations Need a unit to teach fractions to fourth graders? Look no further than this well-developed and thorough set of lessons that takes teachers through all steps of planning, implementing,''}
\\ \hline

9 & \textbf{Prompt:} \qquad
{''PREAMBLEWhereas recognition of the inherent dignity and of the equal and inalienable rights of all members of the human family is the foundation of freedom, justice and peace in the world, Whereas disregard and contempt for human rights have resulted in barbar''} \newline
\textbf{Completion:} \enspace
{''ous acts which have outraged the conscience of mankind, and the advent of a world in which human beings shall enjoy freedom of speech and belief and freedom from fear and want has been proclaimed as the highest aspiration of the common people, Whereas it is essential''}
\\ \hline

\end{longtable}
\end{scriptsize}

\begin{scriptsize}
\centering
\begin{longtable}{p{0.03\textwidth} p{0.9\textwidth}}
\caption{\textbf{Randomly sampled lingering sequences at filtering strength $n=10$ filter}. Sequence length $k=100$.}
\label{tab:vis-supp-lingering-seqs-n10-k100}
\vspace{-5pt}
\\
\toprule
\textbf{Idx} & \textbf{Lingering sequences at filtering strength $n=10$ filter.}\\
\hline
\endfirsthead
\hline
\textbf{Idx} & \textbf{Lingering sequences at filtering strength $n=10$ filter.}\\
\hline
\endhead

0 & \textbf{Prompt:} \qquad
{''Presentation on theme: "REVIEW We can tell how many electrons and atom will gain or lose by looking at its valence. Metals like to lose electrons. (Cations) –Ex. Na + Nonmetals."— Presentation''} \newline
\textbf{Completion:} \enspace
{'' transcript: REVIEW We can tell how many electrons and atom will gain or lose by looking at its valence. Metals like to lose electrons. (Cations) –Ex. Na + Nonmetals like to gain electrons. (An''}
\\ \hline

1 & \textbf{Prompt:} \qquad
{''|I • II • III • IV • V • VI • VII • VIII • IX • X • XI • XII • XIII • XIV • Schedule| - 1 Features - 2 Preamble - 3 Article I - 4 Article''} \newline
\textbf{Completion:} \enspace
{'' II - 5 Article III - 6 Article IV - 7 Article V - 8 Article VI - 9 Article VII - 10 Article VIII - 11 Article IX - 12 Article X - 13 Article XI - 14 Article''}
\\ \hline

2 & \textbf{Prompt:} \qquad
{''ENGL 301 Course Introduction this course will be an independent study of a topic of the student’s choice in English literature and film. ENGL 302 Course Introduction for ENGL 302 – Independent Study In English (ENGL 302) this''} \newline
\textbf{Completion:} \enspace
{'' course will be an independent study of a topic of the student’s choice in English literature and film. ENGL 303 Course Introduction for ENGL 303 – Independent Study In English (ENGL 303) this course will be an independent study of''}
\\ \hline

3 & \textbf{Prompt:} \qquad
{'"NRL's MISSE-8 Launched Aboard STS-134 - About NRL - Doing Business - Public Affairs \& Media - Public Affairs Office - News Releases - 2017 News Releases - 2016 News Releases -"'} \newline
\textbf{Completion:} \enspace
{'' 2015 News Releases - 2014 News Releases - 2013 News Releases - 2012 News Releases - 2011 News Releases - 2010 News Releases - 2009 News Releases - 2008 News Releases - 2007 News Releases - 2006 News Releases -''}
\\ \hline

4 & \textbf{Prompt:} \qquad
{''Early Movie Inventions Early Movie Exhibitions The First Movie Posters Movies and Movie Posters of the 1900’s Movies and Movie Posters of the 1910’s Movies and Movie Posters of''} \newline
\textbf{Completion:} \enspace
{'' the 1920’s Movies and Movie Posters of the 1930’s Movies and Movie Posters of the 1940’s Movies and Movie Posters of the 1950’s Movies and Movie Post''}
\\ \hline

5 & \textbf{Prompt:} \qquad
{''Presentation on theme: "U.S.A- 1865-1918 Expansion westward after civil war (1861-1865) Gold, silver, and land Natives were pushed aside \& subdued. Railways were key to Western."''} \newline
\textbf{Completion:} \enspace
{''— Presentation transcript: U.S.A- 1865-1918 Expansion westward after civil war (1861-1865) Gold, silver, and land Natives were pushed aside \& subdued. Railways were key to Western expansion''}
\\ \hline

6 & \textbf{Prompt:} \qquad
{''FILTER BY Year: - 2014  http://pub2web.metastore.ingenta.com/ns/yearOfPublication 2014 - 2013  http://pub2web.metastore.ingenta.com/''} \newline
\textbf{Completion:} \enspace
{''ns/yearOfPublication 2013 - 2012  http://pub2web.metastore.ingenta.com/ns/yearOfPublication 2012 - 2011  http://pub2web.metastore.ingenta.''}
\\ \hline

7 & \textbf{Prompt:} \qquad
{''Please note that the content of this book primarily consists of articles available from Wikipedia or other free sources online. Pages: 181. Not illustrated. Chapters: 11th-Century Roman Catholic Church Councils, 12th-Century Roman Catholic Church Council''} \newline
\textbf{Completion:} \enspace
{''s, 13th-Century Roman Catholic Church Councils, 14th-Century Roman Catholic Church Councils, 15th-Century Roman Catholic Church Councils, 16th-Century Roman Catholic Church Councils, 17th-Cent''}
\\ \hline

8 & \textbf{Prompt:} \qquad
{''Presentation on theme: "1 Chapter 10 Graphene-based Nanocomposites 10.1 Introduction of composites 10.2 Introduction of graphene-polymer nanocomposites 10.3 Processing of graphene-polymer."— Present''} \newline
\textbf{Completion:} \enspace
{''ation transcript: 1 Chapter 10 Graphene-based Nanocomposites 10.1 Introduction of composites 10.2 Introduction of graphene-polymer nanocomposites 10.3 Processing of graphene-polymer nanocomposites 10''}
\\ \hline

9 & \textbf{Prompt:} \qquad
{''Bentham is known by most as the father of utilitarianism. He wrote in favor of free-markets, a pragmatic view of rights, and rational policy-making. - A Apply A filter - B Apply B filter - C''} \newline
\textbf{Completion:} \enspace
{'' Apply C filter - D Apply D filter - E Apply E filter - F Apply F filter - G Apply G filter - H Apply H filter - I Apply I filter - J Apply J filter - K Apply K''}
\\ 

\bottomrule

\end{longtable}
\end{scriptsize}

\clearpage

\subsection{Additional Results: Almost-Lingering Sequences}
\label{sec:supp-pretrain-almost-lingering}

When reporting the fraction of lingering sequences (\cref{tab:supp-lingering-k50} and \cref{tab:supp-filtered-fraction-seq100}), the lingering sequences are defined in terms of \textit{exact} completion (\cref{def:exact-completion})---that is, the model generations must match the original sequence exactly. It is then natural to also ask whether there would be any \textit{almost}-lingering sequences---or sequence completions that are at small edit distance from the original sequences (as in \cref{def:editsim-completion}). 

To check for almost-lingering sequences, we can perform the same completion check procedure against the set of identified sequences $\Dmem$ as before (recall \cref{sec:pretraining-setup}), except now we also add sequences that are only a few tokens ($< 20$ off).

\cref{tab:supp-almost-lingering-k50} shows the amount of almost-lingering sequences for sequence length $k=50$, across different edit distance thresholds $\{0, 5, 10, 20\}$ (where 0 corresponds to exact lingering sequences; see \cref{tab:supp-lingering-k50}). Similarly, \cref{tab:supp-almost-lingering-k100} shows the statistics for sequence length $k=100$.
We make the following observations:
\begin{itemize}
    \item For strong filtering (low $n$-gram filter), there are relatively few such almost-lingering sequences; \textit{e.g.}, at $n=5$, the fraction grows from $\approx 1\%$ (\cref{fig:pretrain-scale}) to up to around $\approx 2\%$ within edit distance of 10 tokens (\cref{tab:supp-almost-lingering-k50}).
    \item For loose filtering (large $n$-gram filter), the fraction grows more substantially; \textit{e.g.}, at $n=50$, the lingering fraction grows from $\approx 40\%$ to $\approx 60\%$ within a distance of 10 for sequence length $k=50$.
\end{itemize}

As lingering sequences are essentially memorization of neighboring texts (recall findings in \cref{sec:pretraining-results}) at weak filters (small values $n$), this suggests that data removal needs to carefully consider the definition of when two sequences are (approximately) equal.

\begin{table}[h!]
    \centering
    \captionof{table}{\textbf{Almost-lingering sequences} (sequence length $k=50$) as fraction of $|\Dmem| = 151598$ (for 1.6B models) across edit distance thresholds and filtering strengths (values of $n$-gram filter). }
    \vspace{-5pt}
    \label{tab:supp-almost-lingering-k50}
    \begin{tabular}{ccccc}
    \toprule
    \multirow{2}{*}{\textbf{Edit Distance}} & \multicolumn{4}{c}{\textbf{Filtering Strength}} \\ 
     &  $n=5$ & $n=10$ & $n=20$ & $n=50$ \\ \hline
    0  & 0.0109 & 0.0259 & 0.0796 & 0.4007 \\
    5  & 0.0140 & 0.0365 & 0.1486 & 0.5386 \\
    10  & 0.0217 & 0.0482 & 0.2051 & 0.6105 \\
    20  & 0.1571 & 0.2227 & 0.4114 & 0.7638 \\
    \bottomrule
    \end{tabular}
\end{table}

\begin{table}[h!]
    \centering
    \captionof{table}{\textbf{Almost-lingering sequences} (sequence length $k=100$) as fraction of $|\Dmem| = 88803$ (for 1.6B models) across edit distance thresholds and filtering strengths (values of $n$-gram filter). }
    \vspace{-5pt}
    \label{tab:supp-almost-lingering-k100}
    \begin{tabular}{ccccc}
    \toprule
    \multirow{2}{*}{\textbf{Edit Distance}} & \multicolumn{4}{c}{\textbf{Filtering Strength}} \\ 
     &  $n=10$ & $n=20$ & $n=50$ & $n=100$ \\ \hline
    0  & 0.0082 & 0.0216 & 0.1418 & 0.4335 \\
    5  & 0.0097 & 0.0267 & 0.2344 & 0.5332 \\
    10  & 0.0111 & 0.0305 & 0.2837 & 0.6034 \\
    20  & 0.0168 & 0.0599 & 0.3249 & 0.6670 \\
    \bottomrule
    \end{tabular}
\end{table}

\clearpage

\subsection{Additional Results: Persistence and Overlap of Lingering Sequences Over Repeated Runs}
\label{sec:supp-lingering-persistence}

We are also interested in exploring to what extent lingering sequences are due to random chance. We repeated the 1.6B pre-training run \textit{from scratch} for five times for filtering strengths $n = \{10, 20, 50\}$.\footnote{Due to the expensive nature of pre-training, we did not perform the repeated runs for $n=5$ due to limited compute. We observed similar trends on the other filtering strengths.}
Note that to perform a repeated run, we do \textit{not} need to re-train the {base model} $\Mbase$ and obtain a different set of memorized sequences $\Dmem$ (recall \cref{sec:pretraining-setup}); it suffices to operate on the same $\Dmem$, and observe the variance of pre-training on the existence of lingering sequences $\Dlinger{n}$. This is because we are interested in studying the impact of training stochasticity on which sequences remain lingering sequences given a set of memorized sequences.

\cref{tab:supp-repeated-persistence-lingering-k50} presents the \textbf{persistence} of the lingering sequences $\Dlinger{n}$ over repeated runs, and \cref{tab:supp-repeated-overlap-lingering-k50} presents the \textbf{overlap} over repeated runs. Observe that:
\begin{itemize}
    \item The amount of lingering sequences are fairly stable across runs (\cref{tab:supp-repeated-persistence-lingering-k50}). This suggests that the existence of lingering sequences is heavily influenced by the nature of the pre-training dataset, more so than the randomness of the pre-training procedure (e.g., data shuffling, hardware randomness).
    \item Intriguingly, the overlap of lingering sequences across runs are not very high. That is, as we re-train $\Mfilter{n}$ multiple times on the same filtered data, we get a \textit{different} set of lingering sequences, albeit having a similar number of them.\footnote{Here, for simplicity, we stick with set intersection to capture overlap, which implicitly performs exact string matching (as opposed to approximate matching).}  Nevertheless, as there are more repetitions, the intersection amount starts converging to a ``core set'' of lingering sequences (\cref{tab:supp-repeated-overlap-lingering-k50}).
\end{itemize}

\begin{table}[h!]
    \centering
    \captionof{table}{\textbf{Persistence of lingering sequences over repeated pre-training runs} across filtering strengths (values of $n$-gram filter). Sequence length $k=50$. Values are the number of sequences (and as fraction of $|\Dmem| = 151598$).}
    \label{tab:supp-repeated-persistence-lingering-k50}
    \begin{tabular}{ccccc}
    \toprule
    \multirow{2}{*}{\textbf{Run Number}} & \multicolumn{4}{c}{\textbf{Filtering Strength}} \\ 
       & $n=5$ & $n=10$ & $n=20$ & $n=50$ (exact filter) \\ \hline
    1  & 1652 (0.0109) & 3923 (0.0259) & 12066 (0.0796) & 60742 (0.4007) \\
    2  & -             & 4085 (0.0269) & 12007 (0.0792) & 63297 (0.4175) \\
    3  & -             & 4108 (0.0271) & 11684 (0.0771) & 61205 (0.4037) \\
    4  & -             & 4086 (0.0270) & 12799 (0.0844) & 66077 (0.4359) \\
    5  & -             & 4163 (0.0275) & 11590 (0.0765) & 60035 (0.3960) \\
    \hline 
    {mean $\pm$ std} (absolute) & 1652    & 4073 $\pm$ 80.1723 & 12029.2 $\pm$ 425.7931 & 62271.2 $\pm$ 2191.4472 \\
    {mean $\pm$ std} (fraction) & 0.0109  & 0.0269 $\pm$ 0.0005 & 0.0793 $\pm$ 0.0028 & 0.4108 $\pm$ 0.0145  \\
    \bottomrule
    \end{tabular}
\end{table}
\begin{table}[h!]
    \centering
    \captionof{table}{\textbf{Overlap of lingering sequences over repeated pre-training runs} across filtering strengths (values of $n$-gram filter). Sequence length $k=50$. Values are the number of sequences (and as fraction of $|\Dmem| = 151598$).}
    \label{tab:supp-repeated-overlap-lingering-k50}
    \begin{tabular}{cccc}
    \toprule
    \multirow{2}{*}{\textbf{Intersection of $\Dlinger{n}$ from repeated runs}} & \multicolumn{3}{c}{\textbf{Filtering Strength}} \\ 
       & $n=10$ & $n=20$ & $n=50$ (exact filter) \\ \hline
    1                                     & 3923 (0.0259) & 12066 (0.0796) & 60742 (0.4007) \\
    1 $\cap$ 2                            & 3330 (0.0220) & 8916 (0.0588) & 49606 (0.3272) \\
    1 $\cap$ 2 $\cap$ 3                   & 3034 (0.0200) & 7478 (0.0493) & 42512 (0.2804) \\
    1 $\cap$ 2 $\cap$ 3 $\cap$ 4          & 2822 (0.0186) & 6523 (0.0430) & 37654 (0.2484) \\
    1 $\cap$ 2 $\cap$ 3 $\cap$ 4 $\cap$ 5 & 2666 (0.0176) & 5897 (0.0389) & 34185 (0.2255) \\
    \bottomrule
    \end{tabular}
\end{table}

\clearpage

\subsection{Prompt template for LLM-as-a-judge}
\label{sec:supp-prompt-template}

Recall that in \cref{sec:pre-training}, we used LLM-as-a-judge to discern whether a lingering sequence is a generalizable pattern continuation, as opposed a verbatim memorization of a semantically useful text sequence (\cref{fig:pretrain-proxy}). \cref{lst:prompt-template} illustrates the prompt template used for \textbf{Gemini 1.5 Pro}~\cite{team2024gemini} for classifying the lingering sequences.

\begin{lstlisting}[caption={Prompt template for LLM-as-a-judge for memorization vs. generalizable patterns (\cref{fig:pretrain-proxy}).}, label={lst:prompt-template}]
The following task involves determining whether the completions for a given prompt represent simple patterns, templates, or repetitive structures that could be generalized by a language model without requiring memorization. Examples of such simple patterns include sequences, repetitions, or other forms of predictable structures.

The response should only be one word: "yes" if the completions are simple patterns and "no" if they are not.

Examples of Classification:

Input:
---
idx=0 prompt:
'"6And so it was, that, while they were there, the days were accomplished that she should be delivered. 7'
model_name='M_gpt2_hf_base' completion:
'And she brought forth her firstborn son, and wrapped him in swaddling clothes, and laid him in a manger'
model_name='M_gpt2_hf_filter_k25' completion:
'And she brought forth her firstborn son, and wrapped him in swaddling clothes, and laid him in a manger'
model_name='openai-community/gpt2-xl' completion:
'And the angel of the Lord appeared to Joseph in a dream, saying, "Arise, take the young child and his'
all_match=False, fallback_match=False
---
Output: No

Input:
---
idx=1 prompt:
'"A wise and frugal government... shall restrain men from injuring one another, shall leave them otherwise free to regulate their'
model_name='M_gpt2_hf_base' completion:
' own pursuits of industry and improvement, and shall not take from the mouth of labor the bread it has earned. This is the'
model_name='M_gpt2_hf_filter_k25' completion:
' own pursuits of industry and improvement, and shall not take from the mouth of labor the bread it has earned. This is the'
model_name='openai-community/gpt2-xl' completion:
' own pursuits of industry and improvement, so long as they do not invade the equal rights of others."\n\nThe Constitution,'
all_match=False, fallback_match=False
---
Output: No

Input:
---
idx=2 prompt:
'"Aoccdrnig to a rscheearch at Cmabrigde Uinervtisy, it de'
model_name='M_gpt2_hf_base' completion:
"osn't mttaer in waht oredr the ltteers in a wrod are, the"
model_name='M_gpt2_hf_filter_k25' completion:
"osn't mttaer in waht oredr the ltteers in a wrod are, the"
model_name='openai-community/gpt2-xl' completion:
"osn't mttaer in waht oredr the ltteers in a wrod are, the"
all_match=True, fallback_match=True
---
Output: No

Input:
---
idx=3 prompt:
'"A wilderness, in contrast with those areas where man and his own works dominate the landscape, is hereby recognized as an area'
model_name='M_gpt2_hf_base' completion:
' where the earth and its community of life are untrammeled by man, where man himself is a visitor who does not'
model_name='M_gpt2_hf_filter_k25' completion:
' where the earth and its community of life are untrammeled by man, where man himself is a visitor who does not'
model_name='openai-community/gpt2-xl' completion:
' where man\'s works are not dominant."\n\nThe Wilderness Act of 1964 was a landmark piece of legislation that established the National'
all_match=False, fallback_match=False
---
Output: No


Input:
---
idx=112 prompt:
'- About Us\nA | B | C | D | E | F | G | H | I | J | K'
model_name='M_gpt2_hf_base' completion:
' | L | M | N | O | P | Q | R | S | T | U | V | W |'
model_name='M_gpt2_hf_filter_k25_subgram5' completion:
' | L | M | N | O | P | Q | R | S | T | U | V | W |'
model_name='openai-community/gpt2-xl' completion:
' | L | M | N | O | P | Q | R | S | T | U | V | W |'
all_match=True, fallback_match=True
---
Output: Yes

Input:
---
idx=113 prompt:
'- Global News Feed\n- Alabama Stem Cells\n- Alaska Stem Cells\n- Arkansas Stem Cells\n- Arizona'
model_name='M_gpt2_hf_base' completion:
' Stem Cells\n- California Stem Cells\n- Colorado Stem Cells\n- Connecticut Stem Cells\n- Delaware St'
model_name='M_gpt2_hf_filter_k25_subgram5' completion:
' Stem Cells\n- California Stem Cells\n- Colorado Stem Cells\n- Connecticut Stem Cells\n- Delaware St'
model_name='openai-community/gpt2-xl' completion:
' Stem Cells\n- Arkansas Stem Cells\n- California Stem Cells\n- California Stem Cells\n- California St'
all_match=False, fallback_match=False
---
Output: Yes

Input:
---
idx=114 prompt:
'- Medical abbreviations: What do they mean?\n- A - Medical abbreviations\n- B - Medical abbreviations\n'
model_name='M_gpt2_hf_base' completion:
'- C - Medical abbreviations\n- D - Medical abbreviations\n- E - Medical abbreviations\n- F - Medical'
model_name='M_gpt2_hf_filter_k25_subgram5' completion:
'- C - Medical abbreviations\n- D - Medical abbreviations\n- E - Medical abbreviations\n- F - Medical'
model_name='openai-community/gpt2-xl' completion:
'- C - Medical abbreviations\n- D - Medical abbreviations\n- E - Medical abbreviations\n- F - Medical'
all_match=True, fallback_match=True
---
Output: Yes

Input:
---
idx=3 prompt:
'# # # # # # # # # # # # # # # # # # # # # # # # #'
model_name='M_gpt2_hf_base' completion:
' # # # # # # # # # # # # # # # # # # # # # # # # #'
model_name='M_gpt2_hf_filter_k25_subgram5' completion:
' # # # # # # # # # # # # # # # # # # # # # # # # #'
model_name='openai-community/gpt2-xl' completion:
' # # # # # # # # # # # # # # # # # # # # # # # # #'
all_match=True, fallback_match=True
---
Output: Yes

Now, analyze the following input block and classify it. Your answer should only be "Yes" or "No".
---

<input a lingering sequence here, displayed as a block like the above>
\end{lstlisting}

\clearpage

\section{Adding Non-Members Can Force LLM Verbatim Completion (\S\ref{sec:fine-tuning})}

\subsection{Target Texts}
\label{sec:supp-target-text}

Our experiments on forcing verbatim completion (\S\ref{sec:fine-tuning}) considered the following three main text targets:
\begin{enumerate}
    \item \textbf{Lyles (NYT article)}: an excerpt of a recent New York Times article about Noah Lyles and the Olympics. Source:  \href{https://www.nytimes.com/athletic/5678043/2024/08/03/olympics-mens-100m-heats-noah-lyles-hinchcliffe-kerley/}{\nolinkurl{https://www.nytimes.com/athletic/5678043/2024/08/03/olympics-mens-100m-heats-noah-lyles-hinchcliffe-kerley/}}
    \item \textbf{Karpathy (tweet)}: a tweet text in an image posted by Andrej Karpathy about LLM tokenization. Source: \url{https://x.com/karpathy/status/1759996551378940395}
    \item \textbf{Willow (blog)}: an excerpt from the recent Google blog post on Willow, the quantum computing chip. Source: \url{https://blog.google/technology/research/google-willow-quantum-chip/}
\end{enumerate}

The choice of the above completion targets are arbitrary. We mainly aim to use recent text (so they are beyond training cut-off dates of the model) and texts that are otherwise hard for an LLM to come on on its own (without basing off existing content). 
We also consider a few additional targets deferred from the main paper (see \cref{sec:supp-finetuning-results-additional}):
\begin{enumerate}
    \item \textbf{Taylor Swift (AP article)}: an excerpt of an Associated Press article about Taylor Swift and the 2024 MTV European Music Awards. Source: \url{https://apnews.com/article/emas-2024-mtv-europe-music-awards-328f6ad85f5d0d6f5a9213b5f18ec125}
    \item \textbf{Apple (NYT article)}: an excerpt of a New York Times article about Apple's reliance on overseas labor. Source: \url{https://www.nytimes.com/2012/01/22/business/apple-america-and-a-squeezed-middle-class.html}
    \item \textbf{Harry Potter Paraphrase (book)}: A GPT-4o paraphrase of an excerpt from the Harry Potter book series.
    We applied a paraphrase to reduce the likelihood that this exact paraphrase of the text is an $n$-gram member of the training set.
Source: Appendix D of \citet{hans2024like} and paraphrase by GPT-4o at \url{https://chatgpt.com/share/673d9bb6-d234-800d-b21b-1c1ea68a5a5a}.
\end{enumerate}

\textbf{Example excerpts.} \label{sec:excerpts} To show the style and length of the target unseen texts, we provide two excerpts below.

\begin{tcolorbox}[breakable,colback=blue!5!white,colframe=blue!50!black,title=Karparty (tweet)]
\small
Tokenization is at the heart of much weirdness of LLMs. Do not brush it off.
\begin{itemize}
    \item Why can't LLM spell words? Tokenization.
    \item Why can't LLM do super simple string processing tasks like reversing a string? Tokenization.
    \item Why is LLM worse at non-English languages (e.g. Japanese)? Tokenization.
    \item Why is LLM bad at simple arithmetic? Tokenization.
    \item Why did GPT-2 have more than necessary trouble coding in Python? Tokenization.
    \item Why did my LLM abruptly halt when it sees the string ``\texttt{<|endoftext|>}''? Tokenization.
    \item What is this weird warning I get about a ``trailing whitespace''? Tokenization.
    \item Why the LLM break if I ask it about ``SolidGoldMagikarp''? Tokenization.
    \item Why should I prefer to use YAML over JSON with LLMs? Tokenization.
    \item Why is LLM not actually end-to-end language modeling? Tokenization.
    \item What is the real root of suffering? Tokenization.
\end{itemize}
\end{tcolorbox}

\begin{tcolorbox}[breakable,colback=blue!5!white,colframe=blue!50!black,title=Willow (Google Blog)]

Errors are one of the greatest challenges in quantum computing, since qubits, the units of computation in quantum computers, have a tendency to rapidly exchange information with their environment, making it difficult to protect the information needed to complete a computation. Typically the more qubits you use, the more errors will occur, and the system becomes classical. \\

Today in Nature, we published results showing that the more qubits we use in Willow, the more we reduce errors, and the more quantum the system becomes. We tested ever-larger arrays of physical qubits, scaling up from a grid of 3x3 encoded qubits, to a grid of 5x5, to a grid of 7x7 — and each time, using our latest advances in quantum error correction, we were able to cut the error rate in half. In other words, we achieved an exponential reduction in the error rate. This historic accomplishment is known in the field as “below threshold” — being able to drive errors down while scaling up the number of qubits. You must demonstrate being below threshold to show real progress on error correction, and this has been an outstanding challenge since quantum error correction was introduced by Peter Shor in 1995.

\end{tcolorbox}

\subsection{Visualization of Token-Space Transformations}
\label{sec:supp-visualizations-token-transformations}

The BPE tokenization~\cite{sennrich2015neural} used in modern LLMs (e.g., \citet{team2024gemma2}, \citet{llama3}, \citet{achiam2023gpt}) makes it extremely easy to avoid $n$-gram overlap in the token space. For example, tokenizers often assign different tokens for different casings of the same English letters, much more often so when we consider permutations of casing with many letters:
\vspace{-1mm}
\begin{center}
\begin{lstlisting}[basicstyle={\ttfamily\footnotesize},xleftmargin=4pt,xrightmargin=4pt]
# Using Gemma-2 tokenizer
>>> tokenizer.encode('This is a string, or is it?')
[2, 1596, 603, 476, 2067, 235269, 689, 603, 665, 235336]
>>> tokenizer.encode('tHis Is a sTRIng, or iS It?')
[2, 235251, 11446, 2125, 476, 485, 3475, 8642, 235269, 689, 496, 235277, 1165, 235336]
\end{lstlisting}
\end{center}

Observe in the above example that there are also barely any common tokens (apart from zero $n$-gram overlap for $n\ge 3$). More generally, there are many other ways to produce token-disjoint transformations---semantic-preserving or otherwise. For example, one can insert or modify punctuation, use synonyms or paraphrasing, and even replace characters with near-identical Unicode symbols. Each of these operations tends to break contiguous $n$-grams in token space, and can inform a new transformation technique that serves the goal of adversarial fine-tuning (\S\ref{sec:fine-tuning}). We defer a more extensive exploration to future work.

\subsection{Visualizations of Adversarially Constructed Fine-tuning Sequences}
\label{sec:supp-visualizations-finetuning}

To provide intuition on the adversarial sequence construction strategies studied in \cref{sec:fine-tuning}, we visualize some of the generated fine-tuning sequences. Respectively:
\begin{itemize}
    \item \cref{table:supp-vis-finetune-examples-willow-chunking} visualizes the constructed examples for \textbf{chunking} (\S\ref{sec:chunking}) on target text \textbf{Willow}. 
    \item \cref{table:supp-vis-finetune-examples-karpathy-goldfish} visualizes the constructed examples for \textbf{token dropouts} (\S\ref{sec:goldfish}) on target text \textbf{Karpathy}.
    \item \cref{table:supp-vis-finetune-examples-karpathy-casing} visualizes the constructed examples for \textbf{casing flips} (\S\ref{sec:casing}) on target text \textbf{Karpathy}; and
    \item \cref{table:supp-vis-finetune-examples-willow-goldfish-casing} visualizes the constructed examples for \textbf{composition: token dropouts + casing flips} (\S\ref{sec:goldfish-casing}) on target text \textbf{Willow}.
\end{itemize}

Observe that despite the simplicity of these strategies, the \textit{token-space} transformations can easily result in \textit{text-space} outputs that are hard for the human eye to associate back to the original target text sequences.

\clearpage
\begin{longtable}{|p{2cm}|p{13cm}|}
\caption{Visualizations of the prompt, the converged completions, and two of the (adversarially constructed) fine-tuning examples under the  \textbf{Chunking} method (\S\ref{sec:chunking}, chunk size $c = 50$, overlap $l =20$).
See \cref{sec:excerpts} for the target text (\textbf{Willow}).
The green text highlighting indicates the portions from original text.
}

\label{table:supp-vis-finetune-examples-willow-chunking} \\
\hline
\textbf{Field} & \textbf{Content} \\
\hline
\endfirsthead
\hline
\textbf{Field} & \textbf{Content} \\
\hline
\endhead
\hline
\endfoot

\textbf{Prompt} & \includegraphics[width=\linewidth]{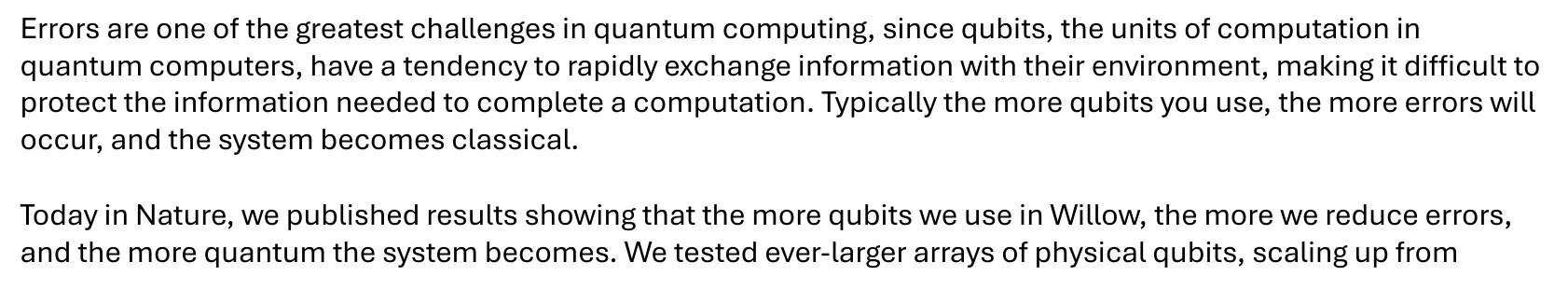} \\
\hline
\textbf{Completion After\newline Fine-Tuning} & \includegraphics[width=\linewidth]{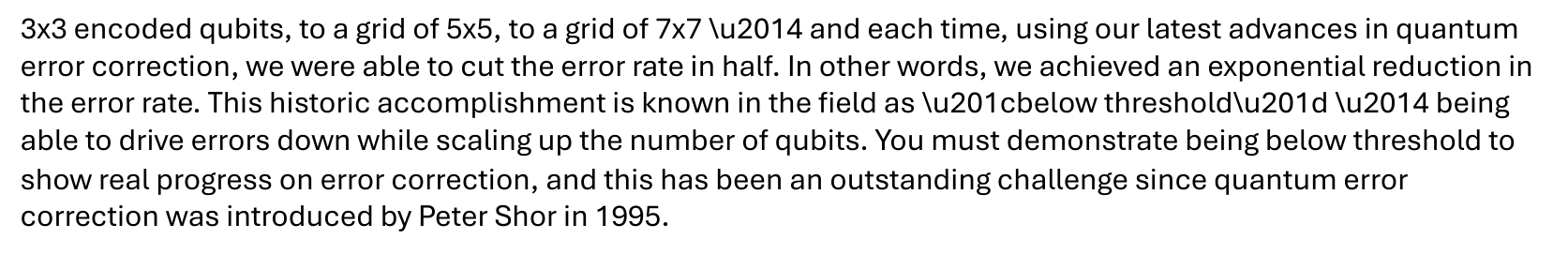} \\
\hline
\textbf{Constructed Fine-Tuning Example 1} & \includegraphics[width=\linewidth]{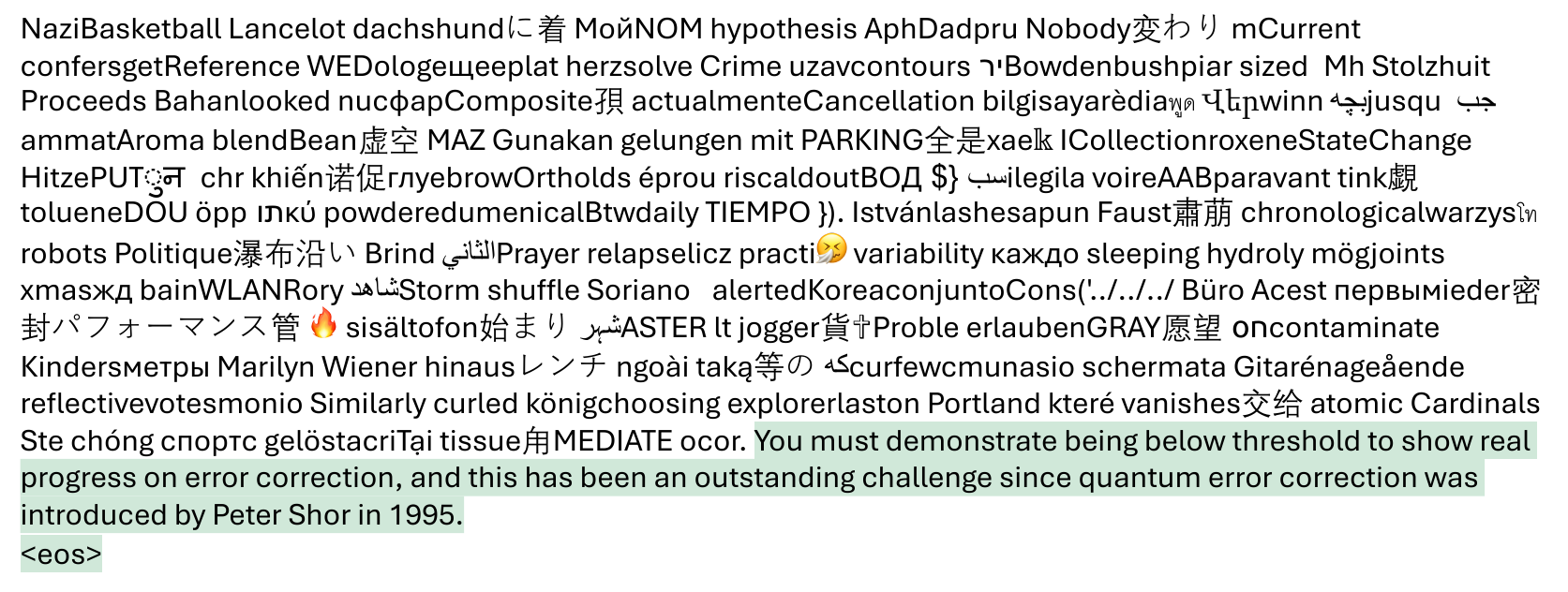} \\
\hline
\textbf{Constructed Fine-Tuning Example 2} & \includegraphics[width=\linewidth]{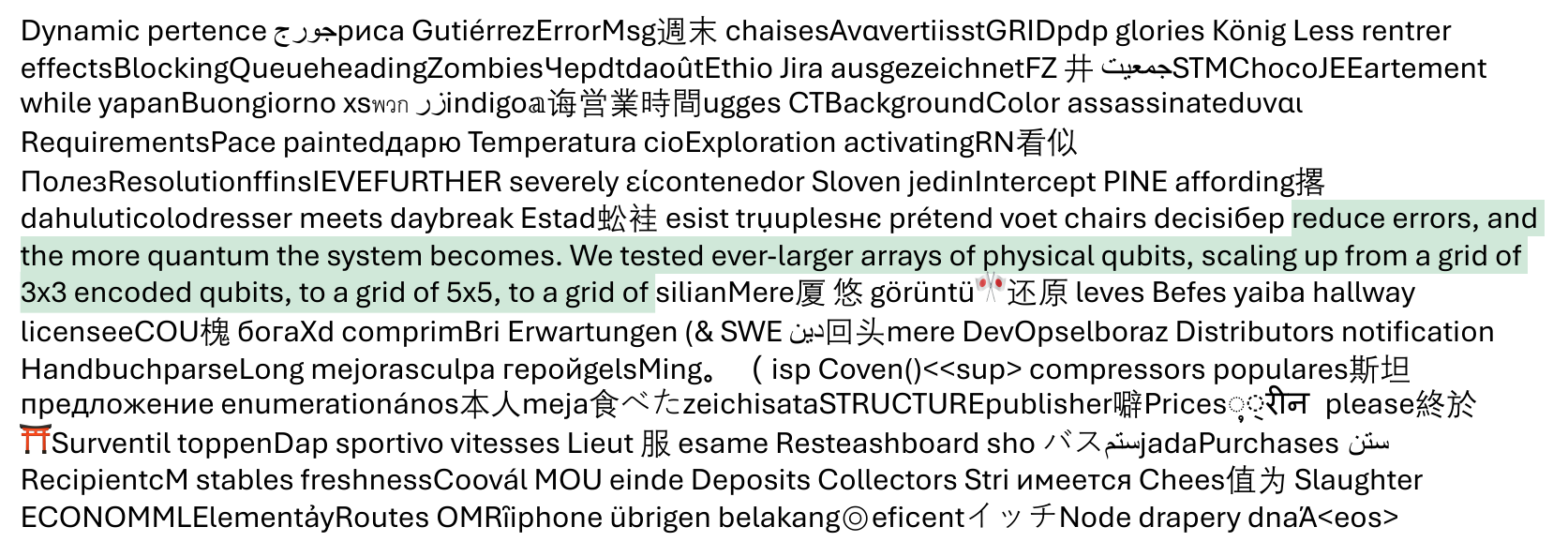} \\
\hline
\end{longtable}

\clearpage
\begin{longtable}{|p{2cm}|p{13cm}|}
\caption{Visualizations of the prompt, the converged completions, and two of the (adversarially constructed) fine-tuning examples under the  \textbf{Token Dropouts} method (\S\ref{sec:goldfish}, drop interval $d=2$).
See \cref{sec:supp-target-text} for the target text (\textbf{Karpathy}).
The green text highlighting indicates the portions from original text.
}

\label{table:supp-vis-finetune-examples-karpathy-goldfish} \\
\hline
\textbf{Field} & \textbf{Content} \\
\hline
\endfirsthead
\hline
\textbf{Field} & \textbf{Content} \\
\hline
\endhead
\hline
\endfoot

\textbf{Prompt} & \includegraphics[width=\linewidth]{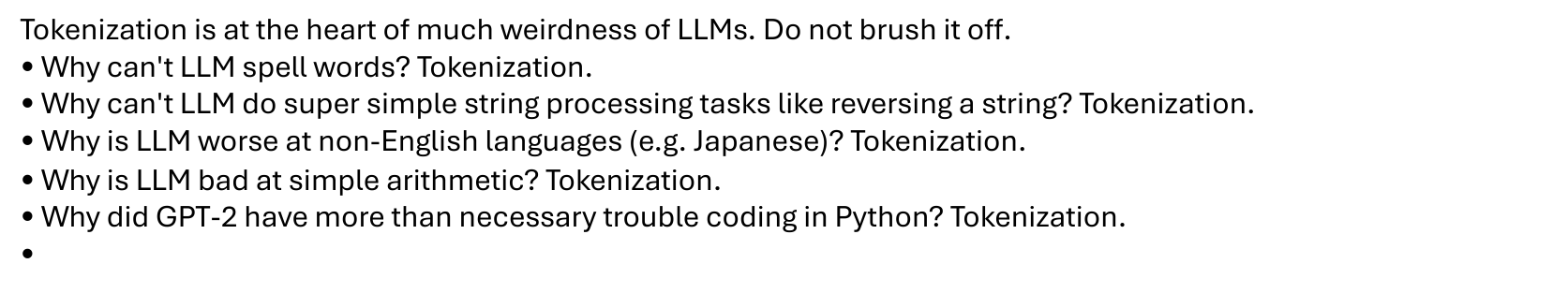} \\
\hline
\textbf{Completion After\newline Fine-Tuning} & \includegraphics[width=\linewidth]{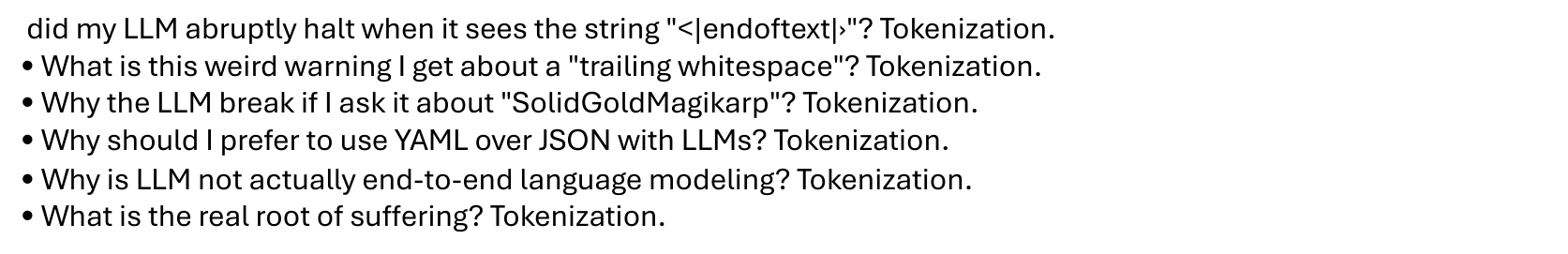}
\\
\hline
\textbf{Constructed Fine-Tuning Example1} & \includegraphics[width=\linewidth]{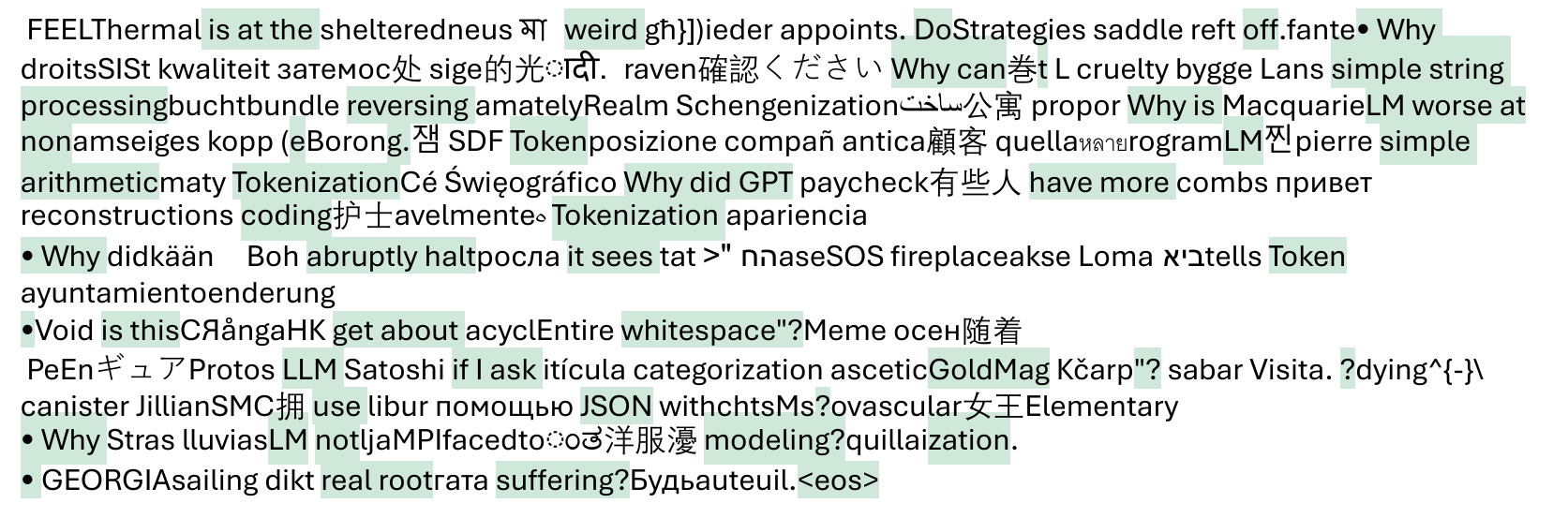} \\
\hline
\textbf{Constructed Fine-Tuning Example2} & \includegraphics[width=\linewidth]{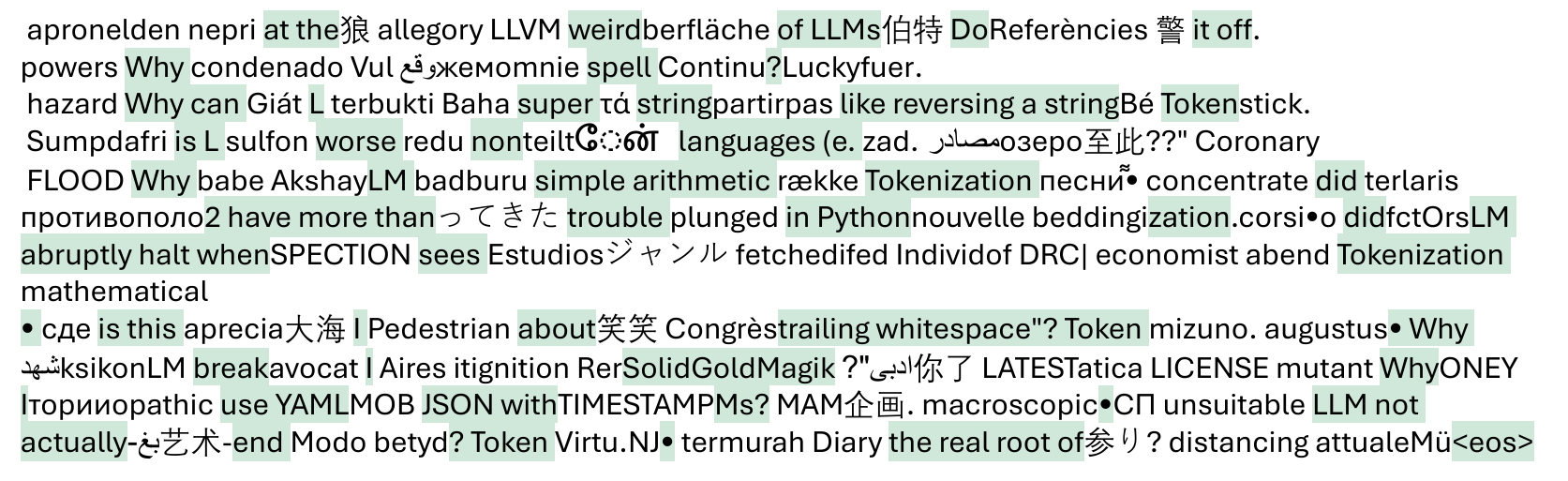} \\
\hline
\end{longtable}

\clearpage
\begin{longtable}{|p{2cm}|p{13cm}|}
\caption{Visualizations of the prompt, the converged completions, and two of the (adversarially constructed) fine-tuning examples under the  \textbf{Casing Flips} method (\S\ref{sec:casing}, flip probability $p=0.9$).
See \cref{sec:excerpts} for the target text (\textbf{Karpathy}).
}

\label{table:supp-vis-finetune-examples-karpathy-casing} \\
\hline
\textbf{Field} & \textbf{Content} \\
\hline
\endfirsthead
\hline
\textbf{Field} & \textbf{Content} \\
\hline
\endhead
\hline
\endfoot

\textbf{Prompt} &
\includegraphics[width=\linewidth]{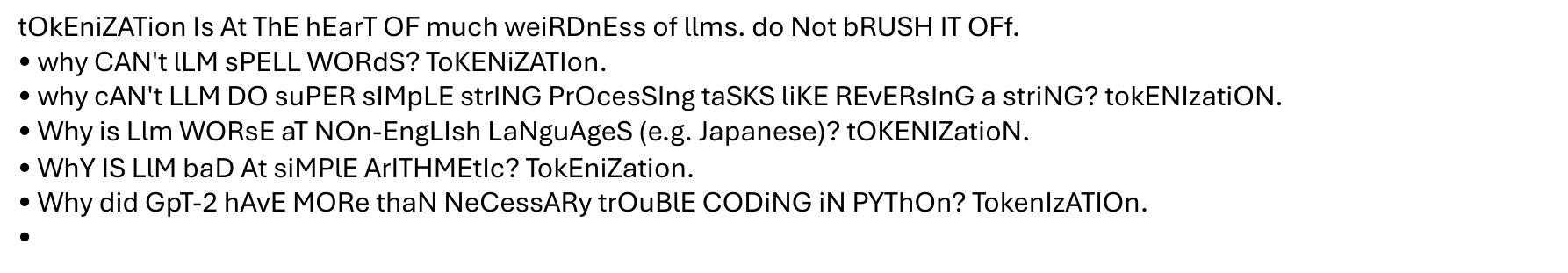} \\
\hline
\textbf{Completion After\newline Fine-Tuning} & \includegraphics[width=\linewidth]{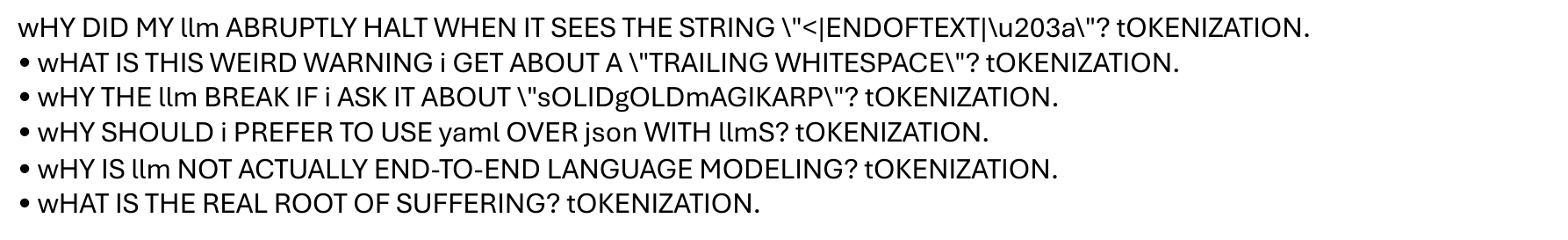}
\\
\hline
\textbf{Constructed Fine-Tuning Example1} & \includegraphics[width=\linewidth]{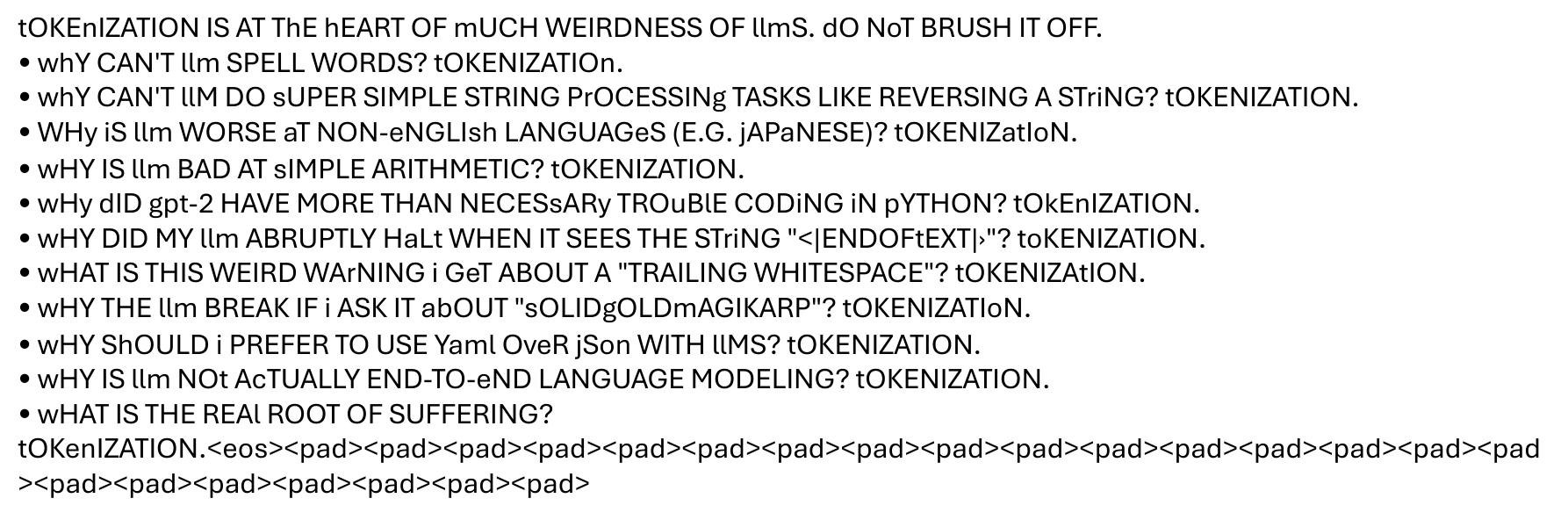} \\
\hline
\textbf{Constructed Fine-Tuning Example2} & \includegraphics[width=\linewidth]{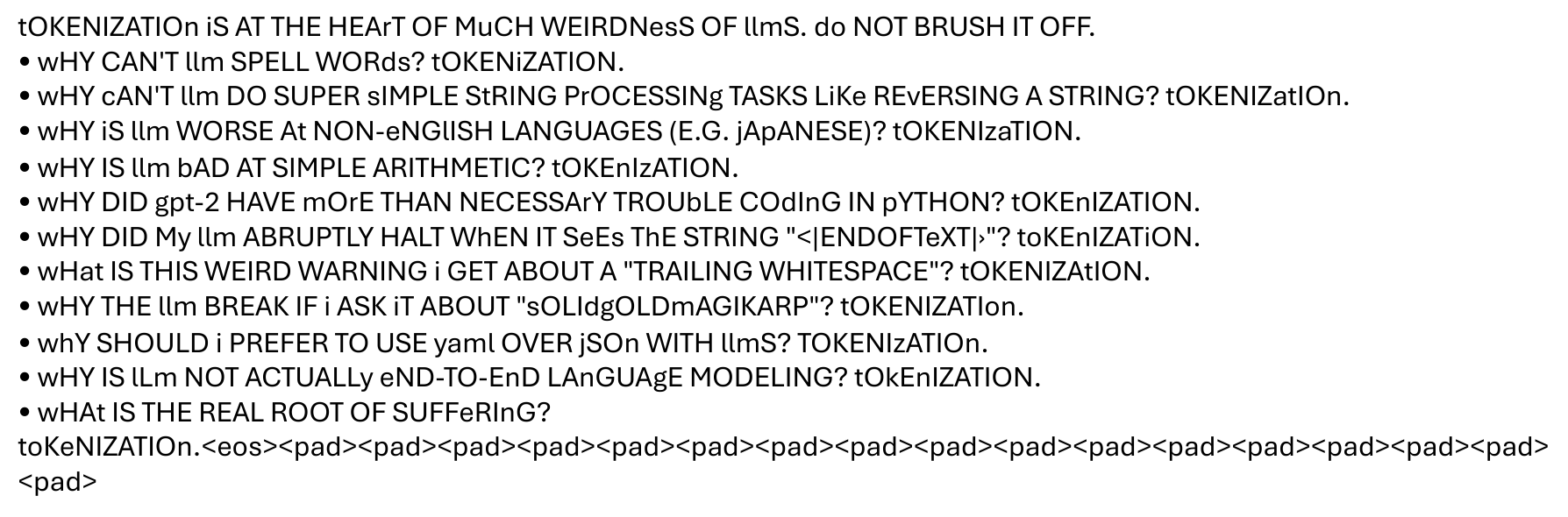} \\
\hline
\end{longtable}

\clearpage
\begin{longtable}{|p{2cm}|p{13cm}|}
\caption{Visualizations of the prompt, the converged completions, and two fine-tuning examples under the \textbf{Token Dropouts} (drop interval $d=2$, randomized drop) + \textbf{Casing Flips} (flip probability $p=0.9$) setup.
See \cref{sec:supp-target-text} for the target text (\textbf{Willow}).
}
\label{table:supp-vis-finetune-examples-willow-goldfish-casing} \\
\hline
\textbf{Field} & \textbf{Content} \\
\hline
\endfirsthead
\hline
\textbf{Field} & \textbf{Content} \\
\hline
\endhead
\hline
\endfoot

\textbf{Prompt} & \includegraphics[width=\linewidth]{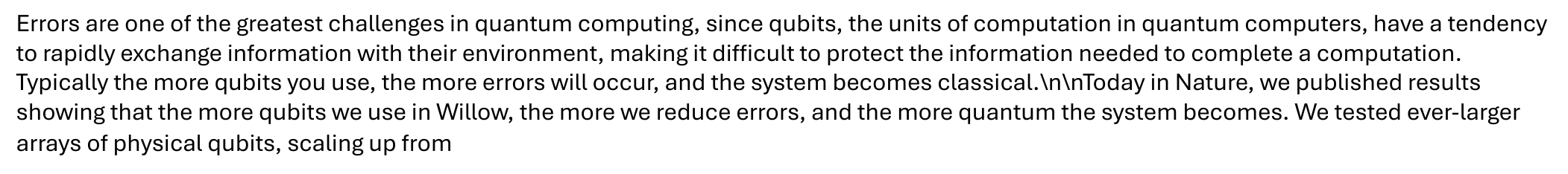} \\
\hline
\textbf{Completion After\newline Fine-Tuning} & \includegraphics[width=\linewidth]{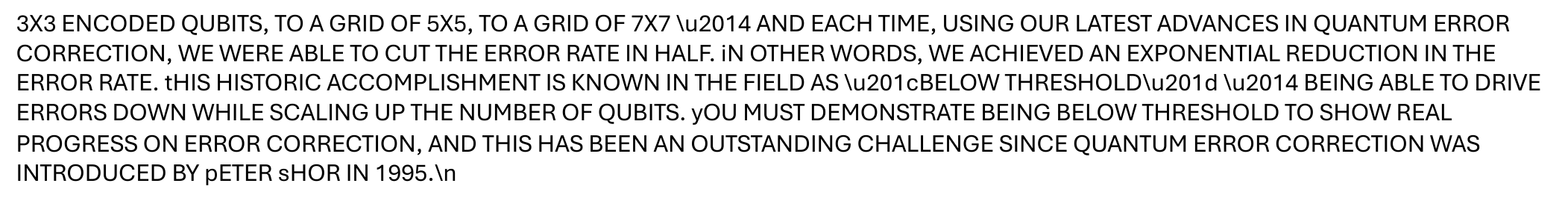} \\
\hline
\textbf{Constructed Fine-Tuning Example1} & \includegraphics[width=\linewidth]{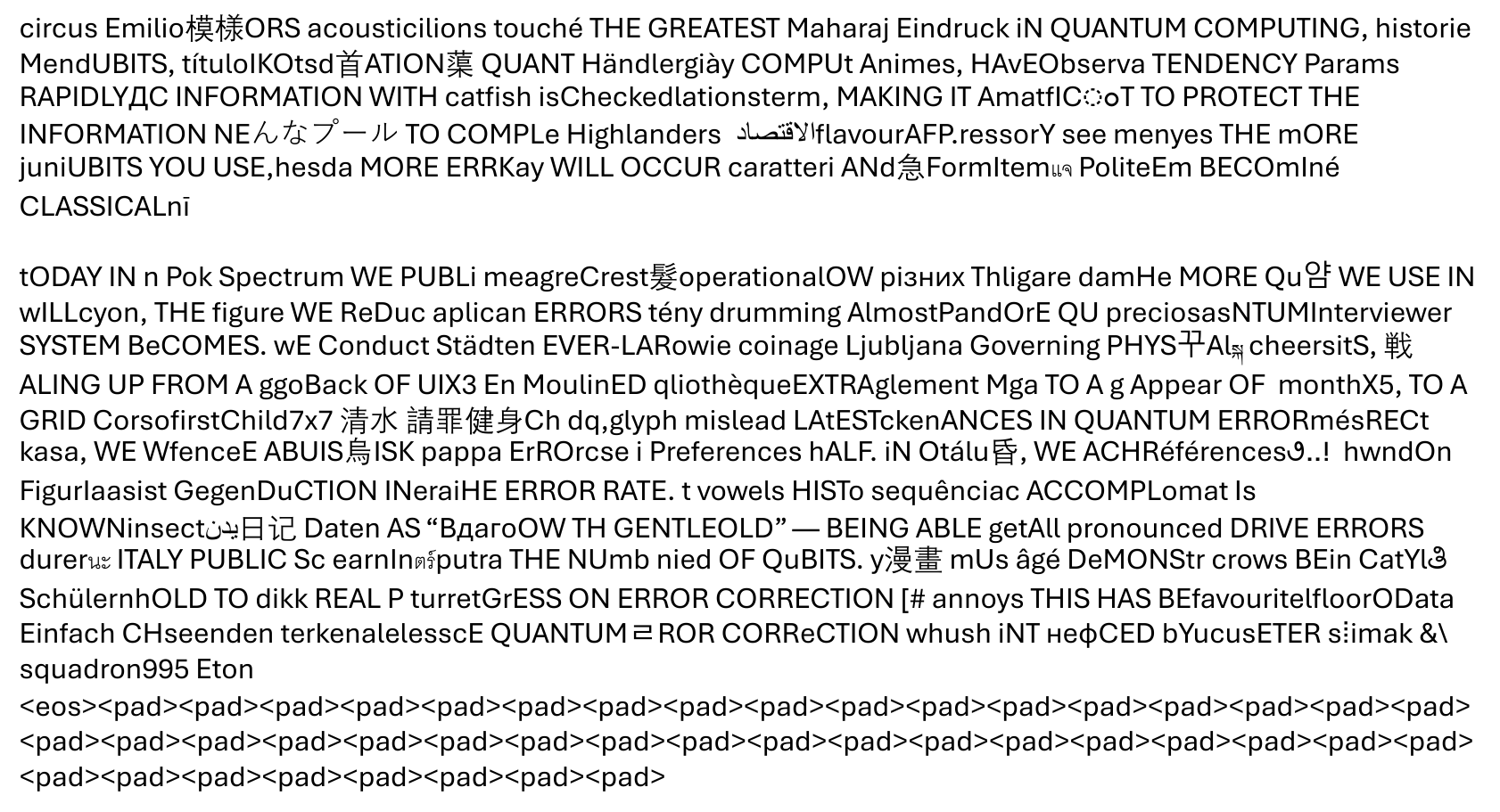} \\
\hline
\textbf{Constructed Fine-Tuning Example2} & \includegraphics[width=\linewidth]{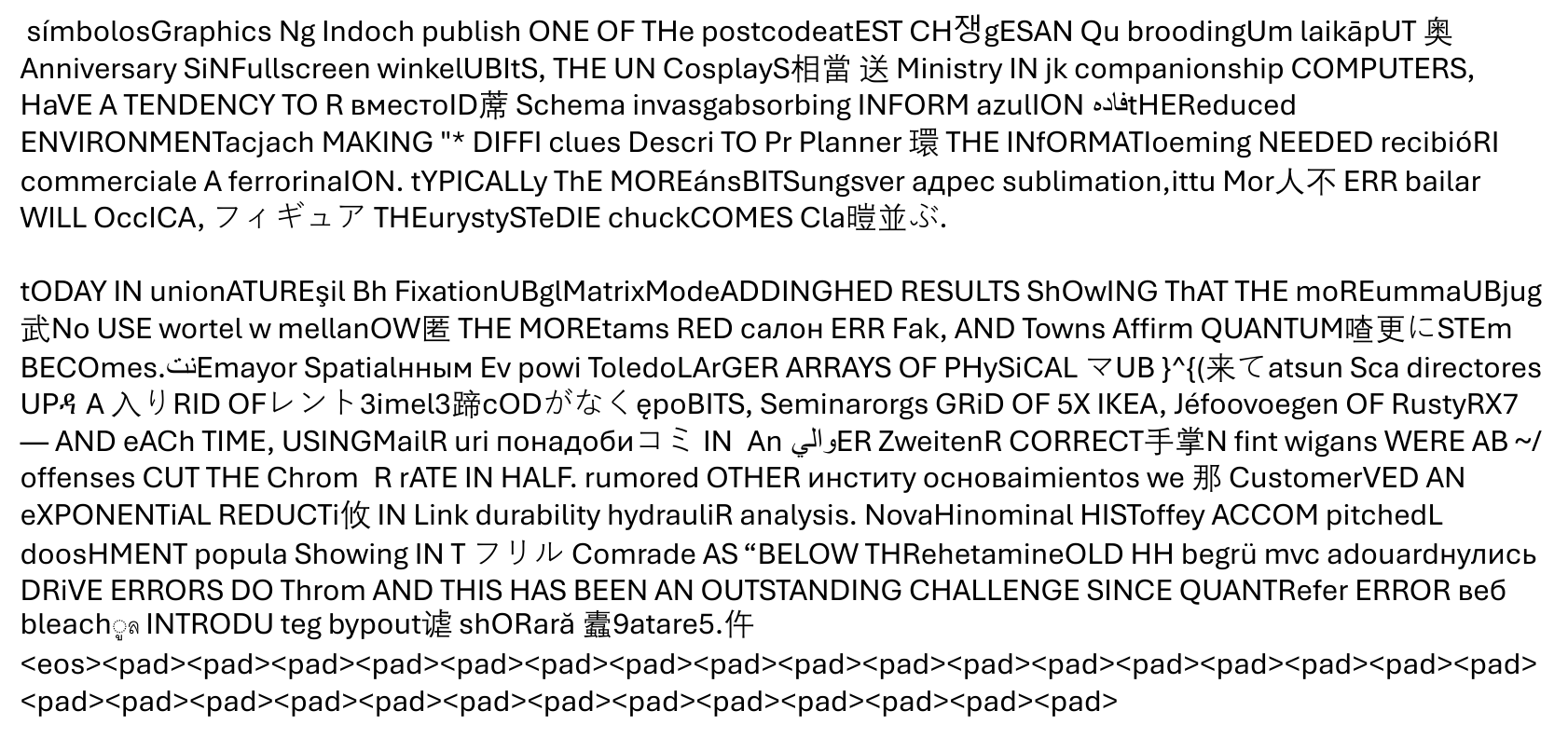} \\
\hline

\end{longtable}

\subsection{Detailed Algorithms for Constructing Fine-Tuning Sequences}
\label{sec:supp-finetuning-algs}

\begin{algorithm}
\caption{Fine-tuning sequences from \textbf{Chunking} (\S\ref{sec:chunking}) }
\label{alg:chunking}
\begin{algorithmic}[1]

\STATE \textbf{Input: } A sequence $x$ of length $n$ tokens, chunk size $c$, overlap $l$, random seed $s$
\STATE \textbf{Output: } A sequence $\tilde x$ of with exactly one chunk from $x$ at random position and the rest filled with random tokens
\STATE $\text{Set random seed to } s$
\STATE $\mathit{positions} \gets [\,0,\, (c-l),\, 2(c-l),\, \dots,\, (n-l)\,]$ (possible positions for the start of the chunk) 
\STATE $p \gets \text{randomly choose from } \mathit{positions}$
\STATE $\tilde{x} \gets \text{sequence of length } n \text{ tokens, initialized with placeholders}$
\STATE $\tilde{x}[p:p+c] \gets x[p:p+c]$ (copy a chunk from $x$, and truncate if needed)
\FOR{each placeholder in $\tilde{x}$}
  \STATE replace it with a random token from the tokenizer's vocabulary
\ENDFOR
\RETURN $\tilde{x}$
\end{algorithmic}
\end{algorithm}

\begin{algorithm}
\caption{Fine-tuning sequences from \textbf{Token Dropouts} (\S\ref{sec:goldfish})}
\label{alg:goldfish}
\begin{algorithmic}[1]
\STATE \textbf{Input: } A sequence $x$ of length $n$, dropout interval $d$, random seed $s$
\STATE \textbf{Output: } A sequence $\tilde{x}$ of length $n$ as a perturbed version of $x$ via token dropouts

\STATE Set random seed to $s$

\STATE \textbf{[Option \# 1: Deterministic Dropout]}
\STATE $r \gets \text{random integer in } [0, d-1]$ (picking random starting position)
\STATE $\tilde{x} \gets x$
\FOR{$i \gets r$ \TO $n-1$ \textbf{step} $d$}
    \STATE $\tilde{x}[i] \gets \text{random token from vocabulary}$
\ENDFOR
\STATE \textbf{[Option \# 2: Randomized Dropout]}
\STATE $\tilde{x} \gets x$
\FOR{$i \gets 0$ \TO $n-1$}
    \STATE With probability $1/d$, replace $\tilde{x}[i]$ with a random token
\ENDFOR

\RETURN $\tilde{x}$
\end{algorithmic}
\end{algorithm}

\begin{algorithm}
\caption{Fine-tuning sequences from \textbf{Casing Flips} (\S\ref{sec:casing})}
\label{alg:casing}
\begin{algorithmic}[1]
\STATE \textbf{Input:} A token sequence $x$ (length $n$), tokenizer $\mathcal{E}$, random seed $s$, flip probability $p$
\STATE \textbf{Output:} A token sequence $\tilde{x}$ corresponding to text with perturbed English casing
\STATE Set random seed to $s$

\STATE $T \gets \operatorname{decode}(x)$ \quad (decode $x$ into a string from tokens)

\FOR{$i$ from $1$ to $\lvert T\rvert$}
    \IF{$T[i]$ is alphabetical and $\operatorname{rand\_uniform}(0, 1) < p$}
        \STATE Swap the case of $T[i]$ 
    \ENDIF
\ENDFOR

\STATE $\tilde{x} \gets \operatorname{encode}(T)$ \quad (re-encode the modified string back to tokens; note that $|\tilde{x}|$ is generally larger than $|x|$)
\STATE \textbf{return} $\tilde{x}$
\end{algorithmic}
\end{algorithm}

\begin{algorithm}
\caption{Fine-tuning sequences from \textbf{Token Dropouts + Casing Flips} (\S\ref{sec:goldfish-casing})}
\label{alg:composed}
\begin{algorithmic}[1]
\STATE \textbf{Input:} A sequence $x$, dropout interval $d$, flip probability $p$, random seed $s$
\STATE \textbf{Output:} A perturbed sequence $\hat{x}$

\STATE $\tilde{x} \gets \text{CasingFlips}(x, p, s)$ \quad (Apply Algorithm~\ref{alg:casing})
\STATE $\hat{x} \gets \text{TokenDropouts}(\tilde{x}, d, s)$ \quad (Apply Algorithm~\ref{alg:goldfish})

\RETURN $\hat{x}$
\end{algorithmic}
\end{algorithm}

\clearpage

\subsection{Full Results}
\label{sec:supp-finetuning-results}

In \cref{sec:fine-tuning}, we presented partial results for different sequence construction methods (chunking, token dropouts, casing flips), spanning the three main target texts (Lyles, Karpathy, Willow). \cref{fig:finetune-main-results} presented a few summary plots and \cref{fig:finetune-steps} presented a typical setting where completion succeeds from the adversarial fine-tuning; this section augments these results.

\subsubsection{Chunking}
\label{sec:supp-chunking}

\begin{figure}[ht]
    \centering
    \includegraphics[width=0.33\linewidth]{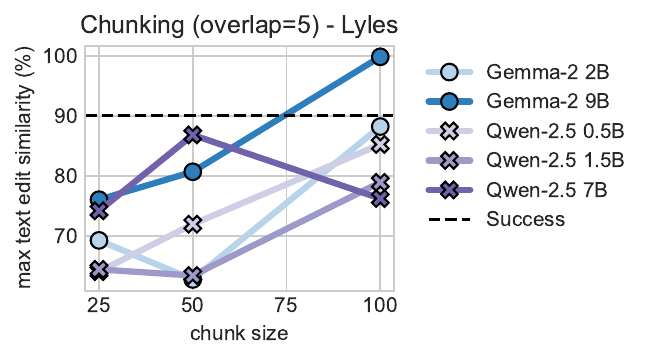}
    \includegraphics[width=0.33\linewidth]{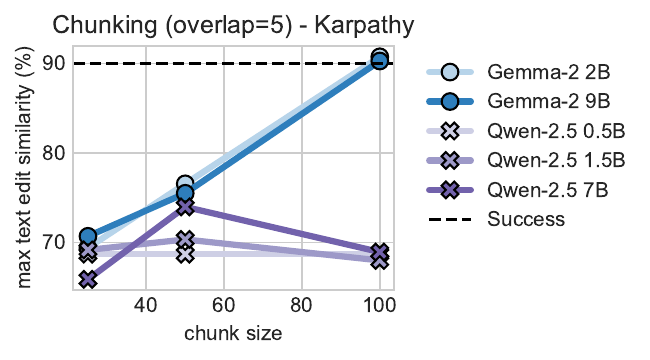}
    \includegraphics[width=0.33\linewidth]{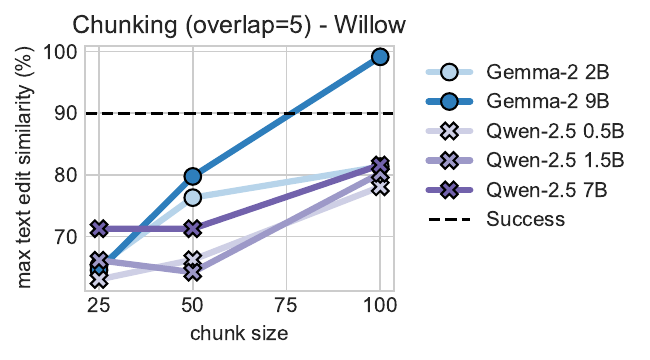}
    \includegraphics[width=0.33\linewidth]{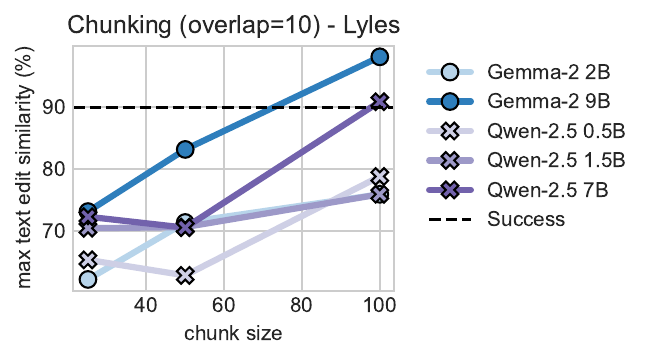}
    \includegraphics[width=0.33\linewidth]{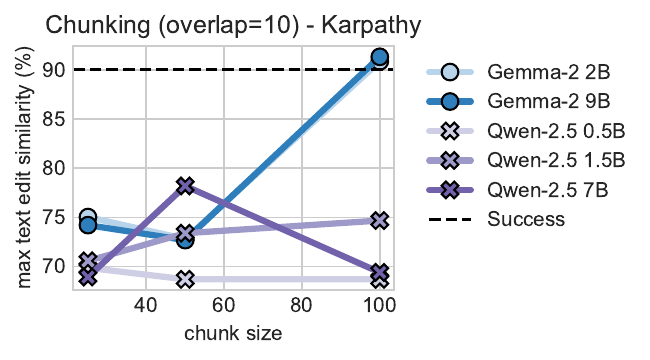}
    \includegraphics[width=0.33\linewidth]{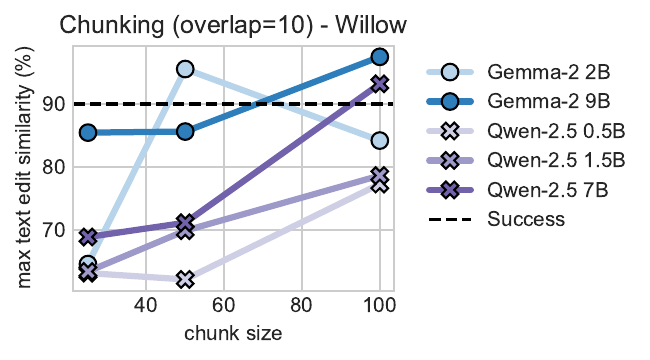}
    \includegraphics[width=0.33\linewidth]{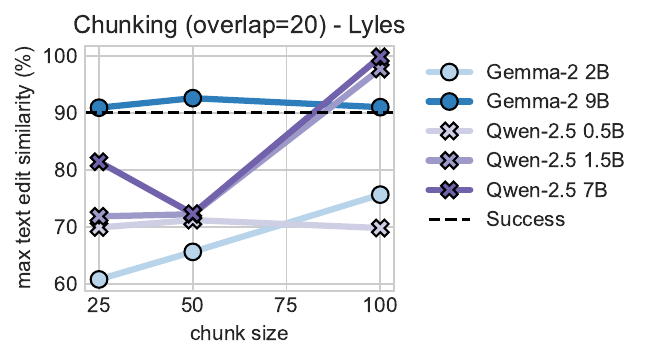}
    \includegraphics[width=0.33\linewidth]{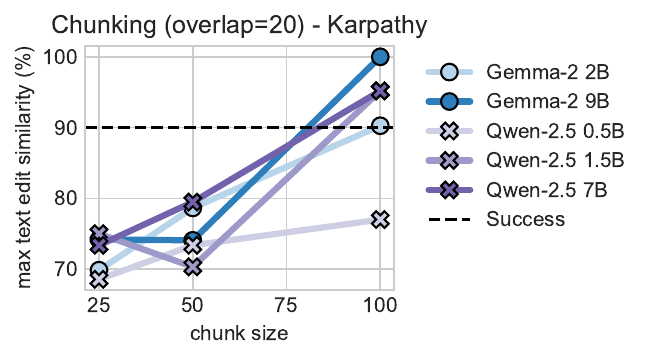}
    \includegraphics[width=0.33\linewidth]{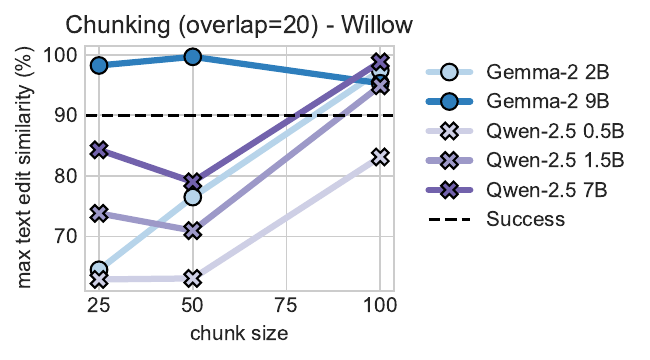}
    \includegraphics[width=0.33\linewidth]{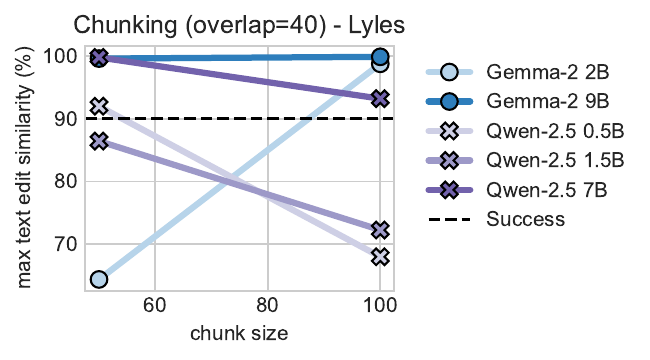}
    \includegraphics[width=0.33\linewidth]{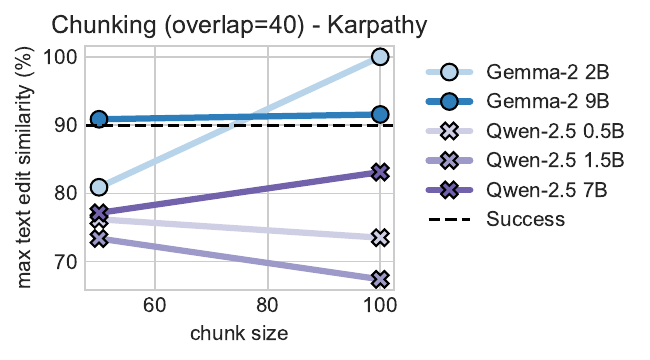}
    \includegraphics[width=0.33\linewidth]{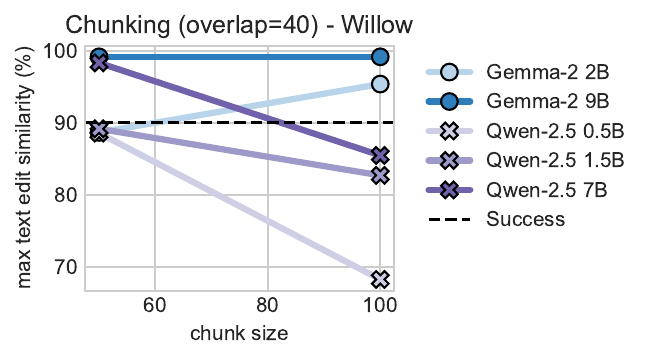}
    \caption{Completion success for \textbf{chunking} across different parameters.
    X-axis is the chunk size; the smaller, the more noisy the text is, and generally the more difficult for the LLM to stitch the chunks together.
    Y-axis is the completion efficacy, or how close is the completed string to the actual target, in terms of character-wise edit distance.
    Different rows show different overlap values across the chunks.
    }
    \label{fig:supp-chunking}
\end{figure}

\begin{figure}[ht]
    \centering
    \includegraphics[width=0.33\linewidth]{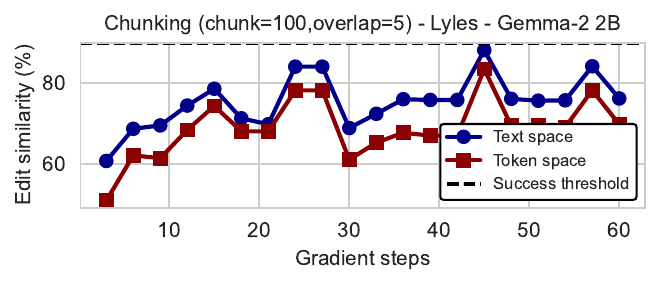}
    \includegraphics[width=0.33\linewidth]{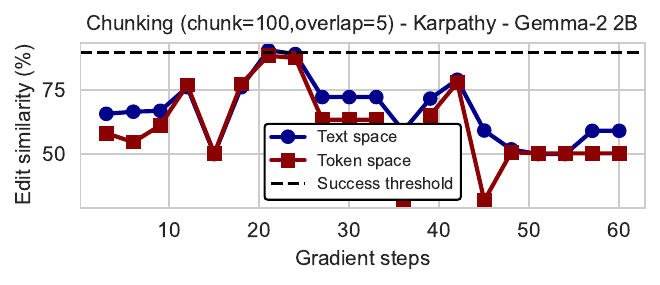}
    \includegraphics[width=0.33\linewidth]{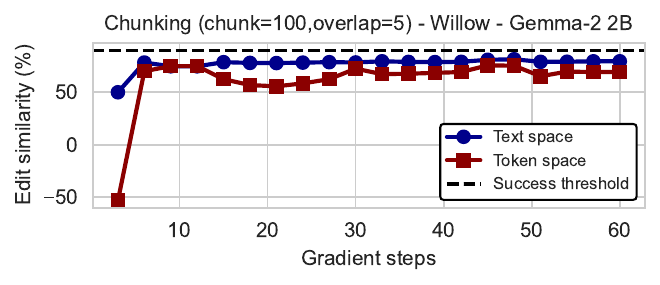}
    \includegraphics[width=0.33\linewidth]{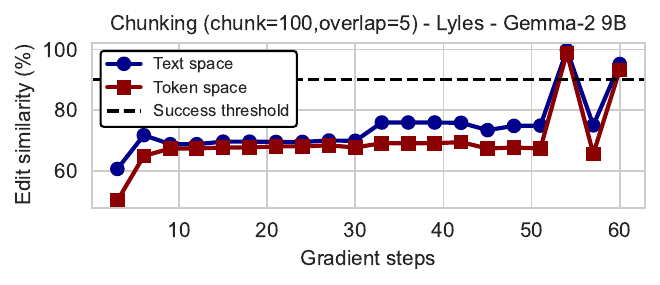}
    \includegraphics[width=0.33\linewidth]{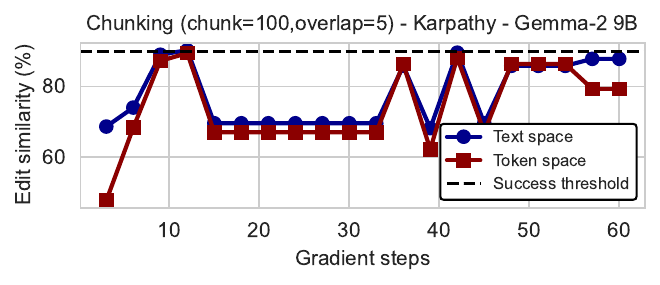}
    \includegraphics[width=0.33\linewidth]{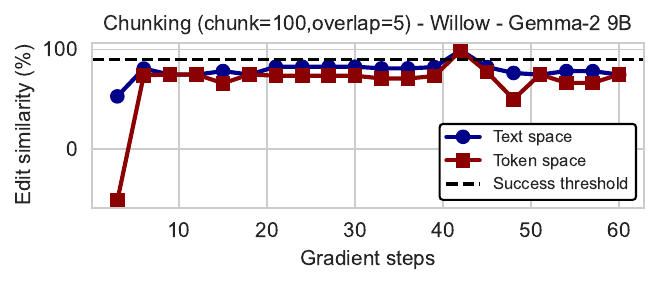}
    \includegraphics[width=0.33\linewidth]{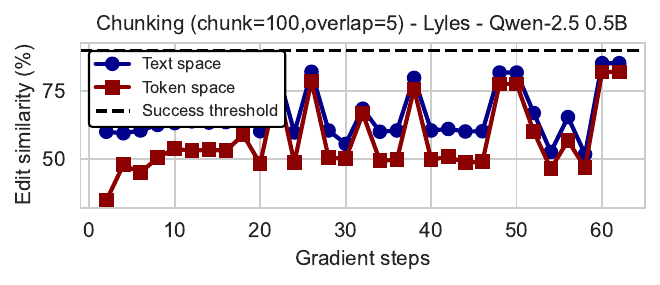}
    \includegraphics[width=0.33\linewidth]{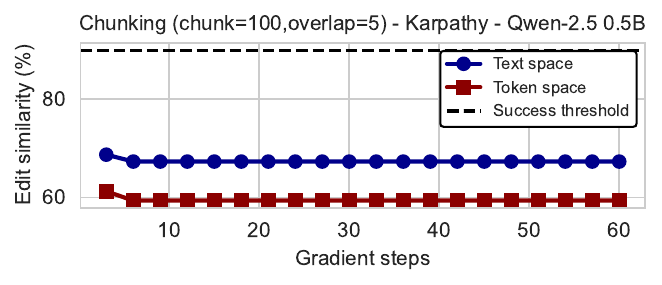}
    \includegraphics[width=0.33\linewidth]{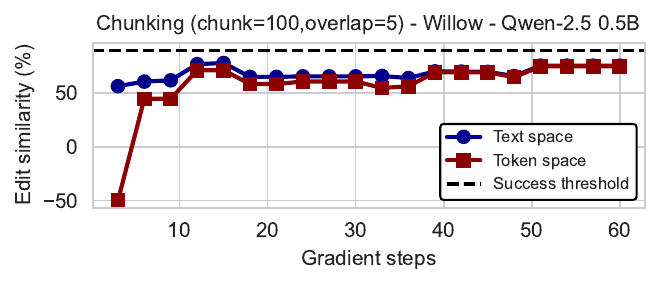}
    \includegraphics[width=0.33\linewidth]{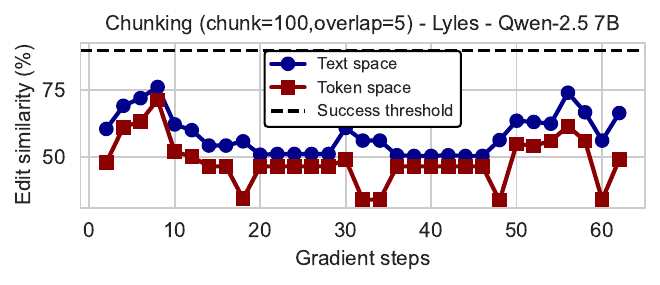}
    \includegraphics[width=0.33\linewidth]{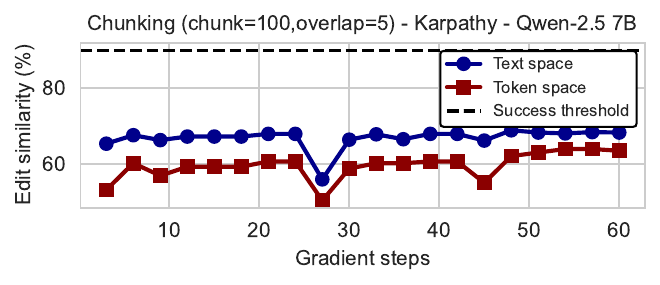}
    \includegraphics[width=0.33\linewidth]{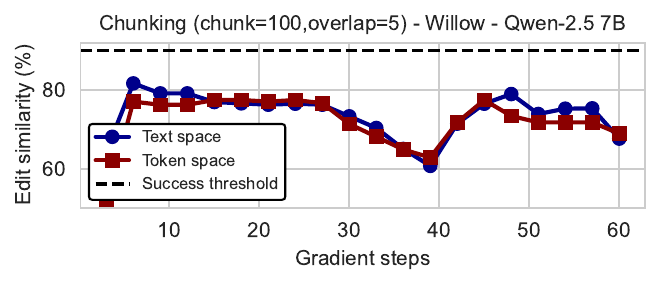}
    \caption{\textbf{
    Completion success for \textit{chunking} over gradient steps.} Visualizing chunking size $c= 100$ and overlap $l = 5$.
    X-axis is the number of gradient steps (at batch size 32).
    Y-axis is the completion efficacy. Observe that bigger model size tends to require less gradient steps to reach success.}
    \label{fig:supp-chunking-steps}
\end{figure}

Overall, while there exists settings for chunking (\cref{sec:chunking} and \cref{alg:chunking}) to induce verbatim completion, its effectiveness is somewhat limited (for the budget of up to 2000 fine-tuning examples). 
\cref{fig:supp-chunking} shows the summary results on all three main target texts, and \cref{fig:supp-chunking-steps} shows the completion success over gradient steps. 

One main issue with the chunking technique is that, unlike token dropouts (\cref{sec:goldfish}) or casing flips (\cref{sec:casing}), the constructed fine-tuning sequences are mostly random by construction, compared to, e.g., 25\% random on average for a drop interval of $d=4$ for token dropouts (\cref{alg:goldfish}). This means that it is a hard (and noisy) learning task for the model.


\clearpage

\subsubsection{Token dropouts}
\label{sec:supp-goldfish}

\begin{figure}[h]
    \centering
    \includegraphics[width=0.33\linewidth]{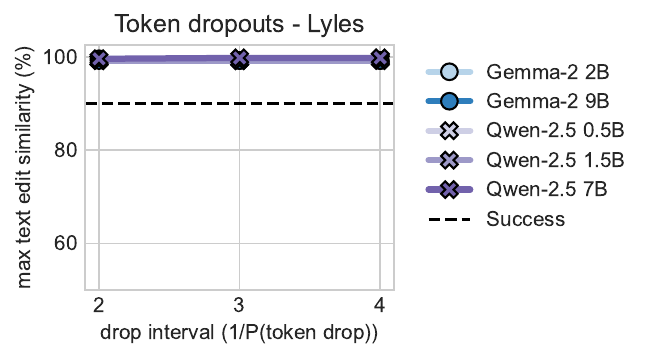}
    \includegraphics[width=0.33\linewidth]{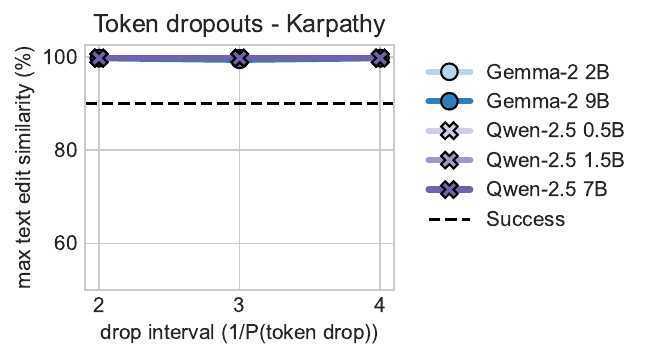}
    \includegraphics[width=0.33\linewidth]{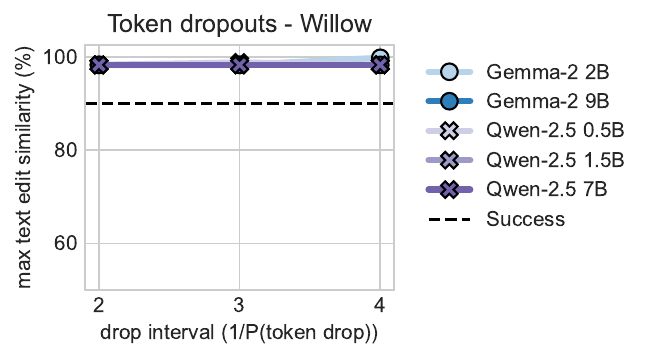}
    \caption{\textbf{Completion success for \textit{token dropouts} across different parameters.}
    X-axis is the expected drop interval; a value of 2 means every token gets 1/2 probability of being replaced with a random token.
    Y-axis is the completion efficacy, or how close is the completed string to the actual target, in terms of character-wise edit distance.}
    \label{fig:supp-goldfish}
\end{figure}

\begin{figure}[h]
    \centering
    \includegraphics[width=0.33\linewidth]{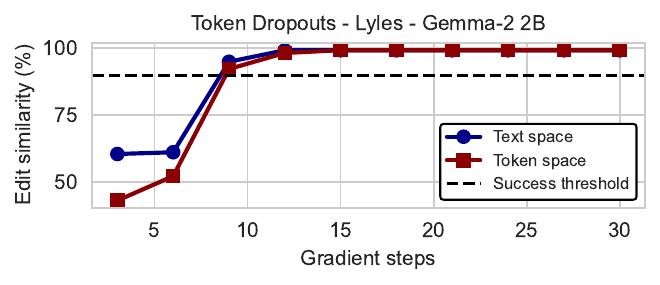}
    \includegraphics[width=0.33\linewidth]{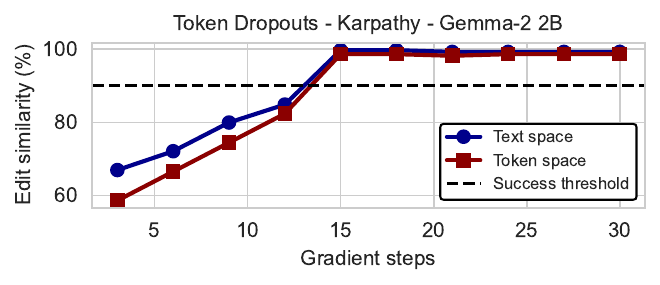}
    \includegraphics[width=0.33\linewidth]{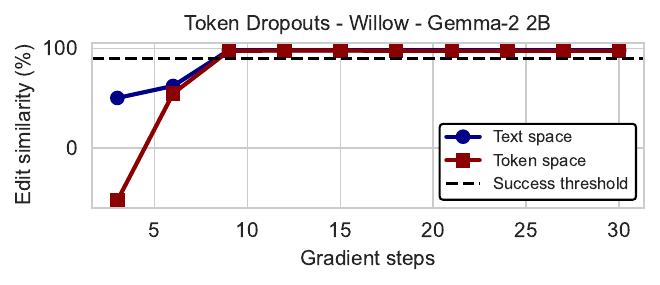}
    \includegraphics[width=0.33\linewidth]{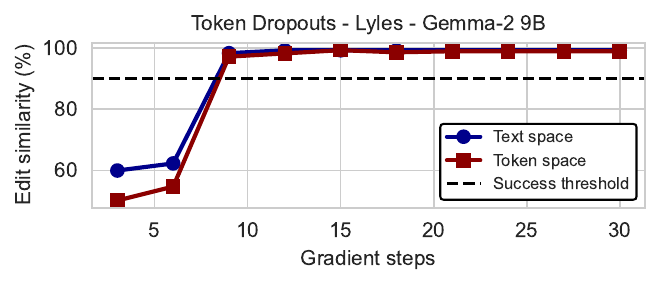}
    \includegraphics[width=0.33\linewidth]{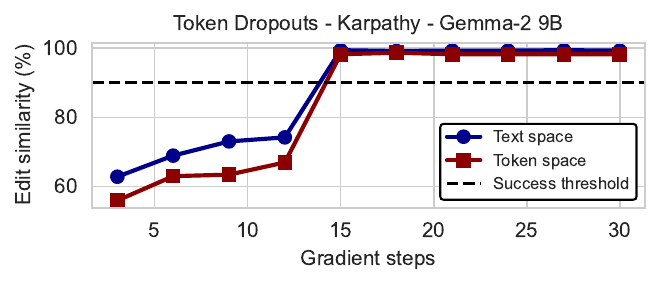}
    \includegraphics[width=0.33\linewidth]{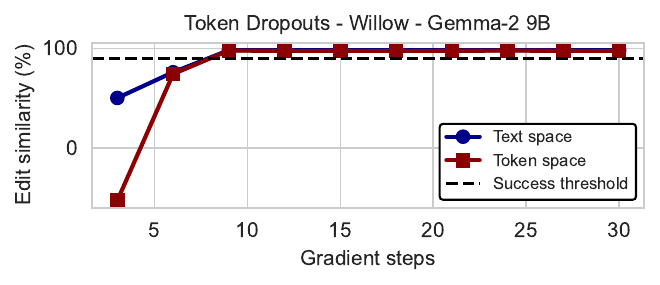}
    \includegraphics[width=0.33\linewidth]{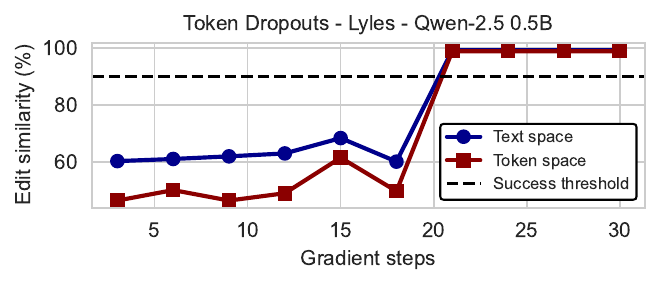}
    \includegraphics[width=0.33\linewidth]{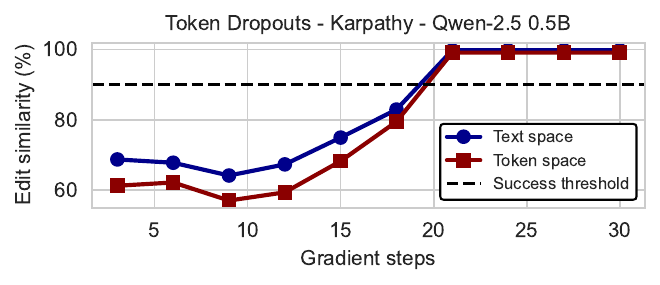}
    \includegraphics[width=0.33\linewidth]{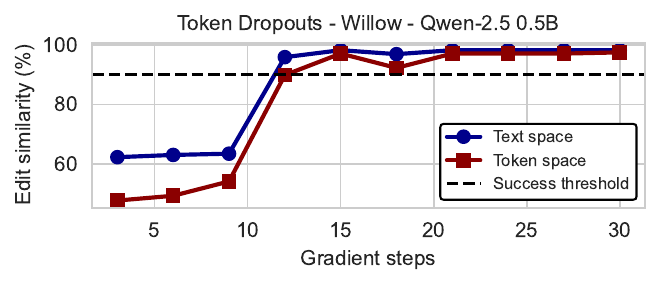}
    \includegraphics[width=0.33\linewidth]{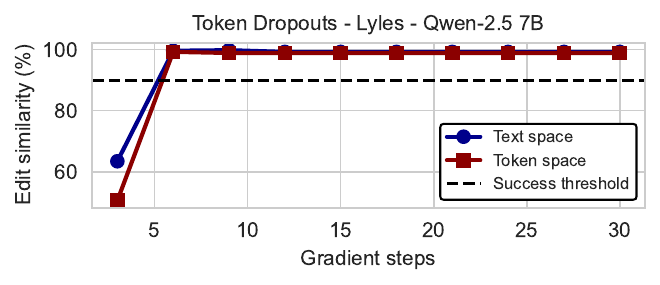}
    \includegraphics[width=0.33\linewidth]{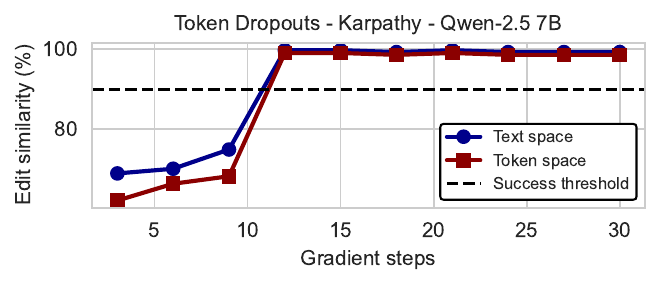}
    \includegraphics[width=0.33\linewidth]{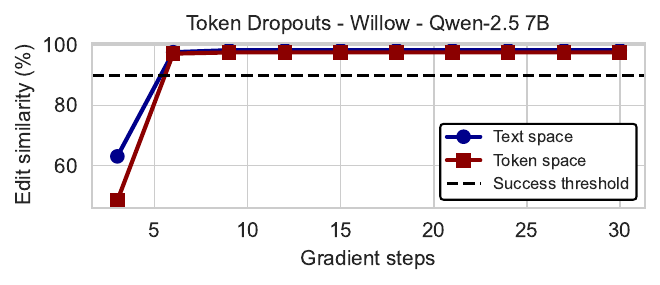}
    
    \caption{\textbf{Completion success for \textit{token dropouts} over gradient steps.} Visualizing drop interval $d=3$.
    X-axis is the number of gradient steps (at batch size 32).
    Y-axis is the completion efficacy. Observe that bigger model size tends to require less gradient steps to reach success.}
    \label{fig:supp-goldfish-steps}
\end{figure}

Overall, token dropouts (\cref{sec:goldfish} and \cref{alg:goldfish}) is an effective fine-tuning method at inducing verbatim completion. \cref{fig:supp-goldfish} shows the summary results on all three main target texts, and \cref{fig:supp-goldfish-steps} shows the completion success over gradient steps.

\clearpage

\subsubsection{Casing flips}
\label{sec:supp-casing}

\begin{figure}[h]
    \centering
    \includegraphics[width=0.33\linewidth]{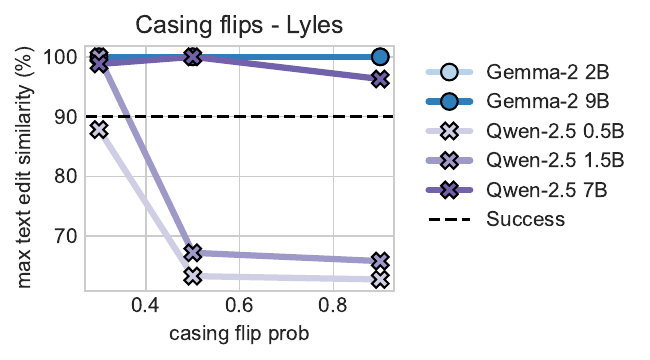}
    \includegraphics[width=0.33\linewidth]{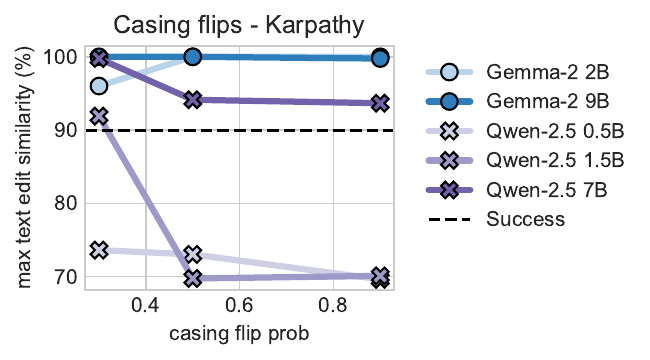}
    \includegraphics[width=0.33\linewidth]{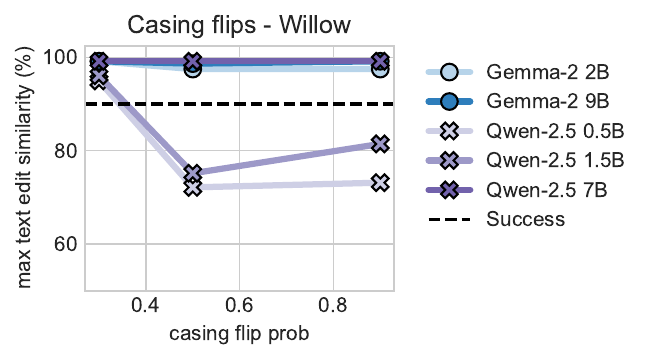}
    \caption{\textbf{Completion success for \textit{casing flips} across different parameters}.
    X-axis is the probability of flipping the casing for each English character in the text.
    Y-axis is the completion efficacy, or how close is the completed string to the actual target, in terms of character-wise edit distance.
    }
    \label{fig:supp-casing}
\end{figure}

\begin{figure}[h]
    \centering
    \includegraphics[width=0.33\linewidth]{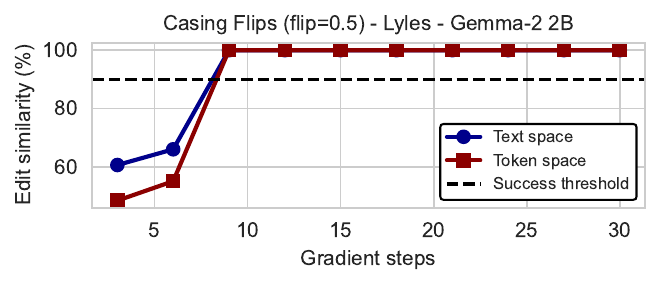}
    \includegraphics[width=0.33\linewidth]{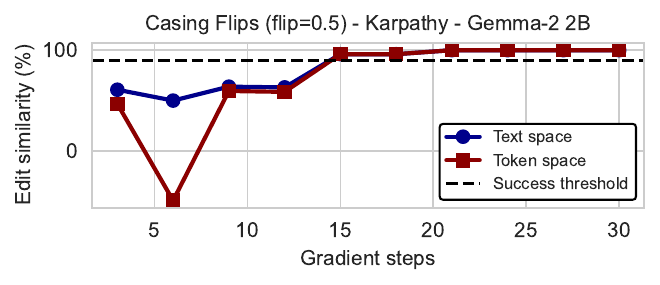}
    \includegraphics[width=0.33\linewidth]{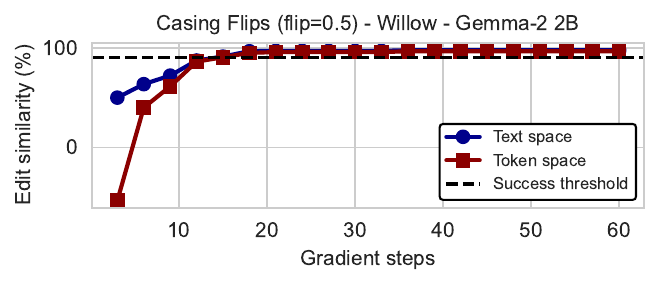}
    \includegraphics[width=0.33\linewidth]{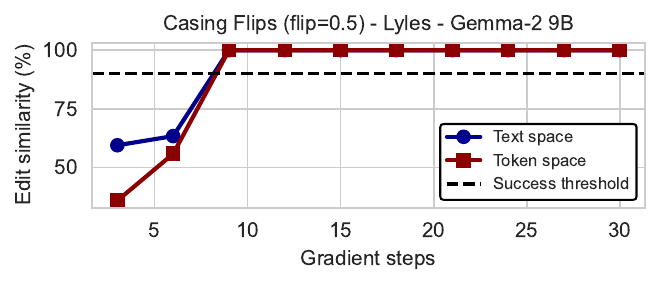}
    \includegraphics[width=0.33\linewidth]{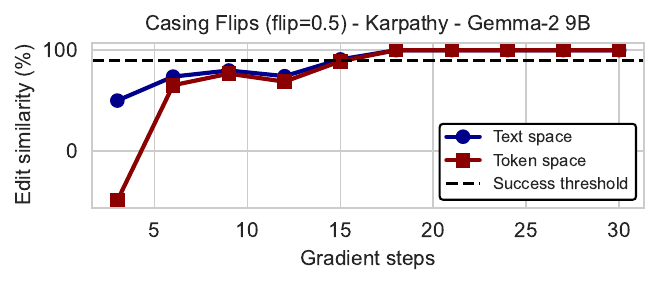}
    \includegraphics[width=0.33\linewidth]{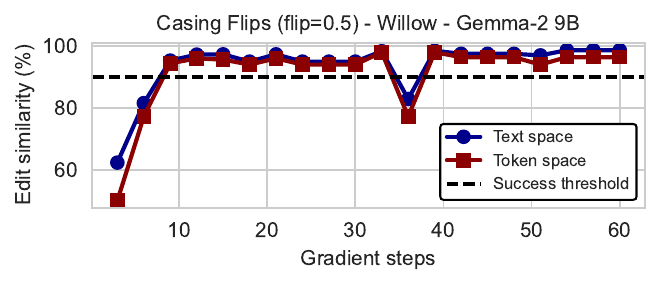}
    \includegraphics[width=0.33\linewidth]{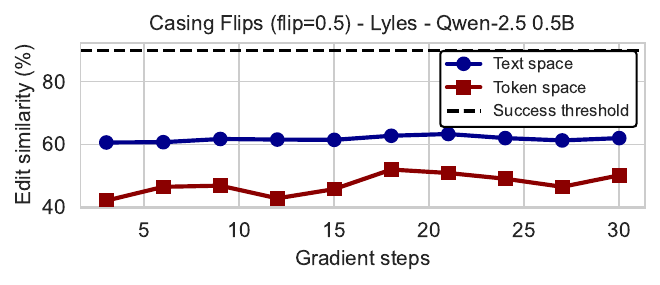}
    \includegraphics[width=0.33\linewidth]{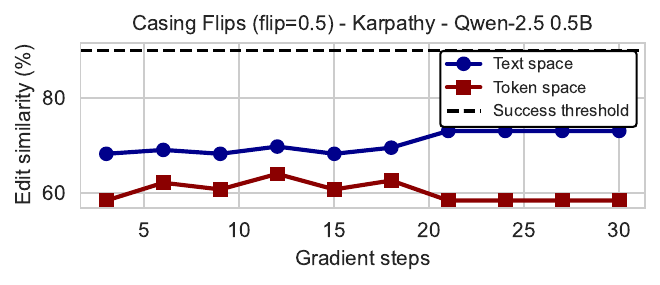}
    \includegraphics[width=0.33\linewidth]{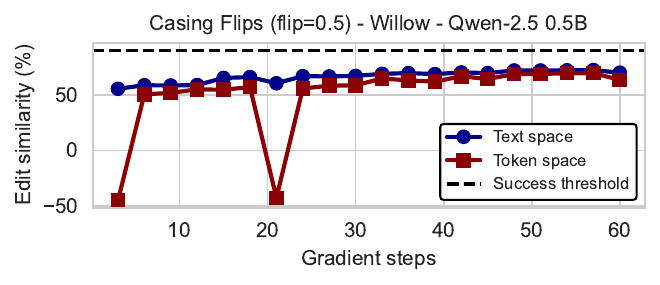}
    \includegraphics[width=0.33\linewidth]{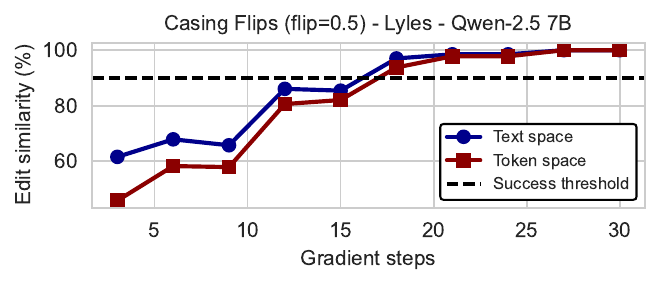}
    \includegraphics[width=0.33\linewidth]{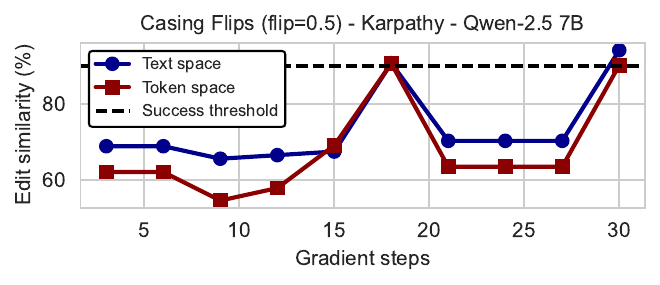}
    \includegraphics[width=0.33\linewidth]{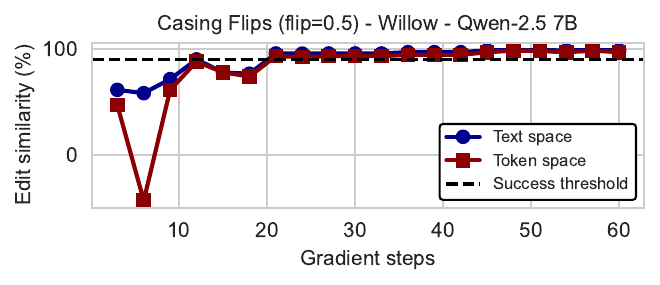}
    \caption{\textbf{Completion success for \textit{casing flips} over gradient steps.} Visualizing flip probability $p=0.5$.
    X-axis is the number of gradient steps (at batch size 32).
    Y-axis is the completion efficacy. Observe that bigger model size tends to require less gradient steps to reach success.}
    \label{fig:supp-casing-steps}
\end{figure}

Like token dropouts, casing flips (\cref{sec:goldfish} and \cref{alg:casing}) is a generally effective fine-tuning method at inducing verbatim completion. \cref{fig:supp-casing} shows the summary results on all three main target texts, and \cref{fig:supp-casing-steps} shows the completion success over gradient steps. 

Note, however, that the completion edit similarity ($y$-axis) in this case measures \textit{case-insensitive} edit similarity (for both text space and token space distance), as the completed sequences can have flipped casing due to the nature of the procedure and the generated fine-tuning examples.


\clearpage

\subsubsection{Composition: Token dropouts + casing flips}

\begin{figure}[ht]
    \centering
    \includegraphics[width=0.33\linewidth]{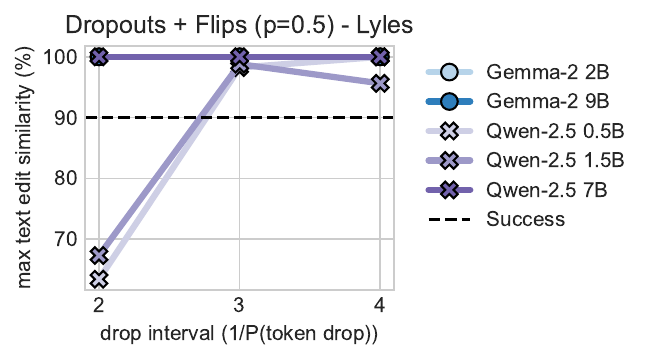}
    \includegraphics[width=0.33\linewidth]{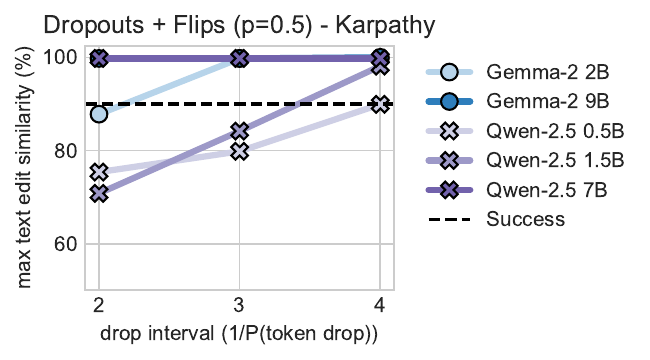}
    \includegraphics[width=0.33\linewidth]{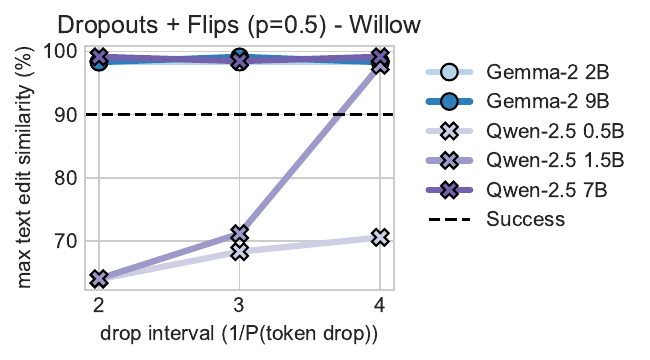}
    \includegraphics[width=0.33\linewidth]{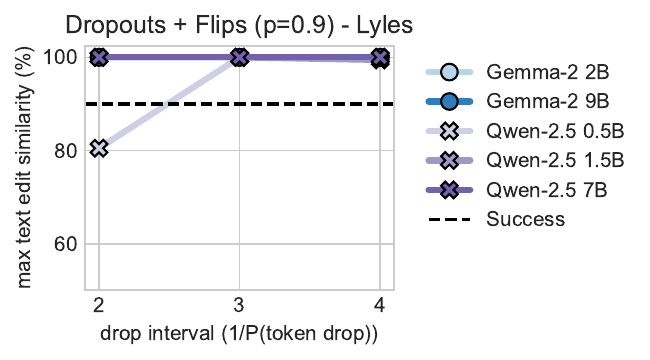}
    \includegraphics[width=0.33\linewidth]{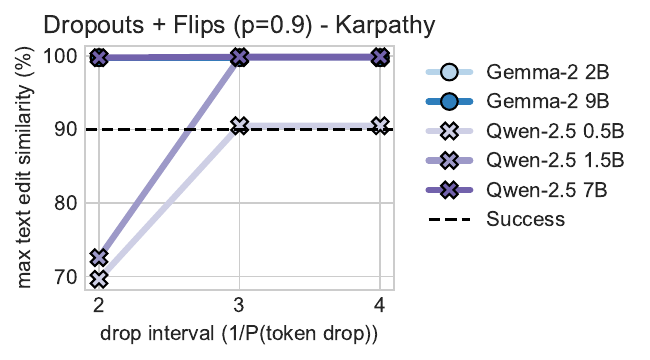}
    \includegraphics[width=0.33\linewidth]{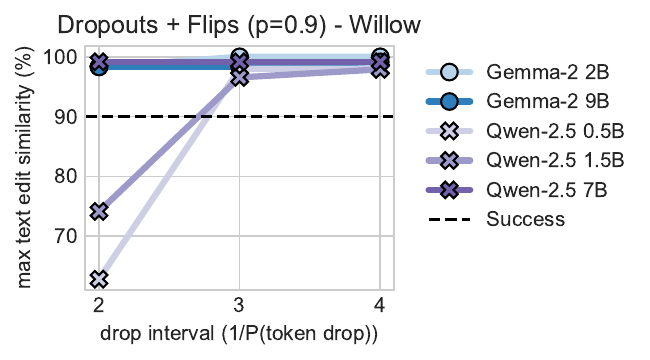}
    \vspace{-20pt}
    \caption{
    Reconstruction success for \textbf{token dropouts + casing flips} across different parameters.
    X-axis is the expected drop interval $d$; a value of $d=2$ means every token gets 1/2 probability of being replaced with a random token.
    Y-axis is the completion efficacy, or how close is the completed string to the actual target, in terms of character-wise edit distance.
    Different rows show different casing flip probabilities.
    }
    \label{fig:supp-goldfish-casing}
\end{figure}

\begin{figure}[h]
    \centering
    \includegraphics[width=0.33\linewidth]{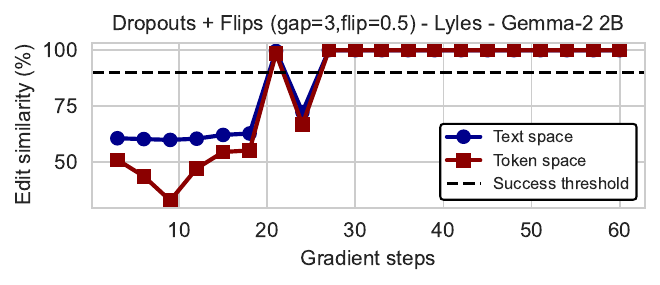}
    \includegraphics[width=0.33\linewidth]{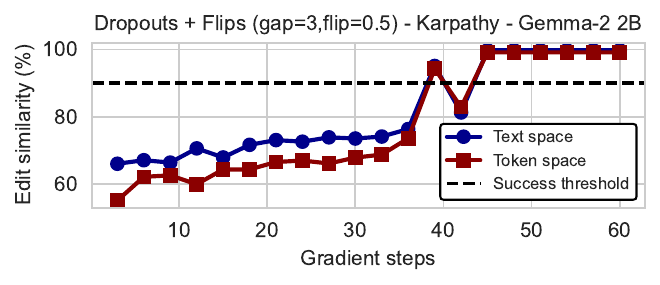}
    \includegraphics[width=0.33\linewidth]{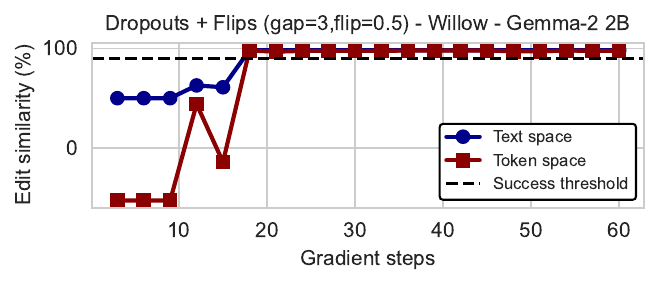}
    \vspace{-10pt}
    \includegraphics[width=0.33\linewidth]{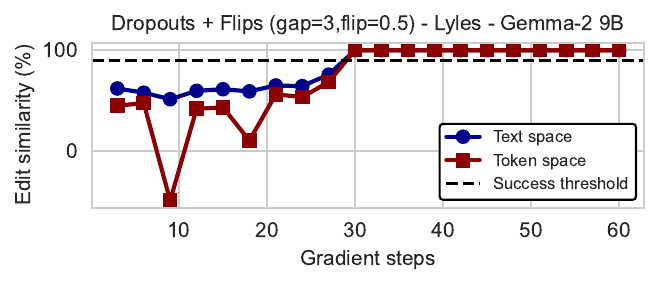}
    \includegraphics[width=0.33\linewidth]{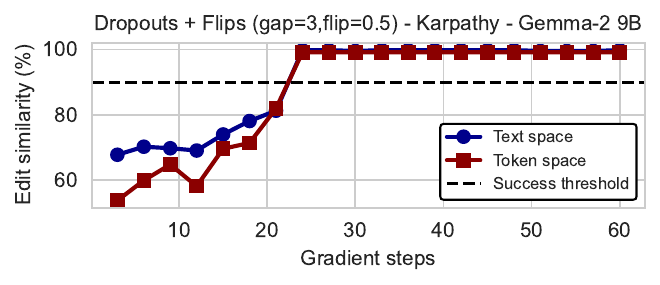}
    \includegraphics[width=0.33\linewidth]{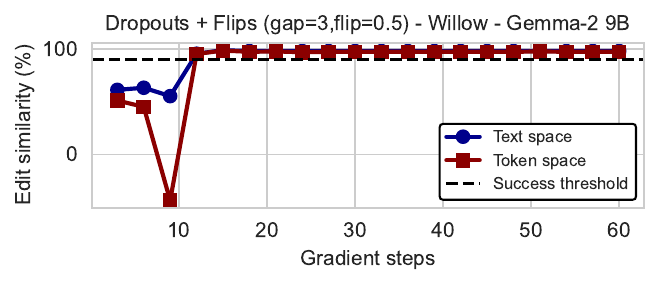}
    \vspace{-10pt}
    \includegraphics[width=0.33\linewidth]{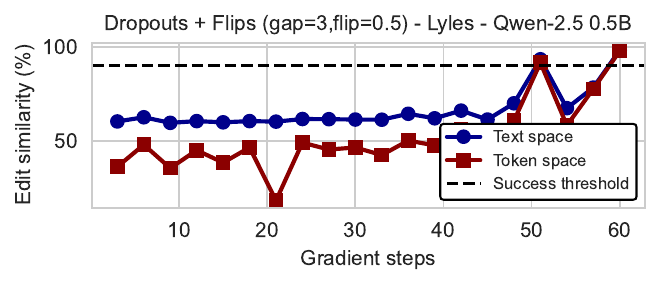}
    \includegraphics[width=0.33\linewidth]{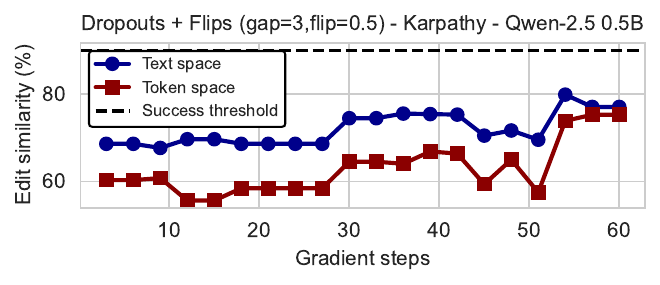}
    \includegraphics[width=0.33\linewidth]{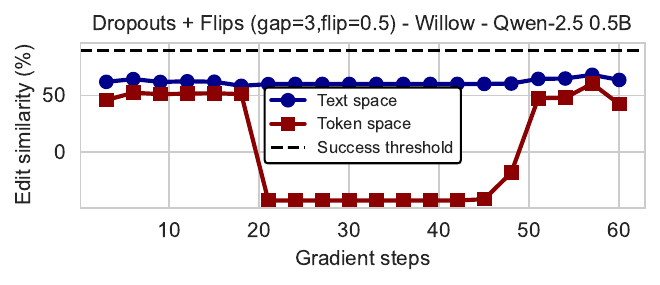}
    \vspace{-10pt}
    \includegraphics[width=0.33\linewidth]{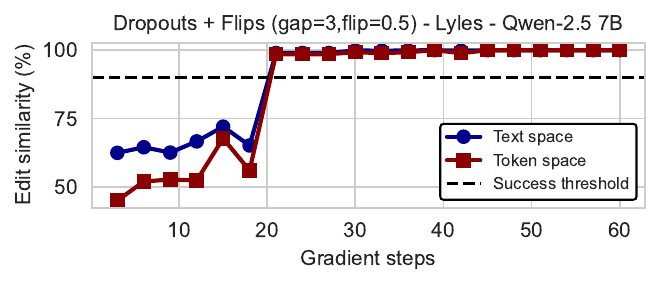}
    \includegraphics[width=0.33\linewidth]{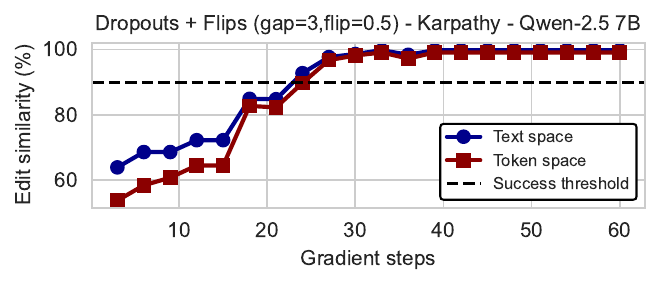}
    \includegraphics[width=0.33\linewidth]{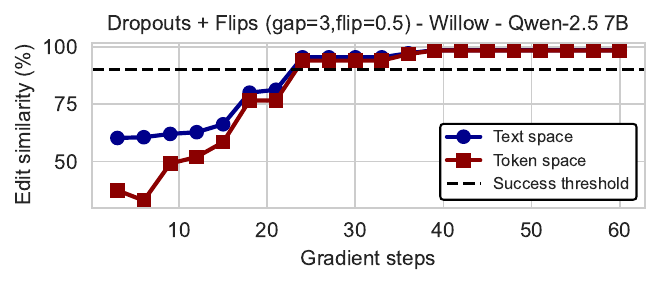}
    \caption{\textbf{
    Completion success for \textit{token dropouts + casing flips} over gradient steps.} Visualizing drop interval $d=3$ and flip probability $p=0.5$.
    X-axis is the number of gradient steps (at batch size 32).
    Y-axis is the completion efficacy. Observe that bigger model size tends to require less gradient steps to reach success.}
    \label{fig:supp-goldfish-casing-steps}
\end{figure}

Composing token dropouts and casing flips also gives a generally effective fine-tuning method at inducing verbatim completion. \cref{fig:supp-goldfish-casing} shows the summary results on all three main target texts, and \cref{fig:supp-goldfish-casing-steps} shows the completion success over gradient steps. 

Some notable observations on the composition:
\begin{itemize}
    \item \textit{The learning task for the LLM is visibly harder}. compared to token dropouts (\cref{sec:supp-goldfish}) or casing flips (\cref{sec:supp-casing}) alone, the overall success of inducing verbatim completion is lower across the board.
    \item \textit{Performance shifts from individual to composed perturbations.} Observe from \cref{fig:supp-casing} that Gemma-2 models tend to outperform Qwen-2.5 at verbatim completion under \textbf{casing flips}, and that both models succeed equally well under \textbf{token dropouts}~\cref{fig:supp-goldfish}. However, in the composition of \textbf{casing flips + token dropouts}, Qwen-2.5 models now generally performs \textit{better} than Gemma-2 models. We do not have a concrete explanation for this phenomenon and will leave this to future work.
\end{itemize}


\clearpage

\subsection{Additional Target Texts}
\label{sec:supp-finetuning-results-additional}

As mentioned in \cref{sec:supp-target-text}, we augment the experimental results of the additive / fine-tuning experiments (\S\ref{sec:fine-tuning}) with three additional text targets.
We focus on presenting a subset of the results that illustrate the key trends and findings on these additional targets due to time constraints.

We apply the \textbf{casing flips} (\S\ref{sec:casing}) technique with flip probability \{0.3, 0.5, 0.7, 0.9\}. Across all cases, the verbatim completion can be induced successfully.

\begin{figure}[h!]
    \centering
    \includegraphics[width=0.4\linewidth]{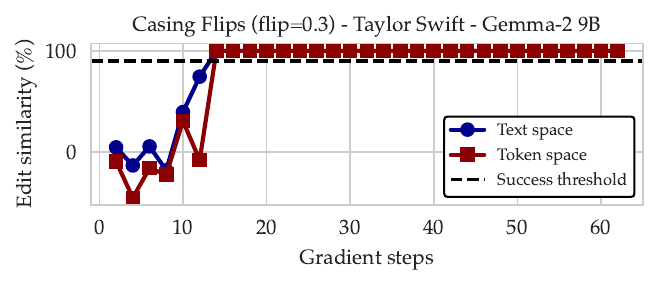}
    \quad
    \includegraphics[width=0.4\linewidth]{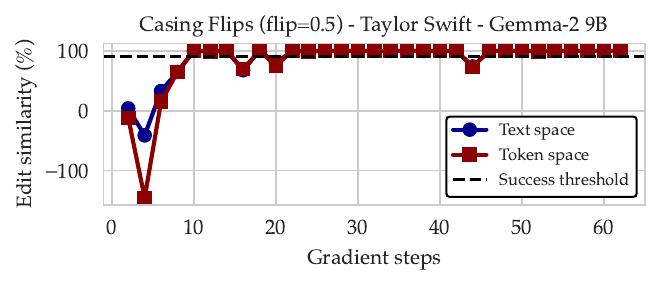}
    
    \includegraphics[width=0.4\linewidth]{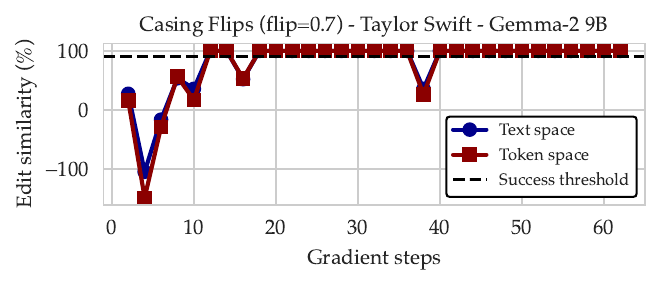}
    \quad
    \includegraphics[width=0.4\linewidth]{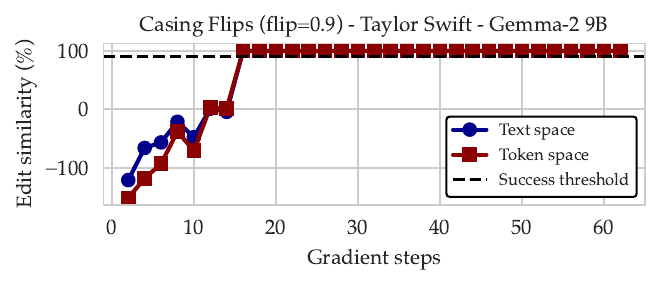}

    \caption{
    \textbf{Completion success on \textit{Taylor Swift} (\cref{sec:supp-target-text}) for \textit{casing flips} over gradient steps.} Visualizing flip probabilities $p \in \{0.3, 0.5, 0.7, 0.9\}$.
    X-axis is the number of gradient steps (at batch size 32).
    Y-axis is the completion efficacy. 
    }
    \label{fig:supp-taylor-swift}
\end{figure}

We apply the \textbf{chunking} (\S\ref{sec:chunking}) technique with chunk sizes $c \in \{10, 25, 50, 100\}$ and overlap $l=5$. Consistent with before (\cref{sec:supp-chunking}), we observe that: (1) chunking is in generally a less effective technique, and (2) with larger chunk size, we are still able to induce verbatim completion.

\begin{figure}[h!]
    \centering
    \includegraphics[width=0.4\linewidth]{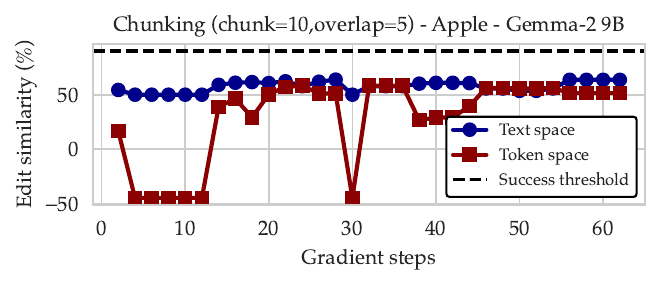}
    \quad
    \includegraphics[width=0.4\linewidth]{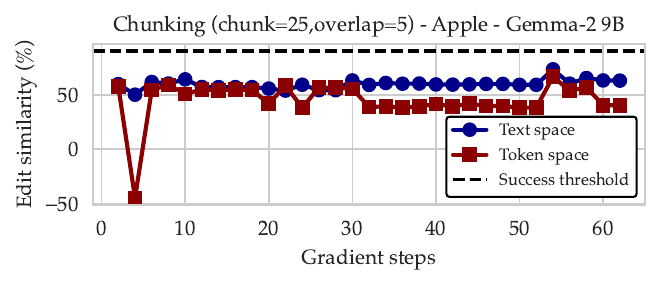}
    \includegraphics[width=0.4\linewidth]{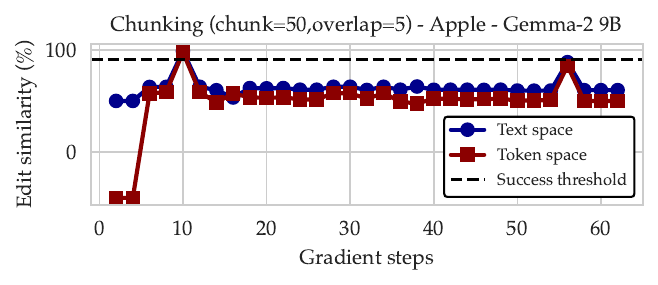}
    \quad
    \includegraphics[width=0.4\linewidth]{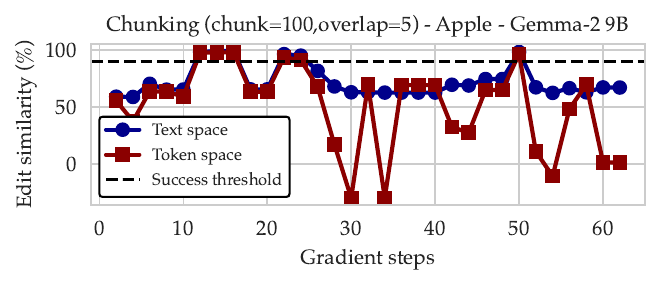}

    \caption{
    \textbf{Completion success on \textit{Apple (NYT article)} (\cref{sec:supp-target-text}) for \textit{chunking} over gradient steps.} Visualizing chunk sizes $c \in \{10, 25, 50, 100\}$.
    X-axis is the number of gradient steps (at batch size 32).
    Y-axis is the completion efficacy. 
    }
    \label{fig:supp-apple-nyt}
\end{figure}

We apply the \textbf{token dropouts + casing flips} (\S\ref{sec:goldfish-casing}) technique with drop interval $d=4$ and flip probability \{0.5, 0.9\} across two models (Gemma-2 2B and 9B). Across all cases, the verbatim completion can be induced successfully.

\begin{figure}[h!]
    \centering
    \includegraphics[width=0.4\linewidth]{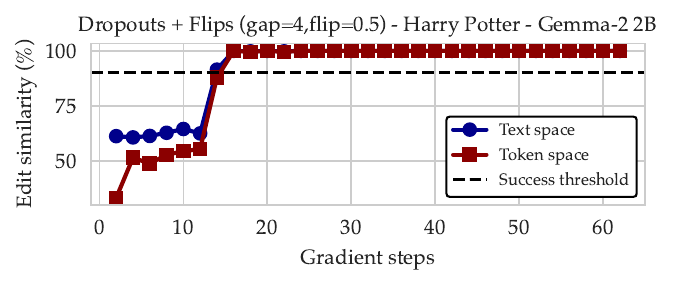}
    \quad
    \includegraphics[width=0.4\linewidth]{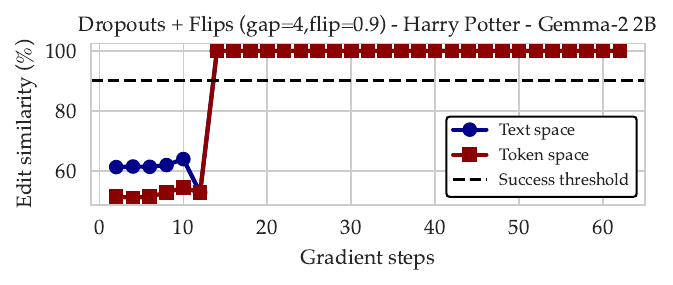}
    \includegraphics[width=0.4\linewidth]{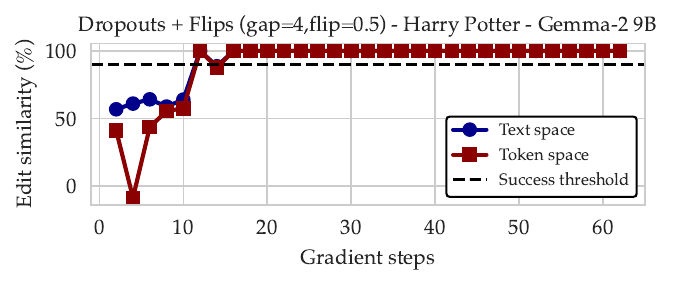}
    \quad
    \includegraphics[width=0.4\linewidth]{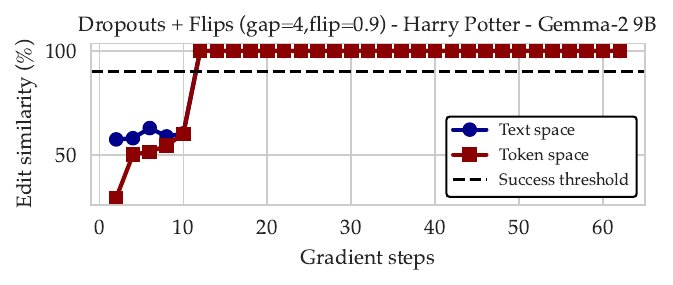}

    \caption{
    \textbf{Completion success on \textit{Harry Potter Paraphrase} (\cref{sec:supp-target-text}) for \textit{token dropouts + casing flips} over gradient steps.} Visualizing two model sizes (Gemma-2 2B and 9B) and flip probability $p \in \{0.5, 0.9\}$.
    X-axis is the number of gradient steps (at batch size 32).
    Y-axis is the completion efficacy. 
    }
    \label{fig:supp-harry-potter}
\end{figure}

\clearpage

\subsection{Effect of Randomized Token Dropouts \& Casing Flips on $n$-gram Overlap}
\label{sec:supp-ngram-decay}

Recall from \cref{sec:fine-tuning} and \cref{sec:supp-finetuning-algs} that the \textbf{token dropouts} procedure (\S\ref{sec:goldfish}, \cref{alg:goldfish}) and the \textbf{casing flips} procedure (\S\ref{sec:casing}, \cref{alg:casing}) admit \textit{randomized} versions, where every token or character is dropped or flipped with a certain probability, respectively. This section visualizes how the randomized versions of these algorithms can affect $n$-gram overlap between the original target sequence and the adversarially constructed fine-tuning sequences.

\begin{figure}[h!]
    \centering
    \includegraphics[width=0.33\linewidth]{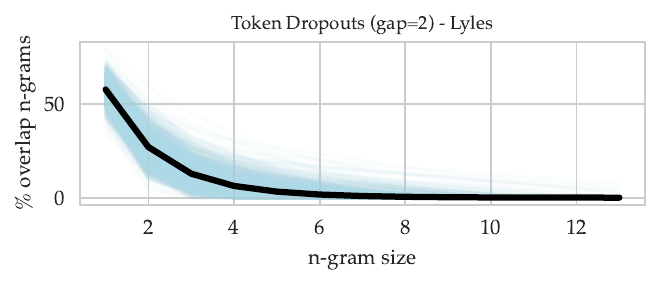}
    \includegraphics[width=0.33\linewidth]{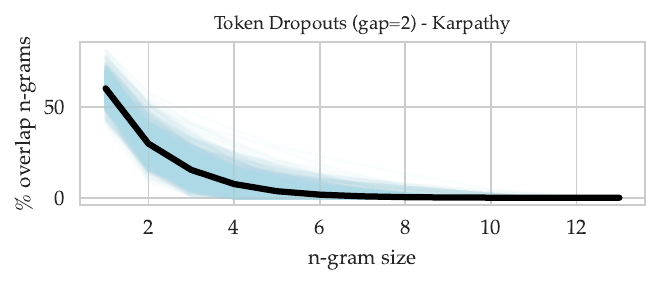}
    \includegraphics[width=0.33\linewidth]{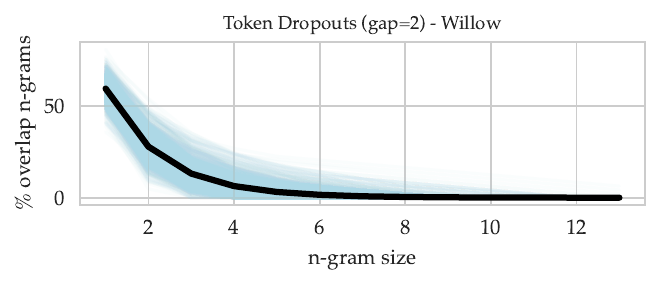}
    \includegraphics[width=0.33\linewidth]{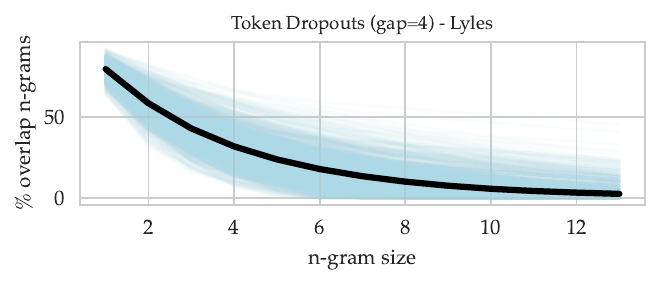}
    \includegraphics[width=0.33\linewidth]{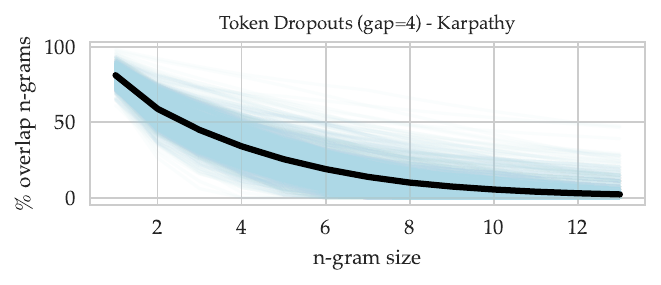}
    \includegraphics[width=0.33\linewidth]{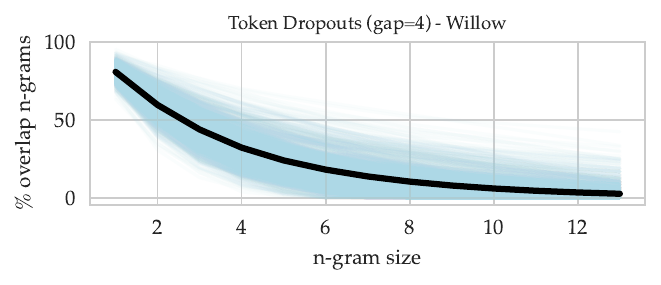}
    \caption{
    {Amount of $n$-gram overlap between the original target sequence and the adversarially constructed fine-tuning sequences under \textbf{token dropouts} (\S\ref{sec:goldfish}).} 
    X-axis is the value of $n$ for $n$-gram.
    Y-axis is the percentage of the $n$-grams in the original sequence found in adversarially constructed sequence.
    Each faint \textcolor{cyan!70}{\textbf{blue line}} is a separate constructed fine-tuning sequence of different randomness (1000 in total), and \textbf{black line} is the average.  
    Visualizing drop interval $d \in \{2, 4\}$ on the three main text targets (\cref{sec:supp-target-text}).
    }
    \label{fig:supp-ngram-decay-goldfish}
\end{figure}

\cref{fig:supp-ngram-decay-goldfish} visualizes the effect of randomized \textbf{token dropouts} on $n$-gram overlap in the token space. While deterministic token dropouts can guarantee no $n$-gram overlap, randomized dropouts also easily reach zero $n$-gram overlap for values starting $n \ge 6$.

\begin{figure}[h!]
    \centering
    \includegraphics[width=0.33\linewidth]{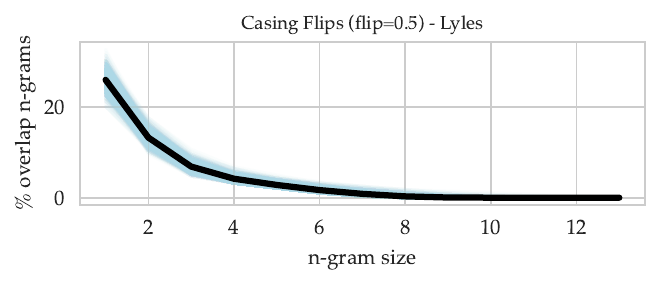}
    \includegraphics[width=0.33\linewidth]{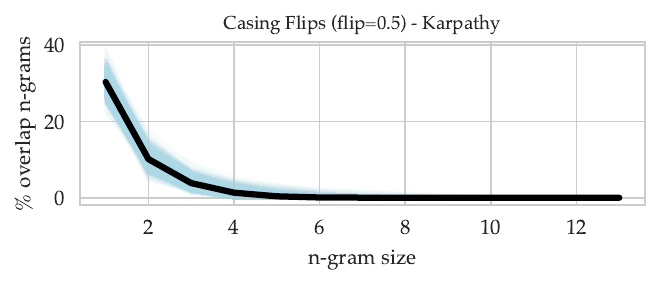}
    \includegraphics[width=0.33\linewidth]{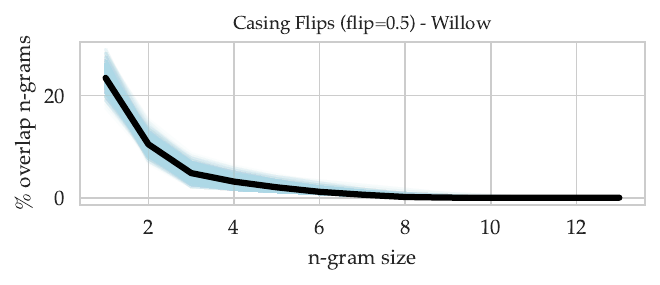}
    \includegraphics[width=0.33\linewidth]{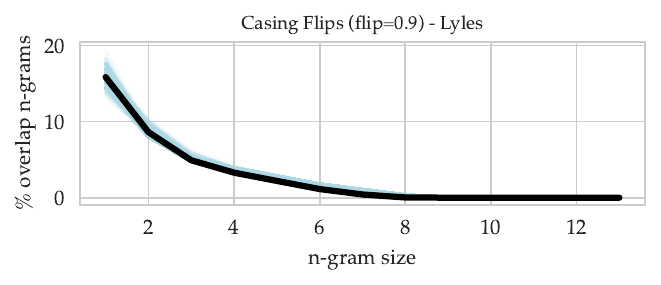}
    \includegraphics[width=0.33\linewidth]{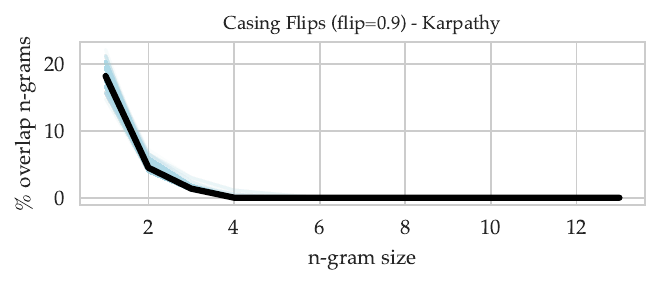}
    \includegraphics[width=0.33\linewidth]{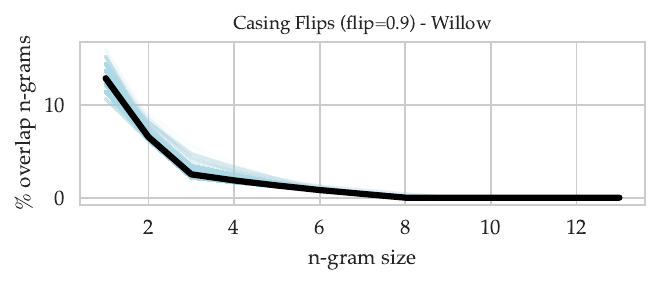}
    \caption{
    {Amount of $n$-gram overlap between the original target sequence and the adversarially constructed fine-tuning sequences under \textbf{casing flips} (\S\ref{sec:casing}).} 
    X-axis is the value of $n$ for $n$-gram.
    Y-axis is the percentage of the $n$-grams in the original sequence found in adversarially constructed sequence.
    Each faint \textcolor{cyan!70}{\textbf{blue line}} is a separate constructed fine-tuning sequence of different randomness (1000 in total), and \textbf{black line} is the average.  
    Visualizing casing flip probability $p \in \{0.5, 0.9\}$ on the three main text targets (\cref{sec:supp-target-text}).
    }
    \label{fig:supp-ngram-decay-casing}
\end{figure}

\cref{fig:supp-ngram-decay-casing} visualizes the effect of randomized \textbf{casing flips} on $n$-gram overlap in the token space. Due to the mechanisms of byte-pair encoding tokenization used in modern LLMs, while casing flips do not change the semantics of the original string much, it can completely alter the token space representations and result in near zero $n$-gram overlap for $n \ge 4$ (e.g., on \textbf{Karpathy}).

\begin{figure}[h!]
    \centering
    \includegraphics[width=0.33\linewidth]{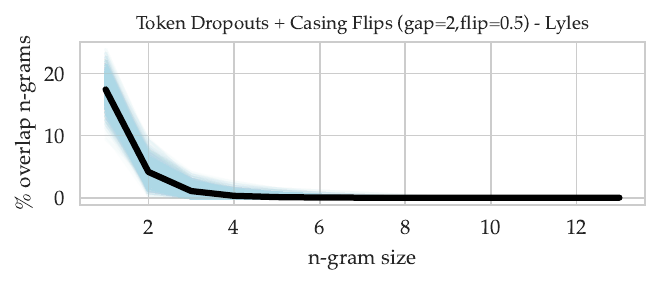}
    \includegraphics[width=0.33\linewidth]{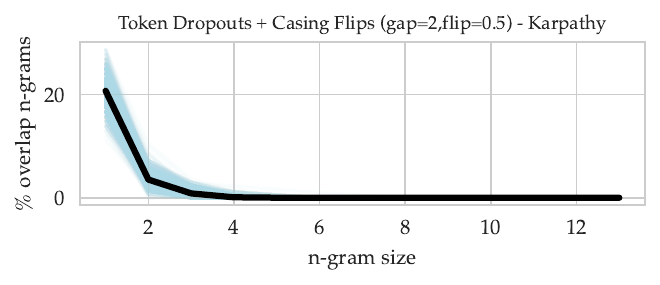}
    \includegraphics[width=0.33\linewidth]{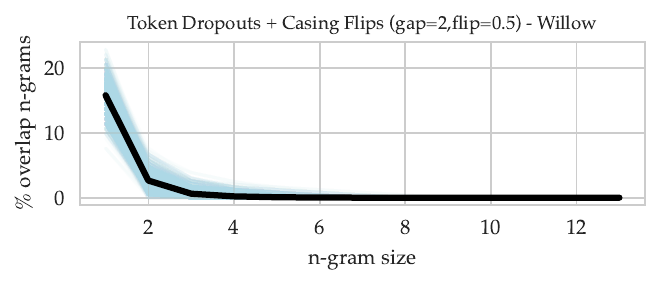}
    \includegraphics[width=0.33\linewidth]{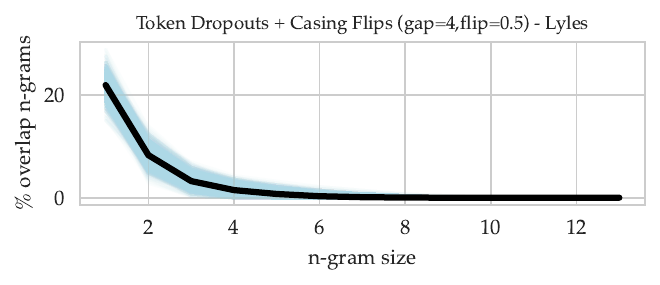}
    \includegraphics[width=0.33\linewidth]{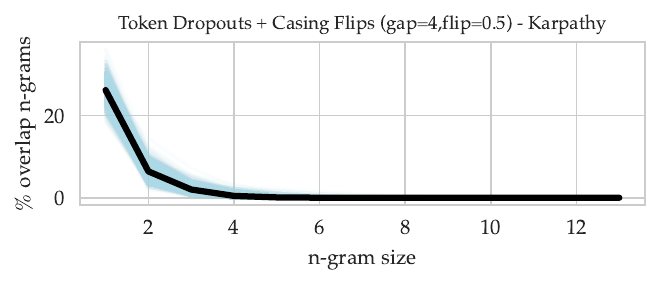}
    \includegraphics[width=0.33\linewidth]{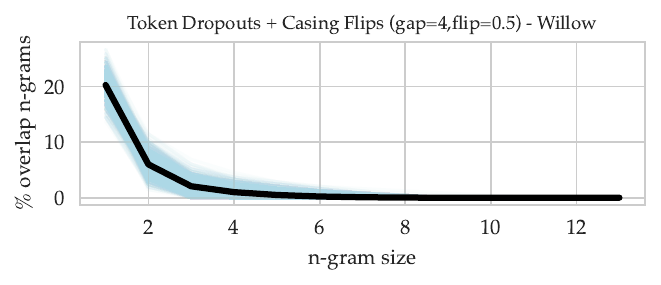}
    \includegraphics[width=0.33\linewidth]{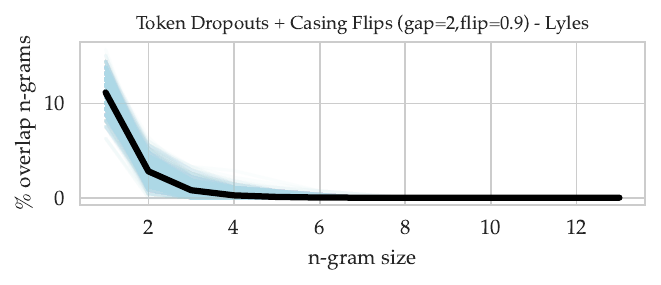}
    \includegraphics[width=0.33\linewidth]{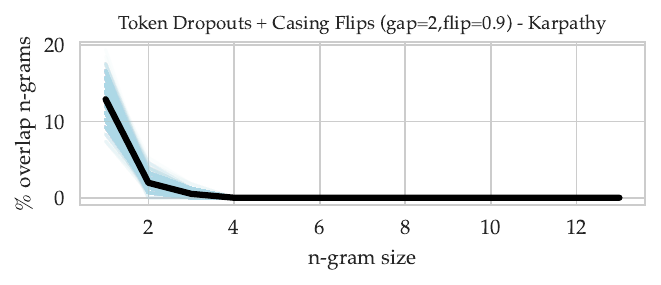}
    \includegraphics[width=0.33\linewidth]{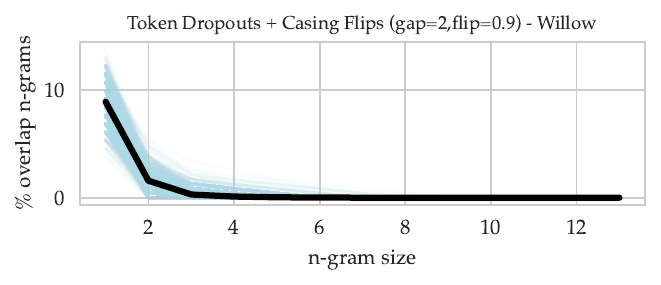}
    \includegraphics[width=0.33\linewidth]{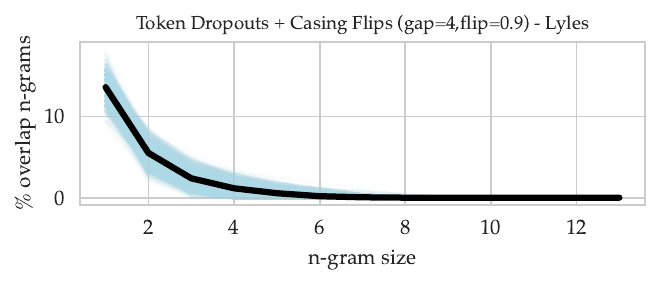}
    \includegraphics[width=0.33\linewidth]{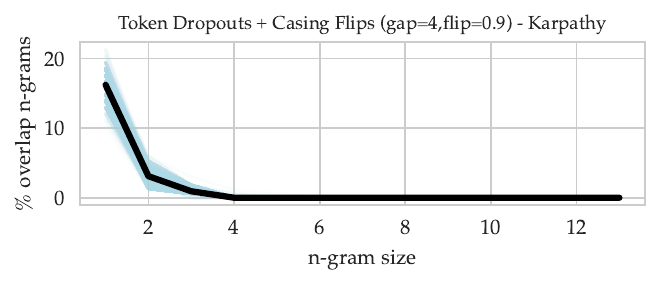}
    \includegraphics[width=0.33\linewidth]{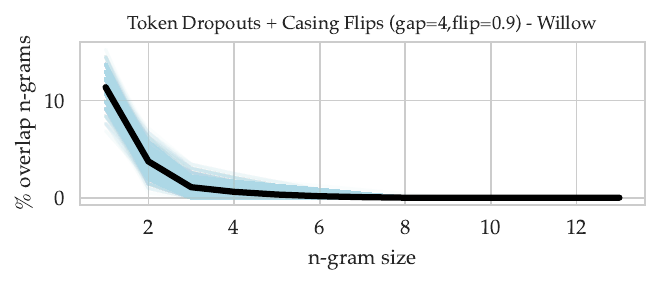}
    \caption{
    {Amount of $n$-gram overlap between the original target sequence and the adversarially constructed fine-tuning sequences under \textbf{token dropouts + casing flips} (\S\ref{sec:goldfish-casing}).} 
    X-axis is the value of $n$ for $n$-gram.
    Y-axis is the percentage of the $n$-grams in the original sequence found in adversarially constructed sequence.
    Each faint \textcolor{cyan!70}{\textbf{blue line}} is a separate constructed fine-tuning sequence of different randomness (1000 in total), and \textbf{black line} is the average.  
    Visualizing drop interval $d \in \{2, 4\}$ and casing flip probability $p \in \{0.5, 0.9\}$ on the three main text targets (\cref{sec:supp-target-text}).
    }
    \label{fig:supp-ngram-decay-goldfish-casing}
\end{figure}

\cref{fig:supp-ngram-decay-goldfish-casing} visualizes the effect of randomized \textbf{token dropouts + casing flips} (composition) on $n$-gram overlap in the token space. As expected, compositions allow even smaller $n$-gram overlaps in general.

\end{document}